%% file: main.tex
\documentclass{article} % For LaTeX2e
\usepackage{iclr2026_conference,times}

\input{template/math_commands}

\usepackage{hyperref}
\usepackage{url}
\usepackage{booktabs}   % 专业表格横线
\usepackage{setspace}   % 行间距控制
\usepackage{makecell}   % 表头多行文本
\usepackage{tabularx} % 新增：自动适配宽度的包
\usepackage{graphicx} % 新增：用于缩放表格（可选）
\usepackage{array}
\usepackage{subcaption} % 用于子图环境
\usepackage[export]{adjustbox} % 提供 valign 选项
\usepackage{calc} % 用于计算宽度
\usepackage{xcolor}
\usepackage{dashrule}
\usepackage{amsmath}  % 用于数学符号如 \lambda
\usepackage[table]{xcolor}

\definecolor{npccolor}{RGB}{70,130,200}
\definecolor{egocolor}{RGB}{225,130,110}
\definecolor{lanecolor}{RGB}{160,160,160}  % 中性浅灰，适配论文黑白打印效果
\definecolor{gtcolor}{RGB}{30,150,70}
\definecolor{predcolor}{RGB}{225,130,110}

\usepackage[utf8]{inputenc}
\usepackage{titletoc}
\titlecontents{section}[0em] % 左侧缩进 0em
{\vspace{12pt}\normalfont\bfseries\color[RGB]{0,102,204}} % 条目前添加间距，字体样式：正常、粗体、浅蓝色
{\thecontentslabel\quad} % 显示编号 (A, B) 并加空格
{} % 无编号时的格式
{\hfill\contentspage} % 无点线，直接右对齐页码
\titlecontents{subsection}[2em] % 左侧缩进 2em
{\normalfont\color[RGB]{0,102,204}} % 字体样式：正常、浅蓝色
{\thecontentslabel\quad} % 显示编号 (B.1)
{}
{\leaders\hbox to 0.5em{\hss.\hss}\hfill\contentspage} % 点线连接页码

\title{HyWorldVLA: A Vision-Language-Action Model with Hybrid World Modeling for Autonomous Driving}

\iclrfinalcopy
\author{
  Quanfu Yu\thanks{Equal contribution. $^\dagger$Corresponding author.}~, ~Xian Wu$^*$, ~Hao Xu$^*$, ~Liulong Ma$^\dagger$ \\
  Automotive New Technology Research Institute, BYD Company Limited \\
  \texttt{\{yqf2018619, wuxian7x, xuhaovic\}@gmail.com, ma.liulong@byd.com} \\
  \quad
}

\begin{document}

\thispagestyle{firstpagestyle}   % 第一页用带 logo 的样式
\pagestyle{otherpagestyle}      % 从第二页起改用无 logo 样式

\maketitle

\begin{abstract}
% Vision-Language-Action (VLA) models enhanced with world modeling have emerged as a promising paradigm for end-to-end autonomous driving. Predicting future visual empowers VLAs with robust spatiotemporal reasoning, yet full pixel reconstruction incurs redundancy calculation and high inference latency. Latent-world VLAs alleviate such efficiency burdens by predicting future states in feature space, but suffer from poor interpretability and potential representation degeneration due to the absence of pixel-level grounding. To address these limitations, we present HyWorldVLA, a hybrid world-VLA framework integrating pixel supervision and latent learning. In pre-training, it predicts video latents encoded by a video VAE for efficient temporal modeling, while simultaneously reconstructing video frames to provide precise pixel-level grounding for latent representations. Then during co-fine-tuning, HyWorldVLA removes the pixel reconstruction branch, predicts only latent feature, which is used by a flow-matching action expert to generate driving trajectories. Extensive experiments on the NAVSIM v1/v2 benchmark demonstrate that HyWorldVLA significantly outperforms BEV and VLA baselines.
Vision-Language-Action (VLA) models augmented with world modeling represent a promising paradigm for end-to-end autonomous driving. While pixel-level future prediction enables fine-grained spatiotemporal reasoning, it compromises robustness in noisy driving scenarios. Conversely, latent-based world models alleviate this sensitivity but often incur limited interpretability and representational degradation due to absent pixel-level grounding. To reconcile this trade-off, we propose HyWorldVLA, a hybrid world-VLA framework that unifies pixel-level supervision and latent representation learning. In the pre-training stage, HyWorldVLA predicts video latents encoded by a pre-trained video VAE, while simultaneously reconstructing video frames to provide precise pixel-level grounding. During the subsequent co-fine-tuning phase, the model exclusively predicts latent features, which are fed into an action expert to generate trajectories. Extensive experiments on NAVSIM v1 and v2 benchmarks demonstrate that HyWorldVLA significantly outperforms both pixel-based and latent-based world model baselines. Notably, we present the first comprehensive qualitative and quantitative analysis of world model noise robustness in autonomous driving, establishing a new benchmark for evaluating future architectures.
\end{abstract}

\input{chapters/introduction}
\input{chapters/related_work}
\input{chapters/method}
\input{chapters/experiments}
\input{chapters/conclusion}

\bibliography{ref}
\bibliographystyle{iclr2026_conference}

\input{chapters/appendix}

\end{document}

%% file: template/math_commands.tex
\usepackage{amsmath,amsfonts,bm}

\def\eqref#1{equation~\ref{#1}}
\def\1{\bm{1}}

\DeclareMathAlphabet{\mathsfit}{\encodingdefault}{\sfdefault}{m}{sl}
\SetMathAlphabet{\mathsfit}{bold}{\encodingdefault}{\sfdefault}{bx}{n}

%% file: chapters/introduction.tex
\section{Introduction}
\label{intro}

End-to-end autonomous driving has emerged as a dominant research direction for building intelligent vehicle systems, aiming to unify perception, reasoning, and control within a single holistic pipeline. As a core driving-centric paradigm, Vision-Language-Action (VLA) models~\citep{wang2023drivemlm, hwang2024emma, li2025recogdrive, lu2026onevl} integrate visual scene perception, language semantic understanding, and low-level action prediction, effectively leveraging large-scale driving datasets and generalized vision-language priors to learn robust driving behaviors. Compared with traditional modular autonomous driving systems that suffer from cascading errors and limited generalization on long-tailed and interactive traffic scenarios, VLA-based frameworks deliver more adaptive and generalizable decision-making capabilities, showing great potential for real-world vehicle deployment.

\begin{figure}[t]
    \centering
    \vspace{-0.5cm}
    \includegraphics[
        width=\textwidth, 
        trim={0 100 0 100},
        clip
    ]{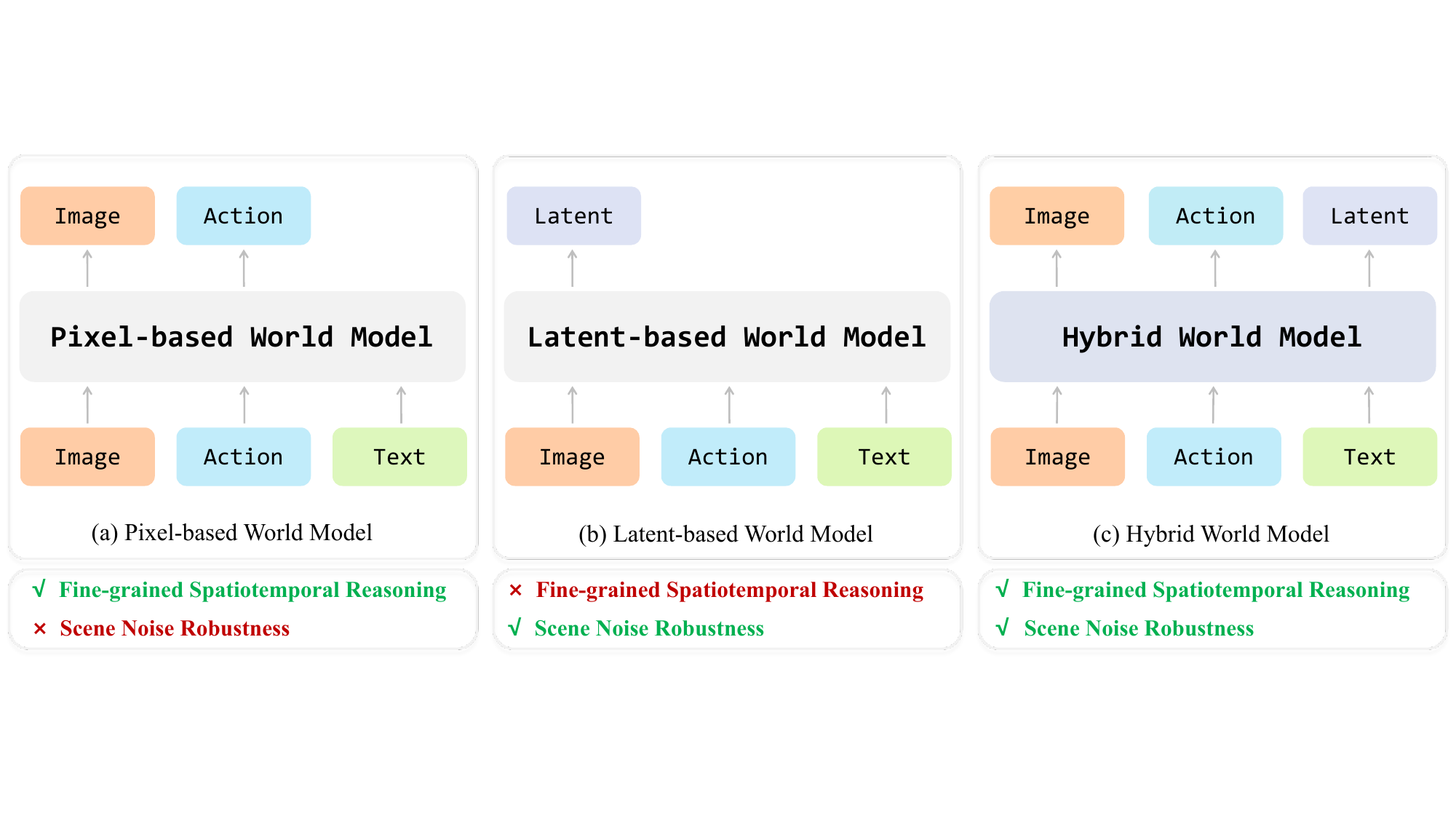} % 图片路径
    \caption{\textbf{Comparison of world model paradigms.} (a) Pixel-based World Model: iteratively predicts actions and images, prone to scene noise; (b) Latent-based World Model: predicts latent states, weak in spatiotemporal dynamics modeling; (c) Hybrid World Model: our method simultaneously models iterative action-image prediction and future latent state prediction.} % 标题
    \label{fig:paradigm}
    \vspace{-0.3cm}
\end{figure}

To further strengthen the spatiotemporal reasoning and scenario generalization of VLA models, recent advances augment standard driving VLA pipelines with world modeling capabilities, which learn to predict future scene states for forward-looking causal reasoning~\citep{jia2023adriver,wang2024drivedreamer,chen2025drivinggpt}. Existing world modeling paradigms for driving VLAs fall into two dominant categories with inherent trade-offs. Pixel-based world models enforce supervision via full future frame reconstruction, providing dense geometric and physical constraints that yield strong generalization on occluded, interactive, and long-tailed driving scenarios, as shown in Figure~\ref{fig:paradigm} (a). However, pixel-level reconstruction is inherently sensitive to environmental variations ranging from rain and fog to fluctuating illumination, which fundamentally constrains the deployment robustness of pixel-based world models under complex driving conditions. In contrast, latent-based world VLAs improve scene noise robustness by forecasting future scene states purely in compact feature space, as shown in Figure~\ref{fig:paradigm} (b). Nevertheless, the absence of explicit pixel-level grounding results in representation degeneration and limited interpretability, which severely deteriorates model performance in complex real-world driving scenarios~\citep{tu2025role}.

To resolve this fundamental dilemma in the world modeling domain, we present HyWorldVLA, a novel hybrid world-VLA framework that unifies the strengths of pixel-grounded supervision and latent feature prediction to deliver both accuracy and robustness for end-to-end autonomous driving. Unlike existing purely pixel-based or latent-based world models, as shown in Figure~\ref{fig:paradigm} (c), HyWorldVLA employs a dual-supervision paradigm in the pre-training stage, jointly optimizing video VAE latent prediction and raw pixel frame reconstruction. During the subsequent co-fine-tuning phase, the model exclusively predicts latent features, which are fed into an action expert to generate driving trajectories. Under this design, pixel-level reconstruction serves as a rigorous structural regularizer during pre-training to refine latent embedding learning and prevent representation collapse. For the co-fine-tuning stage, action generation is conditioned on latent feature modeling that retains compact temporal representations, bolstering the scene noise robustness of the generated driving trajectory. Extensive experimental results demonstrate that HyWorldVLA successfully combines the high performance brought by pixel-based world models’ fine-grained spatiotemporal dynamic modeling with the scene noise robustness intrinsic to latent-based world models. Our contributions are summarized as follows:
\begin{itemize}
    \item We propose HyWorldVLA, which establishes a hybrid world modeling paradigm that unifies the complementary characteristics of world action models and latent world models into a single framework.
    \item We introduce language guidance and latent conditioning mechanisms to effectively steer the model toward semantic information relevant to action generation.
    \item Extensive experiments on the NAVSIM benchmark demonstrate that our method achieves state-of-the-art performance, validating the effectiveness of our hybrid world modeling.
\end{itemize}

%% file: chapters/related_work.tex
\section{Related Work}
\label{related}

\textbf{Vision-Language-Action Models in Driving.}
With remarkable multimodal scene understanding and reasoning capabilities, Vision-Language Models (VLMs) are being extensively integrated into autonomous driving systems, promoting the development of end-to-end Vision-Language-Action (VLA) architectures~\citep{hu2025vision}. Early explorations including DriveMLM~\citep{wang2023drivemlm} and DriveGPT4~\citep{xu2024drivegpt4} directly leverage VLMs' inherent text-generation capabilities to produce textual driving actions, thus fully utilizing the world knowledge from pre-trained VLMs~\citep{hwang2024emma, sima2024drivelm, jiang2025alphadrive}. However, the fundamental gap between discrete language tokens and continuous control spaces introduces precision limits and breaks the end-to-end differentiability. Consequently, recent works augment VLM backbones with mechanisms that produce direct numeric outputs. ReCogDrive~\citep{li2025recogdrive} integrates an autoregressive model with a diffusion planner through a hierarchical cognitive pipeline, while AutoVLA~\citep{zhou2025autovla} unifies semantic reasoning and trajectory planning in a single autoregressive model with dual thinking modes. ORION~\citep{fu2025orion} and OpenDriveVLA~\citep{zhou2025opendrivevla} further bridge the semantic-action gap by aggregating long-term context and performing hierarchical vision-language alignment.

\textbf{World Models for Autonomous Driving.}
World models have been applied to autonomous driving~\citep{zheng2024doe1,hu2023gaia,jia2023adriver,wang2024drivedreamer,li2025end,chen2025drivinggpt,zhang2025epona} due to their ability to anticipate future states and comprehend the temporal dynamics of the world. FSDrive~\citep{zeng2025futuresightdrive} generates unified future frames with explicit physical priors as visual spatiotemporal chain-of-thought. DriveVLA-W0~\citep{li2025drivevla} addresses the ``supervision deficit'' in VLAs by predicting future images as dense self-supervised signals, instantiated via both autoregressive and diffusion world models. OneVL~\citep{lu2026onevl} routes latent reasoning through compact tokens supervised by dual auxiliary decoders, matching answer-only latency. Latent-CoT-Drive~\citep{tan2025latent} represents chain-of-thought reasoning in an action-aligned latent space using action-proposal and world model tokens. Pixel-level prediction is inherently susceptible to scene noise, whereas latent-level prediction struggles to capture the underlying causal relationships of the physical world. To address these complementary limitations, we propose a novel framework that bridges pixel-level and latent-level prediction.

%% file: chapters/method.tex
\section{Methodology}
\label{meth}

\subsection{Problem Definition and Overall Framework}
% The autonomous driving task is formulated as a problem of predicting future T actions $A_{t_t+T}$ given navigation commands, historical ego-vehicle states, and image observations.

\begin{figure}[t]
    \centering
    \vspace{-0.5cm}
    \includegraphics[
        width=\textwidth, 
        trim={10 80 0 60},
        clip
    ]{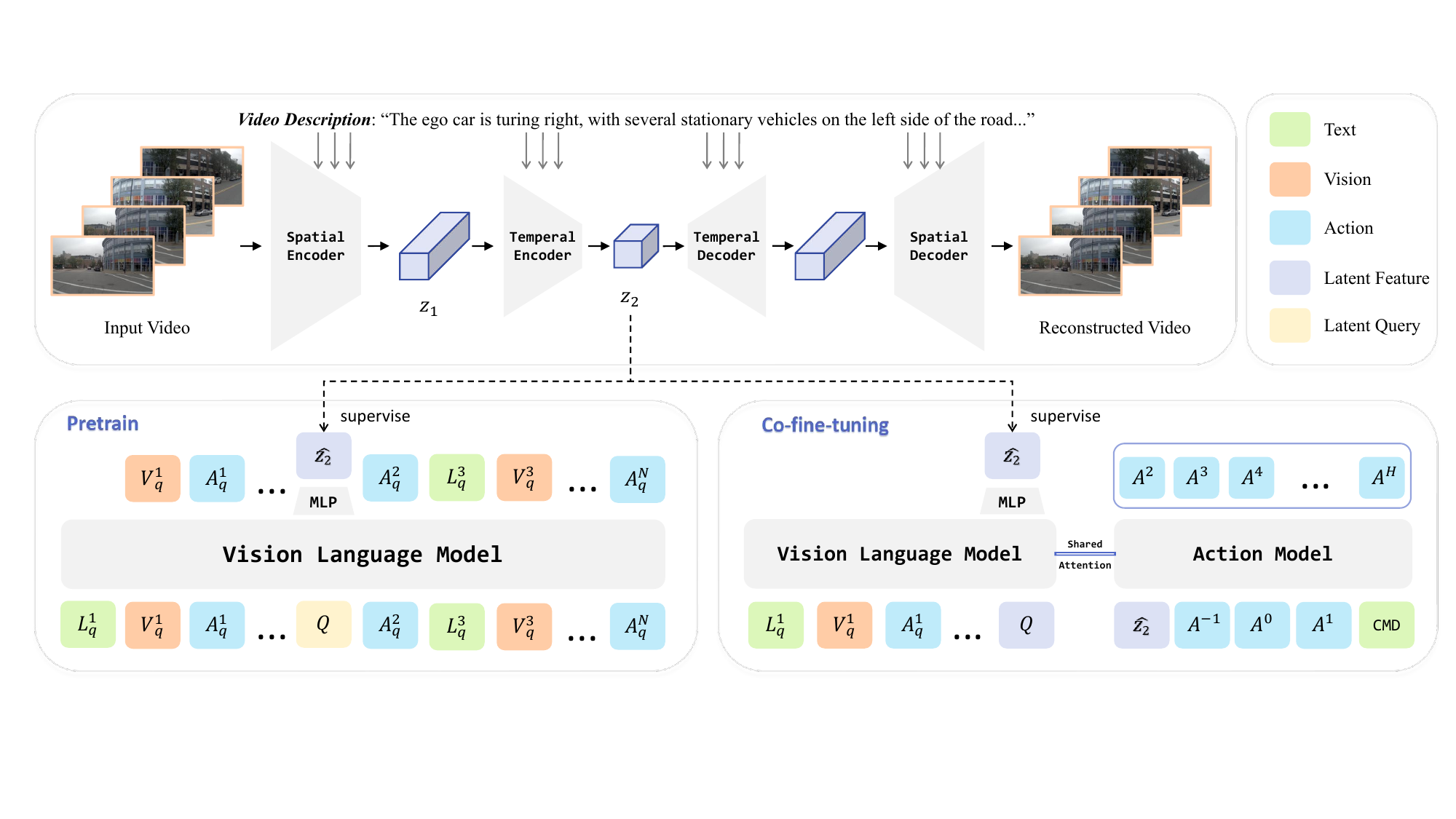} % 图片路径
    \caption{\textbf{Overview of HyWorldVLA.} HyWorldVLA consists of three training stages. First, a video VAE model is trained under text guidance to compress and reconstruct video frames, thereby providing ground-truth targets for subsequent training. In the pre-training stage, sequences composed of text, images, and actions are used to train the world model for iterative prediction, where a learnable latent query $Q$ is introduced to aggregate information and predict future latents. In the co-fine-tuning stage, the historical sequences together with $Q$ are fed into the VLM to obtain global context and predicted latents, which are then passed through a joint attention mechanism to the action expert for action generation.} % 标题
    \label{fig:framework}
    \vspace{-0.3cm}
\end{figure}

HyWorldVLA is formulated to predict future ego-vehicle waypoints conditioned on multi-modal observations. Specifically, at each time step $t$, the model receives a sequence of front-view camera images $\mathbf{V}_{t-H:t} = \left \{ v_{t-H}, \dots, v_t \right \}$, including the current and $H$ preceding frames, and historical vehicle waypoints $\mathbf{W}_{t-H:t} = \left \{ w_{t-H}, \dots, w_t \right \}$, where $w_t = \left ( x_t, y_t, \theta_t \right )$ represents the position and heading angle at step $t$. Additionally, historical navigation commands $\mathbf{L}_{t-H:t}$ (e.g., \textit{Go Straight} and \textit{Turn Right}) are used to specify intended directions explicitly. Then based on these inputs, the model outputs a sequence of future waypoints $\mathbf{W}_{t+1:t+T}$ over a horizon of $T$ time steps.

% As shown in Fig.2, the overall framework of HyWorldVLA consists of three main components and three training stages. \textit{1) Video Encoder:} A sequence of video frames is encoded into a latent feature, providing a compact representation of future world dynamics for VLM supervision. \textit{2) VLM Backbone:} world model. \textit{3) Action Expert:} A flow‑matching‑based action expert, via a joint‑attention mechanism, conditions on the hidden states from each layer of the VLM backbone, the predicted future visual latents, and the ego‑vehicle’s historical motion, to decode trajectory predictions.

As shown in Figure~\ref{fig:framework}, the overall framework of HyWorldVLA consists of three main components and three training stages. \textit{(1) Video Encoder:} Sec.~\ref{sec:latent} elaborates on a video encoder that compresses a sequence of video frames into a latent feature, providing a compact representation of future world dynamics; \textit{(2) VLM Backbone:} Then the latent feature and video frames are jointly predicted by a VLM backbone, which is presented in Sec.~\ref{sec:pre-training}; \textit{(3) Action Expert:} Finally, in Sec.~\ref{sec:co-fine-tuning} we introduce the co-fine-tuning of the VLM model and a query-based action model, which predicts trajectories conditioning on the hidden states from each layer of the VLM backbone, the predicted future visual latents, and the ego‑vehicle’s historical motion.

% During pre-training, we jointly optimize a dense world model and a latent world model built upon a pre‑trained VLM backbone. The dense world model is formulated via next‑token prediction over a discrete sequence composed of text, action, and visual tokens. Each action token corresponds to a fixed‑length action chunk, which is discretized and encoded into a token using the FAST algorithm. Visual tokens are sparsely sampled as frames at uniform temporal intervals. The latent world model leverages a learnable query to aggregate instructional prompts and observational cues, predicting a compact latent representation of future world dynamics; we employ a video compression model, VideoVAE, to supply ground‑truth latent representations for supervision. Additionally, we use a large vision‑language model to annotate decision‑making text, and inject these annotations both after the second image token and into the VideoVAE, thereby steering the model toward predictive emphasis on crucial future information.

% In the post‑training stage, we retain only the historical text, action, and visual token sequences. A flow‑matching‑based action expert, via a joint‑attention mechanism, conditions on the hidden states from each layer of the VLM backbone, the predicted future visual latents, and the ego‑vehicle’s historical motion, to decode action predictions for a future horizon of T time steps.

\subsection{Text-guided Latent Feature Learning}
\label{sec:latent}

% To encode future frames into compact latent space for VLM supervision, we adopt a video variational autoencoder. As shown in Fig.2, its architecture integrates temporal-aware spatial compression and lightweight temporal motion compression, with additional text cross-attention branches for cross-modal semantic alignment. Given a video segment $\mathbf{V} \in \mathbb{R}^{C \times T \times H \times W}$, 

% To encode future frames into compact latent space for VLM supervision, we adopt a video variational autoencoder. As shown in Fig.2, its architecture integrates temporal-aware spatial compression and lightweight temporal motion compression, with additional text cross-attention branches for cross-modal semantic alignment. Given a video segment $\mathbf{V} \in \mathbb{R}^{C \times T \times H \times W}$, 

To encode future frames into compact latent features for VLM supervision, we adopt a video variational autoencoder~\citep{xing2024large}. As shown in Figure~\ref{fig:framework}, taking a video segment $\mathbf{v} \in \mathbb{R}^{C \times T \times H \times W}$ as input, the model first employs a spatial encoder to produce spatial latent $\mathbf{z}_1 \in \mathbb{R}^{c \times T \times \frac{H}{8} \times \frac{W}{8}}$. Built upon a Stable Diffusion image VAE~\citep{rombach2022high}, this spatial encoder stacks 3D convolutions to perform $8 \times$ spatial downsampling while retaining full temporal length. Afterwards, a lightweight temporal autoencoder takes $\mathbf{z}_1$ as input and stacks 3D ResNet blocks to implement $4 \times$ temporal compression, generating latent $\mathbf{z}_2 \in \mathbb{R}^{c' \times \frac{T}{4} \times \frac{H}{8} \times \frac{W}{8}}$ that eliminates redundant inter-frame static information. Then during reconstruction, the full-resolution video $\hat{\mathbf{v}}$ is recovered through a symmetric two-stage decoding pipeline. 

To further enhance reconstruction quality, we embed multi-layer text cross-attention inside each block to introduce textual guidance during feature encoding. We split visual features into patch tokens, take visual patches as query and value, and adopt Flan-T5~\citep{chung2024scaling} text embeddings as keys to compute cross-attention. The attention output is added back to original visual features via residual connection. This textual semantic prior effectively suppresses motion ghosting, edge blurring and temporal flickering, greatly improving reconstruction fidelity for complex traffic scenes.

The VAE is optimized via a multi-component loss that balances reconstruction loss $\mathcal{L}_\mathrm{rec}$, adversarial loss $\mathcal{L}_\mathrm{GAN}$, and KL-divergence regularization loss $\mathcal{L}_\mathrm{KL}$:
\begin{equation}
\mathcal{L}_\mathrm{vae} = \mathcal{L}_\mathrm{rec} + \lambda_\mathrm{GAN} \mathcal{L}_\mathrm{GAN} + \lambda_\mathrm{KL} \mathcal{L}_\mathrm{KL}.
\end{equation}
Through hierarchical spatial-temporal compression and text cross-modal injection, the video VAE generates a compact latent representation, providing effective supervision for downstream VLA pre-training and co-fine-tuning.

\subsection{Pre-training}
\label{sec:pre-training}

The pre-training stage aims to equip the backbone with the ability to predict the world evolution both on the image-level and latent-level.

\textbf{Vision Modeling.}
We follow VQGAN~\citep{esser2021taming} to transform the continuous video stream into discrete visual tokens. Specifically, given a continuous video sequence $\mathbf{V}_{t-H:t+T}$, we first partition the sequence into $N$ non-overlapping chunks, where $N = (H+T) / t_{\text{chunk}}+1$ and $t_{\text{chunk}}$ denotes the chunk duration. For each chunk, we employ a VQ tokenizer to discretize the first visual observation into discrete visual tokens, yielding:
\begin{equation}
V_q = \{V_q^1, V_q^2, \ldots, V_q^N\},
\end{equation}
where $V_q^j$ represents the quantized visual tokens corresponding to the initial frame of the $j$-th chunk.

\textbf{Action Modeling.}
We adopt FAST~\citep{pertsch2025fast} to discretize continuous action sequences into discrete action tokens. Given the waypoint sequence $\mathbf{W}_{t-H:t+T}$, we convert absolute coordinates to ego-relative coordinates to obtain the action sequence ${V}_{t-H:t+T}$, which is split into chunks and encodes ${V}_{t-H:t+T}$ into discrete tokens using the FAST action tokenizer, yielding:
\begin{equation}
{A_q} = \{A_q^1, A_q^2, \ldots, A_q^N\},
\end{equation}
where each $A_q^j$ corresponds to the action tokens within the $j$-th temporal chunk.

\textbf{Language Modeling.}
We utilize the Emu3 tokenizer~\citep{wang2024emu3} to discretize the driving command $\mathbf{CMD}_{t-H:t+T}$ into language tokens:
\begin{equation}
{L_q} = \{L_q^1, L_q^2, \ldots, L_q^N\},
\end{equation}
where each $L_q^j$ corresponds to the $j$-th language token.

\textbf{Future Latent Prediction.}
We strategically insert a learnable query token $Q$ into the input token sequence, and the hidden state at the query position is fed to an MLP to predict the future latent representation $\mathbf{\hat{z}}_2$ during the forward pass.

\textbf{Training Objective.}
Since all modality signals are transformed into discrete tokens drawn from a shared vocabulary, we construct the world model's input sequence as:
\begin{equation}
\mathcal{S} = [L_q^1, V_q^1, A_q^1, L_q^2, V_q^2, Q, A_q^2, L_q^3, \ldots, V_q^N, A_q^N],
\end{equation}
To prevent information leakage, causal masking is applied such that each token can only attend to preceding tokens in the sequence. The final training objective consists of three terms:
\begin{equation}
\mathcal{L}_{\text{world}} = \sum_{j=1}^{N} \text{CE}(\hat{L}_q^j, L_q^j) + \sum_{j=1}^{N} \text{CE}(\hat{A}_q^j, A_q^j) +  \lambda_1\sum_{j=1}^{N} \text{CE}(\hat{v}_q^j, v_q^j) +  \lambda_2\|\mathbf{\hat{z}}_2 - \mathbf{z}_2\|_2^2 .
\end{equation}
Here, the first term is the language tokens prediction loss, the second term is the action prediction loss, the third term is the visual tokens prediction loss, and the last term enforces consistency between the predicted latent representation and the ground-truth compressed feature extracted from the future video sequence. $\lambda_1$ and $\lambda_2$ are the vision token and latent representation loss weights.

\subsection{Co-Fine-Tuning}
\label{sec:co-fine-tuning}
In the co‑fine‑tuning phase, the world model and the action expert are jointly optimized to achieve continuous and reliable action prediction.

% We encode the historical navigation commands, video sequences, and action sequences into text, visual, and action tokens, respectively, and arrange them in the order used in pre-training, then append the latent query Q at the end to form the final input sequence. Feeding this sequence to the VLM yields global context, while the hidden state at Q is decoded via an MLP head to produce a prediction of future latent. Both the global context and the predicted future latent are then provided to the action expert to enable accurate action generation.

% We adopt a joint attention mechanism ~\citep{black2024pi0} for deep information fusion and design two types of action experts: a flow-matching-based expert and a query-based expert. In the flow-matching-based action expert, the historical actions, navigation commands, noise tokens augmented with timestep information, and the predicted latent from the VLM are concatenated as the sequence input to the attention module. After interacting with the VLM context, the hidden state at the noise token positions is decoded via an MLP head to produce the conditional vector ~\citep{lipman2022flow}. 

We integrate the backbone and the action expert via a joint attention mechanism~\citep{black2024pi0}. The historical actions ${V}_{t-H:t}$, the driving commands $\mathbf{CMD}_{t-H:t}$ and the predicted latent representation $\mathbf{\hat{z}}_2$ from the backbone are concatenated as the sequence input to the attention module. After interacting with the backbone context, the hidden state at the action token positions is decoded via an MLP head.

To verify cross-paradigm compatibility, we instantiate our method on the two dominant action model families: selection-based~\citep{li2024hydra} and generative-model-based~\citep{lipman2022flow}.

For the generative-model training objectives, including flow matching, we have the following:
\begin{equation}
\mathcal{L}_{\text{cotrain}} = |\hat{v} - v|_2^2 + \lambda_3|\hat{z} - z|_2^2
\end{equation}
The first term is the flow-matching vector field regression loss, and the second term is the latent prediction loss, and \(\lambda_3\) is the latent loss weight.

% For the query-based action expert, we follow the approach in HydraMDP and employ learnable queries to extract decision-relevant information from both the global context and the predicted latent. The extracted features are then decoded to produce a score for each trajectory in an offline trajectory vocabulary. A ground-truth score is computed by comprehensively considering the expert trajectory as well as metrics such as safety and compliance with driving regions. A KL divergence loss is employed to supervise the predicted score distribution. 

The selection-based training objectives are as follows:
% Total loss
\begin{equation}
\mathcal{L}_{\text{cotrain}} = \mathrm{KL}\left( p_{\text{pred}} \;\middle\|\; p_{\text{gt}} \right) + \lambda_3\,\|\hat{z} - z\|_2^2,
\end{equation}
where $p_\text{pred}$ is the predicted score for each trajectory in an offline trajectory vocabulary, $p_\text{gt}$ is the ground-truth score computed by comprehensively considering the expert trajectory as well as the metrics of the evaluation and a KL divergence loss is employed to supervise the predicted score distribution.

%% file: chapters/experiments.tex
\section{Experiments}
\label{exp}

\subsection{Datasets}
We evaluate our approach on the NAVSIM v1~\citep{dauner2024navsim} and v2~\citep{cao2025pseudo} benchmarks, providing closed-loop evaluation protocols and multi-dimensional quantitative metrics.

\textbf{NAVSIM v1.}
NAVSIM v1 measures driving capability from five core dimensions: No at-fault Collision (NC), Drivable Area Compliance (DAC), Time-To-Collision (TTC), Comfort (C.), and Ego Progress (EP). The overall performance is aggregated into the \textit{Predictive Driver Model Score (PDMS)} via a weighted multiplicative formulation:
\begin{equation}
    \text{PDMS} = \text{NC} \times \text{DAC} \times \frac{5 \times \text{EP} + 5 \times \text{TTC} + 2 \times \text{C.}}{12}.
\end{equation}

\textbf{NAVSIM v2.}
As an extended and refined version, NAVSIM v2~\citep{cao2025pseudo} introduces more fine-grained evaluation dimensions to better align with real-world driving requirements. In addition to the core metrics inherited from v1, it adds Driving Direction Compliance (DDC), Traffic Light Compliance (TLC), Lane Keeping (LK), History Comfort (HC), and Extended Comfort (EC). The corresponding comprehensive metric is the \textit{Extended Predictive Driver Model Score (EPDMS)}, defined as:
\begin{equation}
    \text{EPDMS} = \text{NC} \times \text{DAC} \times \text{DDC} \times \text{TLC} \times \frac{5 \times \text{EP} + 5 \times \text{TTC} + 2 \times \text{LK} + 2 \times \text{HC} + 2 \times \text{EC}}{16}.
\end{equation}

\subsection{Implementation Details}
Our model backbone is based on Emu3~\citep{wang2024emu3}, a unified multimodal autoregressive model.

\textbf{VideoVAEPlus Fine-tuning.}
We initialize from the 4-channel VideoVAEPlus~\citep{xing2024large} pre-trained model and fine-tune it on the NuPlan dataset~\citep{caesar2021nuplan}. For each video segment, we uniformly sample 8 frames resized to $216 \times 216$ and use Qwen3.6-plus to generate a short scene description. The training is performed on 4 PPUs~\citep{ali_ppu_2026} with a per-PPU batch size of 1. The weights of the adversarial loss and the KL loss are set to 0.5 and 1e-6 respectively, and the adversarial loss is enabled after 50K steps. We train the model for 100K steps, enabling it to learn compact representations from the videos.

\textbf{Pre-training.}
During the pre-training stage, we utilize the OpenScenes dataset~\citep{openscene2023} comprising over 120 hours of driving video recordings which are segmented into clips of 20.0 s. During training, for each clip, we randomly sample a frame sequence with a future horizon $T=4.0$ s and a history horizon $H=1.0$ s. Given the chunk duration $t_{\text{chunk}}=1.0$ s, the number of non-overlapping chunks num $N$ is 6. The training configuration adopts a batch size of 4 across two nodes, each equipped with 16 PPUs~\citep{ali_ppu_2026}. We set the learning rate to $2.2 \times 10^{-4}$ with a maximum of 4000 training steps. The loss weighting coefficients are configured as $\lambda_1 = 0.5$ and $\lambda_2 = 0.1$.

\textbf{Co-fine-tuning.}
During the co-fine-tuning phase, we fine-tune the pre-trained model on the NAVSIM trainval dataset~\citep{dauner2024navsim} which comprises over 100,000 frames with ground-truth ego-vehicle trajectory annotations. Consistent with the pre-training configuration, the future horizon $T$ and historical horizon $H$ are set to 4 s and 1 s, respectively. We train for 4000 steps with a total batch size of 96. The learning rate is set to $5\times10^{-5}$, with a latent prediction loss weight $\lambda_3 = 0.1$.

\subsection{Comparison with State-of-the-Art Methods}
% We compare HyWorldVLA with several representative approaches, including end-to-end methods, vision-language-action (VLA) models, world models, and latent world models. As shown in Table\ref{tab:navsim-v1} and \ref{tab:navsim-v2}, our method achieves state-of-the-art (SOTA) performance on both Navsim V1 and V2 benchmarks, significantly surpassing competitive baselines such as DriveVLA-W0 and LatentWAM. This superior performance benefits from our hybrid world modeling design, which effectively mitigates the limitations of single-paradigm modeling and substantially enhances the capabilities of the VLA.
We benchmark HyWorldVLA against representative pixel-based world model (WM) methods, latent-based world model methods, as well as several canonical end-to-end and VLA-based baselines. As shown in Table \ref{tab:navsim-v1} and Table \ref{tab:navsim-v2}, our approach achieves state-of-the-art (SOTA) performance on both v1 and v2 benchmarks, significantly outperforming the competitive pixel-based WM baselines such as DriveVLA-W0~\citep{li2025drivevla} and the latent-based WM baselines including Latent-WAM~\citep{wang2026latent}. This performance gain can be ascribed to our hybrid world modeling paradigm, which reconciles the fine-grained spatiotemporal modeling capability of pixel-based world models with the intrinsic high-level semantic representation capacity of latent world models.

\begin{table}[t]
\caption{Comparison with state-of-the-art methods on the NAVSIM v1. 
Abbreviations: 1x Cam (single front-view camera), Nx Cam (surround-view cameras), L (LiDAR).}
\label{tab:navsim-v1}
\centering
% 缩小字体（可选，根据需求调整）
\small 
\setlength{\tabcolsep}{3pt} % 【新增】减少列间距离
\renewcommand{\arraystretch}{0.9} % 【新增】压缩行高，使表格更紧凑
% 定义表格：X列（方法名，自动换行）+ 8个固定宽度的c列
\begin{tabularx}{\linewidth}{>{
    \raggedright\arraybackslash}p{4.5cm} | c | c | c c | c c c | >{\centering\arraybackslash}p{0.95cm}
}
\toprule
\textbf{Method} & \textbf{Ref} & \textbf{Sensors} & \textbf{NC\(\uparrow\)} & \textbf{DAC\(\uparrow\)} & \textbf{TTC\(\uparrow\)} & \textbf{C.\(\uparrow\)} & \textbf{EP\(\uparrow\)} & \textbf{PDMS\(\uparrow\)} \\
\midrule
Human & - & - & 100 & 100 & 100 & 99.9 & 87.5 & \cellcolor{blue!8} 94.8 \\
\midrule
\multicolumn{9}{@{}l@{}}{\hspace{\tabcolsep}\thead[l]{\textit{End-to-end-based Methods}}} \\
UniAD~\citep{hu2023planning} & CVPR'23 & 6x Cam & 97.8 & 91.9 & 92.9 & \textbf{100.0} & 78.8 & \cellcolor{blue!8} 83.4 \\
TransFuser~\citep{prakash2021multi} & TPAMI'23 & 3x Cam + L & 97.7 & 92.8 & 92.8 & \textbf{100.0} & 79.2 & \cellcolor{blue!8} 84.0 \\
PARA-Drive~\citep{weng2024drive} & CVPR'24 & 6x Cam & 97.9 & 92.4 & 93.0 & 99.8 & 79.3 & \cellcolor{blue!8} 84.0 \\
Hydra-MDP~\citep{li2024hydra} & arXiv'24 & 3x Cam + L & 98.3 & 96.0 & 94.6 & \textbf{100.0} & 78.7 & \cellcolor{blue!8} 86.5 \\
DiffusionDrive~\citep{liao2025diffusiondrive} & CVPR'25 & 3x Cam + L & 98.2 & 96.2 & 94.7 & \textbf{100.0} & 82.2 & \cellcolor{blue!8} 88.1 \\
\midrule
\multicolumn{9}{@{}l@{}}{\hspace{\tabcolsep}\thead[l]{\textit{VLA-based Methods}}} \\
AutoVLA~\citep{zhou2025autovla} & NeurIPS'25 & 3x Cam & 98.4 & 95.6 & \textbf{98.0} & 99.9 & 81.9 & \cellcolor{blue!8} 89.1 \\
ReCogDrive-8B~\citep{li2025recogdrive} & ICLR'26 & 3x Cam & 97.8 & 97.7 & 94.9 & \textbf{100.0} & 86.3 & \cellcolor{blue!8} 90.5 \\
\midrule
\multicolumn{9}{@{}l@{}}{\hspace{\tabcolsep}\thead[l]{\textit{Pixel-based Word Model Methods}}} \\
WoTE~\citep{li2025end} & ICCV'25 & 3x Cam + L & 98.5 & 96.8 & 94.4 & 99.9 & 81.9 & \cellcolor{blue!8} 88.3 \\
FSDrive~\citep{zeng2026futuresightdrive} & NeurIPS'25 & 1x Cam & 98.2 & 93.8 & 93.3 & 99.9 & 80.1 & \cellcolor{blue!8} 85.1 \\
ReSim~\citep{yang2026resim} & NeurIPS'25 & 1x Cam & - & - & - & - & - & \cellcolor{blue!8} 86.6 \\
PWM~\citep{zhao2026forecasting} & NeurIPS'25 & 1x Cam & 98.6 & 95.9 & 95.4 & \textbf{100.0} & 81.8 & \cellcolor{blue!8} 88.1 \\
CoWorld-VLA~\citep{huang2026coworld} & arXiv'26 & 1x Cam & \textbf{99.2} & 96.8 & 96.6 & \textbf{100.0} & 83.6 & \cellcolor{blue!8} 89.8 \\
DriveLaW~\citep{xia2026drivelaw} & CVPR'26 & 1x Cam & 99.0 & 97.1 & 96.7 & \textbf{100.0} & 81.3 & \cellcolor{blue!8} 89.1 \\
DriveVLA-W0~\citep{li2025drivevla} & ICLR'26 & 1x Cam & 98.7 & \textbf{99.1} & 95.3 & 99.3 & 83.3 & \cellcolor{blue!8} 90.2 \\
\midrule
\multicolumn{9}{@{}l@{}}{\hspace{\tabcolsep}\thead[l]{\textit{Latent-based World Model Methods}}} \\
LAW~\citep{li2025enhancing} & ICLR'25 & 1x Cam & 96.4 & 95.4 & 88.7 & 99.9 & 81.7 & \cellcolor{blue!8} 84.6 \\
World4Drive~\citep{zheng2025world4drive} & ICCV'25 & 1x Cam & 97.4 & 94.3 & 92.8 & \textbf{100.0} & 79.9 & \cellcolor{blue!8} 85.1 \\
Epona~\citep{zhang2025epona} & ICCV'25 & 1x Cam & 97.9 & 95.1 & 93.8 & 99.9 & 80.4 & \cellcolor{blue!8} 86.2 \\
DreamerAD~\citep{yang2026dreamerad} & arXiv'26 & 1x Cam & 98.0 & 97.2 & 94.3 & \textbf{100.0} & 83.1 & \cellcolor{blue!8} 88.7 \\
OneVL~\citep{lu2026xiaomi} & arXiv'26 & 1x Cam & - & - & - & - & - & \cellcolor{blue!8} 88.8 \\
WorldRFT~\citep{yang2026worldrft} & AAAI'26 & 1x Cam & 97.8 & 96.8 & 94.0 & \textbf{100.0} & 81.7 & \cellcolor{blue!8} 87.8 \\
ResWorld~\citep{zhang2026resworld} & ICLR'26 & 4x Cam + L & 98.9 & 96.5 & 95.6 & \textbf{100.0} & 83.1 & \cellcolor{blue!8} 89.0 \\
\midrule
Ours & - & 1x Cam & 98.6 & 98.9 & 95.1 & 99.3 & \textbf{84.6} & \cellcolor{blue!8} \textbf{90.59} \\
\bottomrule
\end{tabularx}
\end{table}

\begin{table}[t]
\caption{Comparison with state-of-the-art methods on the NAVSIM v2 with extended metrics.}
\label{tab:navsim-v2}
\centering
\fontsize{8pt}{9pt}\selectfont   % 字号8pt，行距9pt
\setlength{\tabcolsep}{4pt} % 调整列间距离以适应更多列
\renewcommand{\arraystretch}{0.9} % 压缩行高
% 定义表格：X列（方法名）+ 11个固定宽度的c列
\begin{tabularx}{\linewidth}{
    >{\raggedright\arraybackslash}X | c c c c | c c c c c | c
}
\toprule
\textbf{Method} & \textbf{NC$\uparrow$} & \textbf{DAC$\uparrow$} & \textbf{DDC$\uparrow$} & \textbf{TLC$\uparrow$} & \textbf{EP$\uparrow$} & \textbf{TTC$\uparrow$} & \textbf{LK$\uparrow$} & \textbf{HC$\uparrow$} & \textbf{EC$\uparrow$} & \textbf{EPDMS$\uparrow$} \\
\midrule
\multicolumn{9}{@{}l@{}}{\hspace{\tabcolsep}\thead[l]{\textit{End-to-end-based Methods}}} \\
TransFuser~\citep{prakash2021multi} & 96.9 & 89.9 & 97.8 & 99.7 & 87.1 & 95.4 & 92.4 & 98.3 & 87.2 & \cellcolor{blue!8} 76.7 \\
HydraMDP++~\citep{li2025hydra} & 97.2 & 97.5 & 99.4 & 99.6 & 83.1 & 96.5 & 94.4 & 98.2 & 70.9 & \cellcolor{blue!8} 81.4 \\
DriveSuprim~\citep{yao2026drivesuprim} & 97.5 & 96.5 & 99.4 & 99.6 & \textbf{88.4} & 96.6 & 95.5 & 98.3 & 77.0 & \cellcolor{blue!8} 83.1 \\
ARTEMIS~\citep{feng2025artemis} & 98.3 & 95.1 & 98.6 & 99.8 & 81.5 & 97.4 & 96.5 & 98.3 & - & \cellcolor{blue!8} 83.1 \\
DiffusionDrive~\citep{liao2025diffusiondrive} & 98.2 & 95.9 & 99.4 & 99.8 & 87.5 & 97.3 & 96.8 & 98.3 & \textbf{87.7} & \cellcolor{blue!8} 84.5 \\
\midrule
\multicolumn{9}{@{}l@{}}{\hspace{\tabcolsep}\thead[l]{\textit{VLA-based Methods}}} \\
ReCogDrive~\citep{li2025recogdrive} & 98.3 & 95.2 & 98.3 & 99.8 & 87.1 & 97.5 & 96.6 & \textbf{99.5} & 86.5 & \cellcolor{blue!8} 83.6 \\
\midrule
\multicolumn{9}{@{}l@{}}{\hspace{\tabcolsep}\thead[l]{\textit{Pixel-based Word Model Methods}}} \\
DriveVLA-W0~\citep{li2025drivevla} & 98.5 & \textbf{99.1}  & 98.0 & 99.7 & 86.4 & 98.1 & 93.2 & 97.9 & 58.9 & \cellcolor{blue!8} 86.1 \\
ExploreVLA~\citep{sheng2026explorevla} & \textbf{98.8} & 96.2 & \textbf{99.6} & 99.8 & 87.1 & 98.2 & \textbf{97.8} & 98.3 & 86.8 & \cellcolor{blue!8} 88.8 \\
\midrule
\multicolumn{9}{@{}l@{}}{\hspace{\tabcolsep}\thead[l]{\textit{Latent-based World Model Methods}}} \\
DriveWorld-VLA~\citep{liu2026driveworld} & 98.6 & \textbf{99.1} & \textbf{99.6} & 99.8 & 87.4 & 97.9 & 97.0 & 97.9 & 78.6 & \cellcolor{blue!8} 86.8 \\
DreamerAD~\citep{yang2026dreamerad} & 98.6 & \textbf{99.1} & \textbf{99.6} & 99.8 & 87.4 & 97.9 & 97.0 & 97.9 & 78.6 & \cellcolor{blue!8} 86.8 \\
Latent-WAM~\citep{wang2026latent} & 98.0 & 97.2 & 99.5 & 99.8 & 87.8 & 97.4 & 97.5 & 98.3 & 72.4 & \cellcolor{blue!8} 87.7 \\
\midrule
Ours & \textbf{98.8} & 98.2 & 99.5 & \textbf{99.9} & 87.8 & 98.1 & 96.2 & 98.3 & 77.4 & \cellcolor{blue!8} \textbf{89.71} \\
\bottomrule
\end{tabularx}
\end{table}

\subsection{Ablation Analysis}

\textbf{Ablation on Components.}
Table~\ref{tab:ablation_on_module} presents the ablation results for each component of HyWorldVLA. Removing the pixel-regression ability (Pure LWM) yields the lowest score (87.50), confirming the crucial role of its spatiotemporal modeling capability in understanding the environment and generating accurate actions. Removing the latent representational capacity (Pure WAM) also leads to a significant drop (89.91), indicating its positive effect on learning high-level semantics. Ablating the language guidance from the latent module decreases the score to 90.35, demonstrating that language effectively guides the model toward planning-relevant semantics. Removing the latent conditioning in the action expert model drops the score to 90.29, suggesting that explicitly leveraging predicted future information benefits more accurate action generation. Eliminating the latent supervision during co-fine-tuning results in a score of 90.17, highlighting the importance of preserving pre-trained latent semantics and preventing over-optimization.

\textbf{Ablation on Loss Weights in Pre-training.}
As shown in Table \ref{tab:ablation_on_pretrain_loss_weight}, We first fix $\lambda_2 = 0.1$ and examine the effect of $\lambda_1$. When $\lambda_1$ increases from 0.1 to 0.5 and 1.0, the score improves from 90.48 to 90.59 and drops significantly to 89.83. This indicates that the weight for visual tokens prediction should not be too high. When we fix $\lambda_1 = 0.5$ and increase $\lambda_2$ from 0.1 to 0.2, 0.5 and 1.0, the score decreases from 90.59 to 90.47 and then continues to drop to 90.34 and 90.16, respectively. This demonstrates that simultaneously introducing visual tokens prediction ($\lambda_1$ = 0.5) and latent motion ($\lambda_2$ = 0.1) during the pre-training phase most effectively guides the trajectory.

\begin{table}[t]
    \caption{Ablation study on the NAVSIM benchmark.}
    \label{tab:ablation_on_module}
    \centering
    % \fontsize{8pt}{9pt}\selectfont
    % \setlength{\tabcolsep}{3pt}
    \renewcommand{\arraystretch}{0.85}
    % 关键：@{\extracolsep{\fill}} 自动填充空白
    \begin{tabular*}{0.95\linewidth}{
      @{\extracolsep{\fill}}
      l |
      c c c c c | c 
      @{}
    }
    \toprule
    \textbf{Config} & \textbf{NC$\uparrow$} & \textbf{DAC$\uparrow$} & \textbf{TTC$\uparrow$} &
    \textbf{C.$\uparrow$} & \textbf{EP$\uparrow$} & \textbf{PDMS$\uparrow$} \\
    \midrule
    Pure WAM & 98.3 & 98.6 & 94.5 & 99.6 & 84.0 & 89.91 \\
    Pure LWM & 97.2 & 97.8 & 92.0 & 98.9 & 82.4 & 87.50 \\
    w/o language guidance in latent & 98.5 & 98.8 & 94.7 & 99.6 & 84.5 & 90.35 \\
    w/o latent condition in action expert & \textbf{98.6} & 98.7 & 94.9 & 99.3 & 84.3 & 90.29 \\
    w/o latent supervision in co-fine-tuning & \textbf{98.6} & \textbf{98.9} & \textbf{95.3} & \textbf{99.8} & 83.3 & 90.17 \\
    \midrule
    Ours & \textbf{98.6} & \textbf{98.9} & 95.1 & 99.3 & \textbf{84.6} & \textbf{90.59} \\
    \bottomrule
    \end{tabular*}
    \end{table}

\textbf{Ablation on Loss Weights in co-fine-tuning.}
As shown in Table \ref{tab:ablation_on_cotraining_loss_weight}, starting from $\lambda_3 = 0.05$, we observe that increasing it to 0.1 raises the score from 90.47 to 90.59, indicating that appropriate supervision prevents performance degradation caused by semantic collapse of the latent space. However, when $\lambda_3$ is further increased to 0.2 and 1.0, the score drops to 90.01 and 89.75, respectively, suggesting that overly strong supervision hinders the model from autonomously learning the latent semantics most relevant to action generation.

\begin{table}[htbp]
\centering
% ===== 左侧表格：Pretrain loss weight =====
\begin{minipage}[b]{0.48\textwidth}
\centering
\caption{Ablation study of the loss weights in pre-training stage.}
\label{tab:ablation_on_pretrain_loss_weight}
\small
\setlength{\tabcolsep}{2pt}
\renewcommand{\arraystretch}{0.85}
\begin{tabularx}{\linewidth}{
  >{\centering\arraybackslash}p{0.5cm} 
  >{\centering\arraybackslash}p{0.5cm}  |
  >{\centering\arraybackslash}X
  >{\centering\arraybackslash}X
  >{\centering\arraybackslash}X
  >{\centering\arraybackslash}X
  >{\centering\arraybackslash}X |
  >{\centering\arraybackslash}p{1.0cm}
}
\toprule
$\lambda_1$ & $\lambda_2$ & \textbf{NC$\uparrow$} & \textbf{DAC$\uparrow$} & \textbf{TTC$\uparrow$} & \textbf{C.$\uparrow$} & \textbf{EP$\uparrow$} & \textbf{PDMS$\uparrow$} \\
\midrule
0.1 & 0.1  & 98.6 & 98.7 & 94.9 & 99.6 & 84.6 & 90.48 \\
0.5 & 0.1 & 98.6 & 98.9 & 95.1 & 99.3 & 84.6 & \textbf{90.59} \\
1.0 & 0.1  & 98.5 & 98.5 & 94.9 & 99.6 & 83.5 & 89.83 \\
0.5 & 0.2 & 98.6 & 98.9 & 95.0 & 99.5 & 84.4 & 90.47 \\
0.5 & 0.5 & 98.6 & 98.6 & 94.5 & 99.4 & 84.9 & 90.34 \\
0.5 & 1.0 & 98.6 & 98.6 & 94.7 & 99.5 & 84.4 & 90.16 \\
\bottomrule
\end{tabularx}
\end{minipage}%
\hfill
% ===== 右侧表格：Co-fine-tuning loss weight =====
\begin{minipage}[b]{0.48\textwidth}
\centering
\caption{Ablation study of the loss weights in co-fine-tuning stage.}
\label{tab:ablation_on_cotraining_loss_weight}
\small
\setlength{\tabcolsep}{2pt}
\renewcommand{\arraystretch}{0.85}
\begin{tabularx}{\linewidth}{
  >{\centering\arraybackslash}p{0.7cm} |
  >{\centering\arraybackslash}X
  >{\centering\arraybackslash}X
  >{\centering\arraybackslash}X
  >{\centering\arraybackslash}X
  >{\centering\arraybackslash}X |
  >{\centering\arraybackslash}p{1.0cm}
}
\toprule
$\lambda_3$ & \textbf{NC$\uparrow$} & \textbf{DAC$\uparrow$} & \textbf{TTC$\uparrow$} & \textbf{C.$\uparrow$} & \textbf{EP$\uparrow$} & \textbf{PDMS$\uparrow$} \\
\midrule
0.05  & 98.5 & 98.8 & 95.0 & 99.3 & 84.6 & 90.47 \\
0.1   & 98.6 & 98.9 & 95.1 & 99.3 & 84.6 & \textbf{90.59} \\
0.2   & 98.6 & 98.7 & 94.6 & 99.4 & 83.9 & 90.01 \\
1.0   & 98.6 & 98.3 & 95.0 & 99.4 & 83.4 & 89.75 \\
\bottomrule
\end{tabularx}
\vspace{-8.4ex} % 补偿高度差：左表6行，右表4行
\end{minipage}
\end{table}

\subsection{Scene Noise Robustness Analysis}
\textbf{Quantitative Analysis.}
To further analyze the behavior of HyWorldVLA under the noisy conditions, a noise robustness test set was built by collecting 655 cases from the Openscene test dataset ~\citep{openscene2023}, all of which contain non-uniform noise due to rain and fog. Several methods and configurations are compared in terms of performance: WoTE~\citep{li2025end}, DriveLaW~\citep{xia2026drivelaw}, DriveVLA-W0~\citep{li2025drivevla}, HyWorldVLA(Pure WM), HyWorldVLA(w/o latent supervision) and HyWorldVLA. As shown in Table~\ref{tab:scene_noise_robustness}, HyWorldVLA achieves the highest performance of 86.87 on corrupted cases, while WoTE~\citep{li2025end}, DriveLaW~\citep{xia2026drivelaw} and DriveVLA-W0~\citep{li2025drivevla} obtain only 60.65, 67.49 and 61.18, respectively, far below that of HyWorldVLA, demonstrating the superior robustness of our method. Pure WAM and w/o latent supervision in co-fine-tuning configurations obtain 69.95 and 73.18, which further clarifies that latent representation design of the HyWorldVLA yields this robust performance in noisy scenarios.

\begin{table}[htbp]
    \caption{Scene noise robustness comparison with other methods on the NAVSIM benchmark.}
    \label{tab:scene_noise_robustness}
    \centering
    \renewcommand{\arraystretch}{0.95}
    % 关键：@{\extracolsep{\fill}} 自动填充空白
    \begin{tabular*}{0.95\linewidth}{
      @{\extracolsep{\fill}}
      l |
      c c c c c | cp{2.0cm}
      @{}
    }
    \toprule
    \textbf{Method/Config} & \textbf{NC$\uparrow$} & \textbf{DAC$\uparrow$} & \textbf{TTC$\uparrow$} &
    \textbf{C.$\uparrow$} & \textbf{EP$\uparrow$} & \textbf{PDMS$\uparrow$} \\
    \midrule
    WoTE~\citep{li2025end} & 98.9 & 76.0 & 64.4 & \textbf{100.0} & 64.4 & \cellcolor{blue!8} 60.65 \\
    DriveLaW~\citep{xia2026drivelaw} & 99.4 & 75.7 & 82.9 & \textbf{100.0} & 69.5 & \cellcolor{blue!8} 67.49 \\
    DriveVLA-W0~\citep{li2025drivevla} & 93.4 & 76.2 & 83.1 & 98.6 & 59.8 & \cellcolor{blue!8} 61.18 \\
    Pure WAM & 99.1 & 75.7 & 94.3 & 99.8 & 67.8 & \cellcolor{blue!8} 69.95 \\
    w/o latent supervision in co-fine-tuning & 99.2 & 78.5 & 94.8 & 99.7 & 70.9 & \cellcolor{blue!8} 73.18 \\
    \midrule
    Ours & \textbf{99.7} & \textbf{91.1} & \textbf{96.9} & 99.2 & \textbf{84.4} & \cellcolor{blue!8} \textbf{86.87} \\
    \bottomrule
    \end{tabular*}
    \end{table}

\begin{figure}[!t]
\centering
\setlength{\abovecaptionskip}{2pt}
\setlength{\belowcaptionskip}{0pt}
\begin{subfigure}{\textwidth}
    % ===== 第一行：左侧文字 + 第一行图片 =====
    \begin{minipage}[b]{0.2\textwidth}
    \centering
    \rotatebox{90}{\textbf{Light}}
    \vspace{0.8cm}
    \end{minipage}%
    \hspace{-2.0cm}%
    \begin{minipage}[b]{0.85\textwidth}
    \centering
    \begin{minipage}[b]{0.42\textwidth}
    \centering
    \textbf{Front-view Image}
    \includegraphics[
        height=2.5cm, 
        width=\linewidth, 
        keepaspectratio, 
        valign=b,
        trim={0 106 0 0},
        clip
    ]{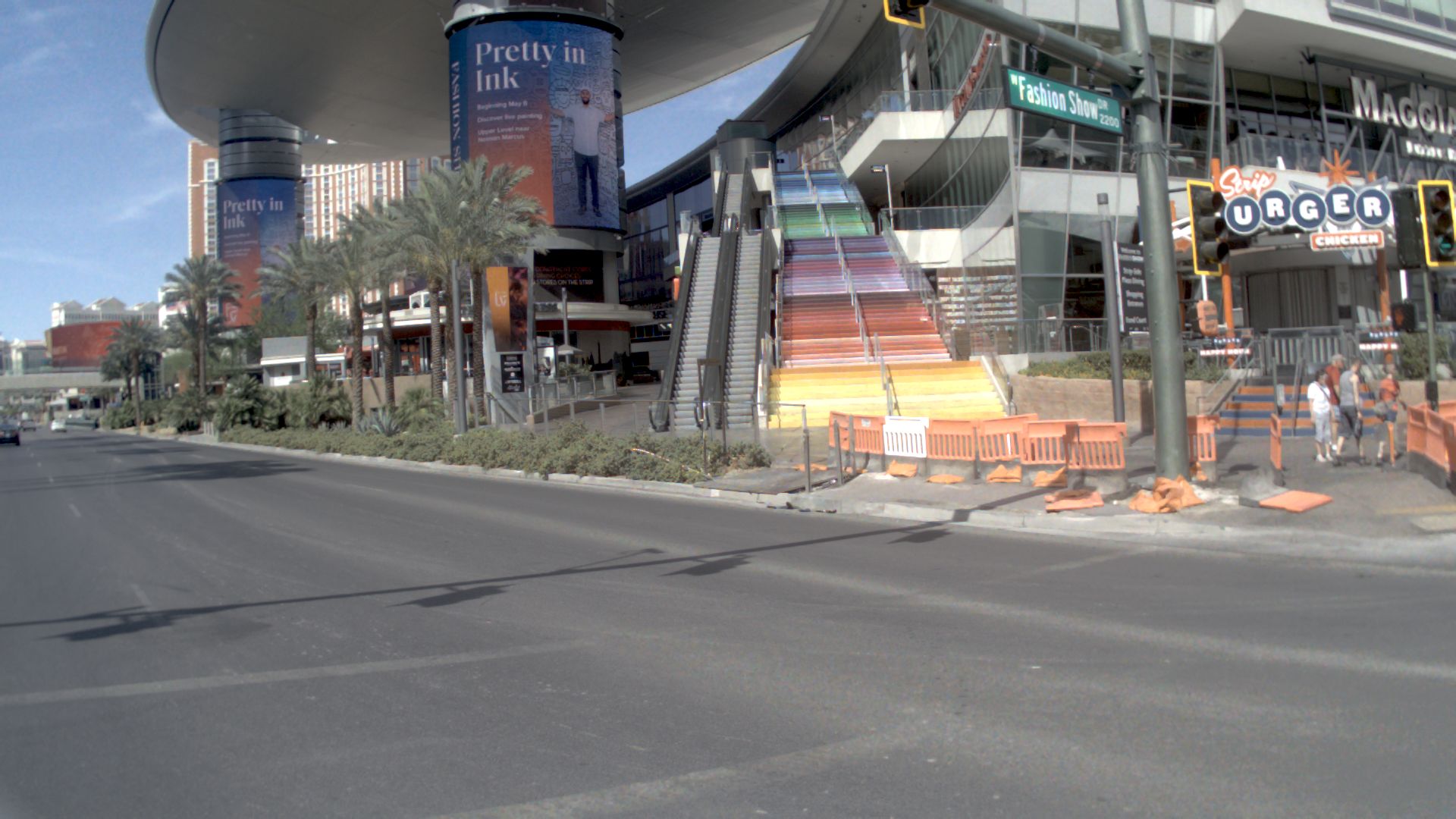}
    \end{minipage}
    \hspace*{0.01cm}%
    \begin{minipage}[b]{0.21\textwidth}
    \centering
    \textbf{Ours}
    \includegraphics[
        height=2.5cm, 
        width=\linewidth, 
        keepaspectratio, 
        valign=b,
        trim={5 5 5 5},
        clip
    ]{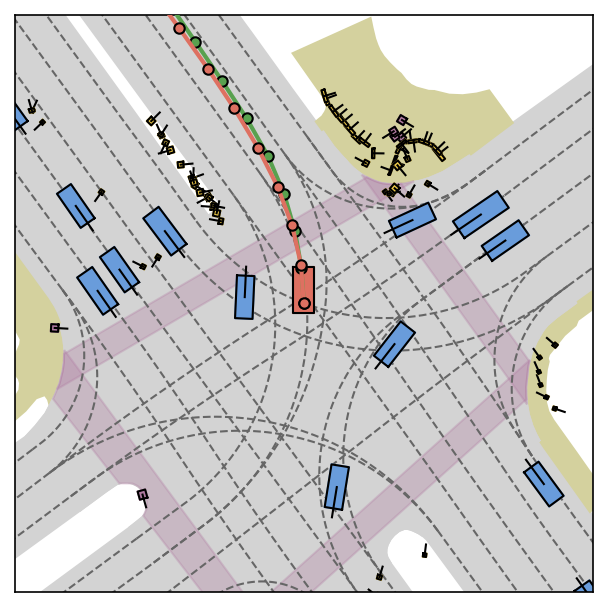}
    \end{minipage}
    \hspace*{0.01cm}%
    \begin{minipage}[b]{0.21\textwidth}
    \centering
    \textbf{DriveVLA-W0}
    \includegraphics[
        height=2.5cm, 
        width=\linewidth, 
        keepaspectratio, 
        valign=b,
        trim={5 5 5 5},
        clip
    ]{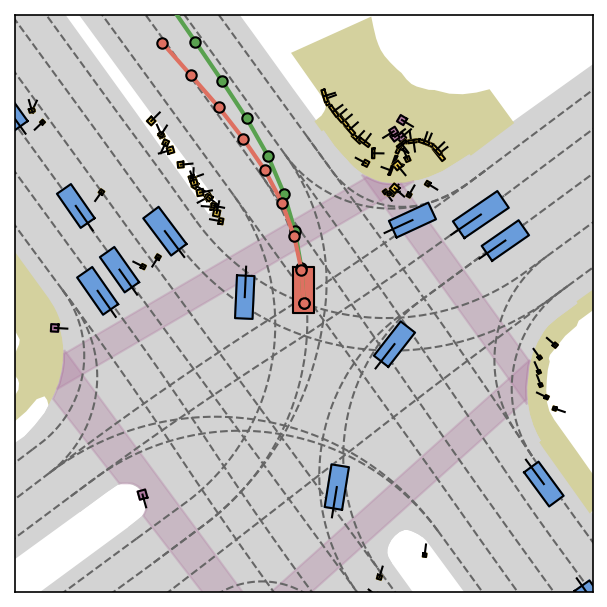}
    \end{minipage}
    \end{minipage}
    % ===== 第二行：左侧文字 + 第二行图片 =====
    \begin{minipage}[b]{0.2\textwidth}
    \centering
    \rotatebox{90}{\textbf{Dark}}
    \vspace{0.8cm}
    \end{minipage}%
    \hspace{-2.0cm}%
    \begin{minipage}[b]{0.85\textwidth}
    \centering
    \begin{minipage}[b]{0.42\textwidth}
    \centering
    \includegraphics[
        height=2.5cm, 
        width=\linewidth, 
        keepaspectratio, 
        valign=b,
        trim={0 150 0 0},
        clip
    ]{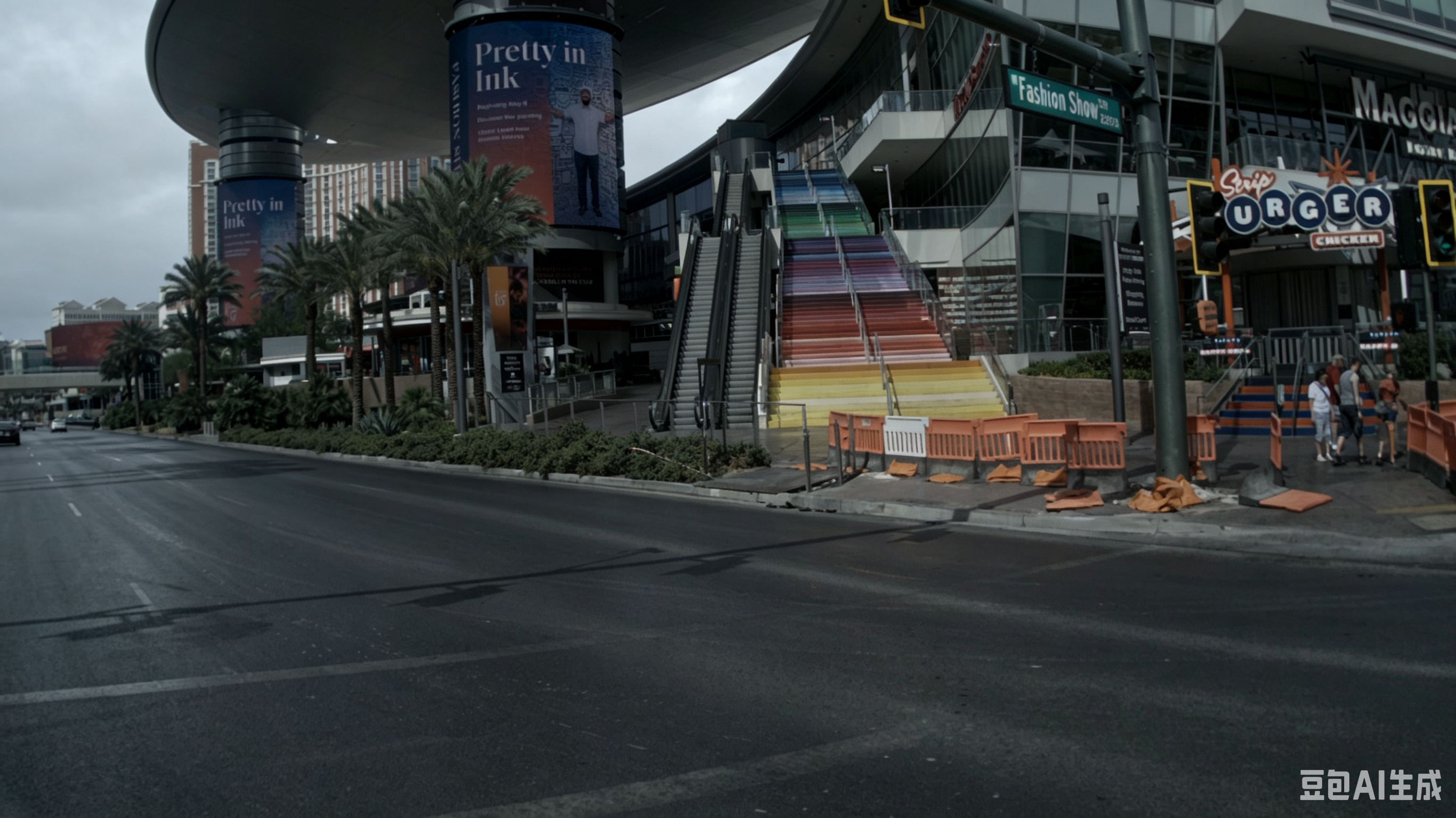}
    \end{minipage}
    \hspace{0.01cm}%
    \begin{minipage}[b]{0.21\textwidth}
    \centering
    \includegraphics[
        height=2.5cm, 
        width=\linewidth, 
        keepaspectratio, 
        valign=b,
        trim={5 5 5 5},
        clip
    ]{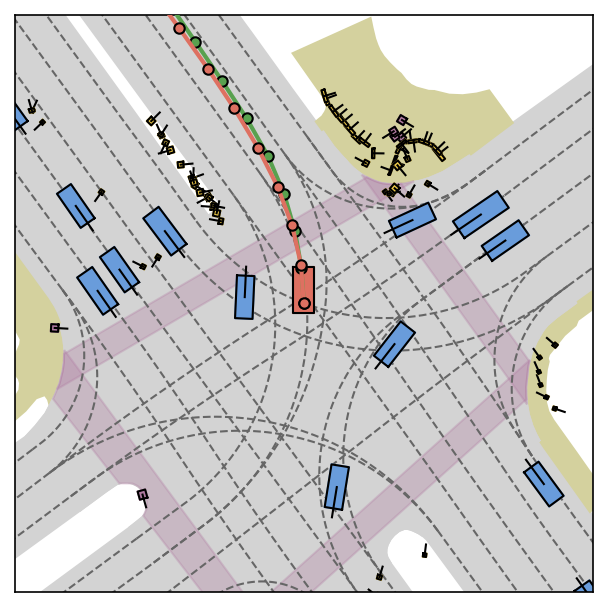}
    \end{minipage}
    \hspace{0.01cm}%
    \begin{minipage}[b]{0.21\textwidth}
    \centering
    \includegraphics[
        height=2.5cm, 
        width=\linewidth, 
        keepaspectratio, 
        valign=b,
        trim={5 5 5 5},
        clip
    ]{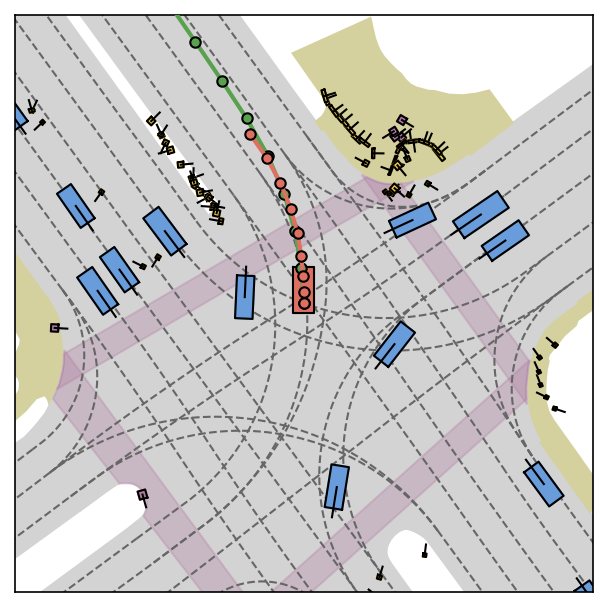}
    \end{minipage}
    \end{minipage}
    \caption{Turn right under the fluctuating illumination.}
    \label{fig:robustness_to_light_intensity_left_turn}
\end{subfigure}
\begin{subfigure}{\textwidth}
    % ===== 第一行：左侧文字 + 第一行图片 =====
    \begin{minipage}[b]{0.2\textwidth}
    \centering
    \rotatebox{90}{\textbf{Hazy}}
    \vspace{0.8cm}
    \end{minipage}%
    \hspace{-2.0cm}%
    \begin{minipage}[b]{0.85\textwidth}
    \centering
    \begin{minipage}[b]{0.42\textwidth}
    \centering
    \includegraphics[
        height=2.5cm, 
        width=\linewidth, 
        keepaspectratio, 
        valign=b,
        trim={0 100 0 0},
        clip
    ]{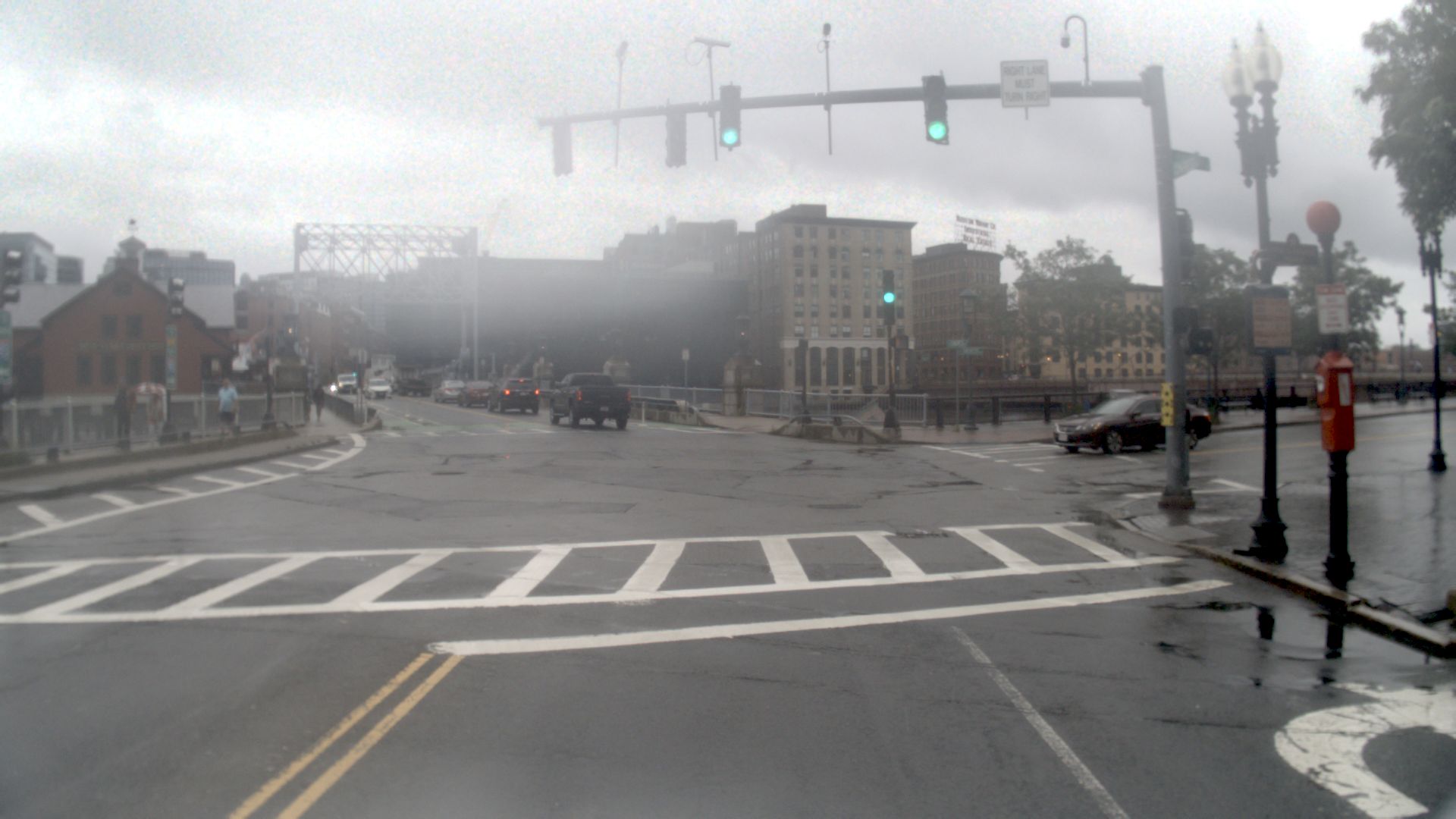}
    \end{minipage}
    \begin{minipage}[b]{0.21\textwidth}
    \centering
    \includegraphics[
        height=2.5cm, 
        width=\linewidth, 
        keepaspectratio, 
        valign=b,
        trim={5 5 5 5},
        clip
    ]{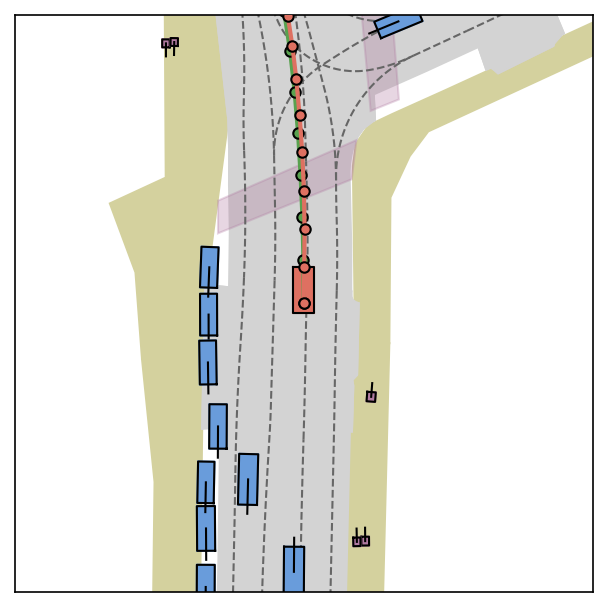}
    \end{minipage}
    \begin{minipage}[b]{0.21\textwidth}
    \centering
    \includegraphics[
        height=2.5cm, 
        width=\linewidth, 
        keepaspectratio, 
        valign=b,
        trim={5 5 5 5},
        clip
    ]{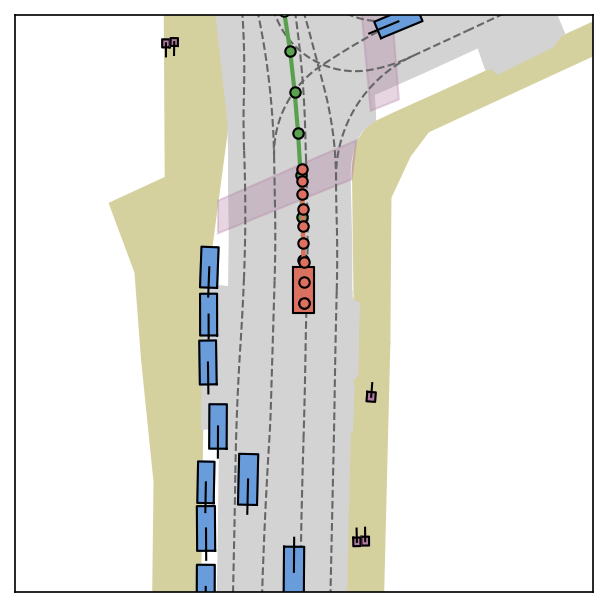}
    \end{minipage}
    \end{minipage}
    % ===== 第二行：左侧文字 + 第二行图片 =====
    \begin{minipage}[b]{0.2\textwidth}
    \centering
    \rotatebox{90}{\textbf{Blurry}}
    \vspace{0.8cm}
    \end{minipage}%
    \hspace{-2.0cm}%
    \begin{minipage}[b]{0.85\textwidth}
        \centering
    \begin{minipage}[b]{0.42\textwidth}
    \centering
    \includegraphics[
        height=2.5cm, 
        width=\linewidth, 
        keepaspectratio, 
        valign=b,
        trim={0 100 0 0},
        clip
    ]{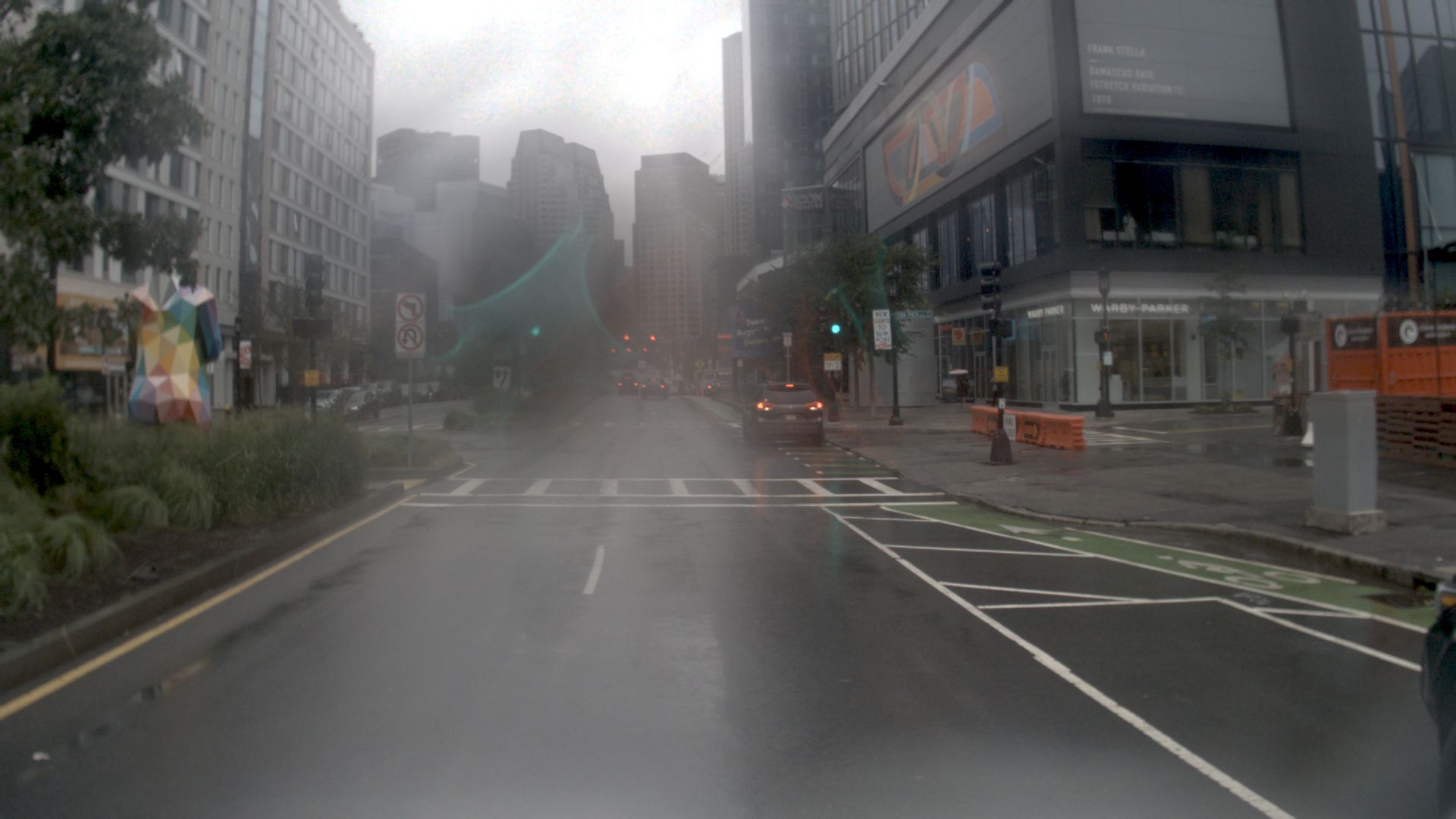}
    \end{minipage}
    \begin{minipage}[b]{0.21\textwidth}
    \centering
    \includegraphics[
        height=2.5cm, 
        width=\linewidth, 
        keepaspectratio, 
        valign=b,
        trim={5 5 5 5},
        clip
    ]{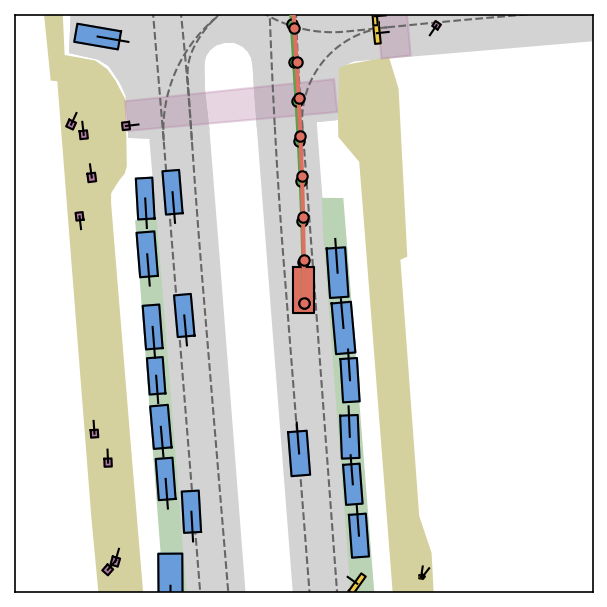}
    \end{minipage}
    \begin{minipage}[b]{0.21\textwidth}
    \centering
    \includegraphics[
        height=2.5cm, 
        width=\linewidth, 
        keepaspectratio, 
        valign=b,
        trim={5 5 5 5},
        clip
    ]{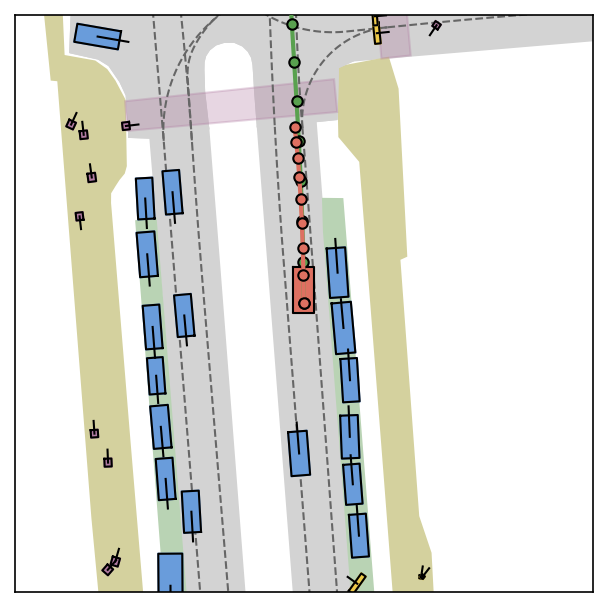}
    \end{minipage}
    \end{minipage}
    \caption{Turn left and go straight under the non-uniform noise from rain and fog.}
    \label{fig:robustness_to_hazy_and_blurry}
\end{subfigure}
\vspace{0.1cm} % 与图例的间距
% ========== 第三行：统一图例 ==========
\centering
\begin{tabular}{c @{\hspace{0.4em}} c @{\qquad} c @{\hspace{0.4em}} c @{\qquad} c @{\hspace{0.4em}} c @{\qquad} c @{\hspace{0.4em}} c @{\qquad} c @{\hspace{0.4em}} c}
  \textcolor{npccolor}{\rule{1.2em}{0.8em}} & NPC &
  \textcolor{egocolor}{\rule{1.2em}{0.8em}} & Ego Car &
  \textcolor{lanecolor}{\hdashrule[0.4ex]{1.6em}{0.05em}{2pt 2pt}} & Lane Center Line &
  \textcolor{gtcolor}{\Large$\bullet$} & GT Traj. &
  \textcolor{predcolor}{\Large$\bullet$} & Pred Traj. \\
\end{tabular}
% ========== 第四行：图题 ==========
\vspace{0.1cm} % 与图例的间距
\caption{\textbf{Hybrid World modeling improves the robustness to the scene noise.} 
(a) DriveVLA-W0 suffers from severe behavioral variance under illumination changes, while our approach retains remarkable consistency. (b) DriveVLA-W0 exhibits overly conservative behavior that sacrifices traffic efficiency, while our method retains accurate driving with human-aligned trajectories.}
\label{fig:robustness_to_light_and_rain}
\vspace{-0.3cm}
\end{figure}

\textbf{Qualitative Analysis.}
% We modify the illumination in front-view camera images, shifting from sunny to overcast conditions, and then benchmark the model’s performance. As shown in Figure~\ref{fig:robustness_to_light_intensity_left_turn}, HyWorldVLA exhibits consistent behavior that remains unaffected by illumination changes, whereas DriveVLA-W0 shows substantial inconsistency. Figure~\ref{fig:robustness_to_hazy_and_blurry} further presents two cases with non-uniform noise from the noise robustness test set we built. Our model still makes correct driving decisions under such degradation, while DriveVLA-W0 produces failed decisions. For more qualitative analysis, including
% the more fluctuating illumination cases(Figure~\ref{fig:robustness_to_light_intensity_right_turn}, Figure~\ref{fig:robustness_to_light_intensity_go_straight}) and noise from rain and fog cases(Figure~\ref{fig:robustness_red_light_stop}, Figure~\ref{fig:robustness_green_light_stop}), please refer to Appendix \ref{sec:more_vis}.
For a qualitative robustness assessment,we systematically modulate front-view illumination, transitioning from sunny to overcast conditions. As illustrated in Fig.~\ref{fig:robustness_to_light_intensity_left_turn}, HyWorldVLA maintains consistent behavioral patterns largely invariant to lighting variations, whereas DriveVLA-W0 exhibits significant instability. Furthermore, Fig.~\ref{fig:robustness_to_hazy_and_blurry} showcases two additional cases drawn from our proposed noise robustness test set, featuring non-uniform visual degradations. Under these challenging conditions, HyWorldVLA continues to execute correct driving maneuvers, while DriveVLA-W0 fails to respond appropriately. Additional cases are in Appendix \ref{sec:more_vis} (Figs.~\ref{fig:robustness_to_light_intensity}-\ref{fig:robustness_green_light_stop}).

% =======================================
% ============ light dark ===============
% =======================================

%% file: chapters/conclusion.tex
\section{Conclusion}
\label{conc}
% In this paper, we propose a hybrid world model for autonomous driving that reconciles the fine-grained spatiotemporal modeling capacity of pixel-regression world models with the inherent noise robustness of latent world models. To bridge this gap, the pixel-level reconstruction preserve fine-grained spatial structures and realistic dynamic details. In parallel, the latent prediction learns semantically meaningful scene state representations in the VAE latent space, which naturally endows the model with robustness against the scene noise. Extensive experiments on the NAVSIM V1 and NAVSIM V2 benchmarks demonstrate that our proposed method achieves state-of-the-art performance, outperforming both representative pixel-regression baselines and leading latent world models by a notable margin. Furthermore, dedicated robustness evaluations on rain- and fog-corrupted camera inputs validate the scene noise robustness of our hybrid design. 

% Despite the promising results, our work has several limitations that motivate future investigation. Our dual-supervision formulation is currently built upon monocular camera input; extending the hybrid paradigm to multi-modal sensor configurations. Moreover, the latent motion space is bounded by the quality and domain scope of the pretrained video VAE, which may induce distribution shift across novel scenarios. We expect that strengthening the coupling of latent dynamics with focusing on more critical transportation elements will extend our method's applicability to real-world driving.
We present a hybrid autonomous driving world model that combines the fine-grained spatiotemporal modeling of pixel-based methods with the noise robustness of latent variants. Pixel-level reconstruction preserves detailed spatial and dynamic information, while latent prediction in the VAE space inherently improves noise resilience. Experiments on NAVSIM v1/v2 show our method achieves state-of-the-art performance, with further validation on rain/fog-corrupted inputs. Limitations and future work include extending the monocular-based framework to multi-modal sensors and strengthening latent dynamics coupling with critical traffic elements.

\clearpage

%% file: chapters/appendix.tex
\clearpage
\newpage
\appendix
\newlength{\wmimageheight}

% 在附录最前面插入局部目录（大标题 "APPENDIX"）
\section*{Appendix}
\begin{spacing}{0.85}
\startcontents[appendix]
\printcontents[appendix]{}{0}{}
\end{spacing}

\section{Overall}
This appendix contains supplementary materials that accompany the main paper. \S~\ref{sec:more_exps} reports additional experiments, which include the comparison of the flow matching action expert on the NAVSIM benchmark, as well as an ablation study of the latent loss weights for the scene noise robustness. Further qualitative findings are presented in \S~\ref{sec:more_vis}, images produced by our world model, featuring visualizations of challenging case studies and the qualitative analyses on illumination interference and blur noise cases. Finally, \S~\ref{sec:use_llms} discloses our use of LLMs for writing assistance.

\section{More Experiments}
\label{sec:more_exps}
\subsection{Flow Matching Based Experiments}
In order to verify the versatility of our approach on different types of action models, in addition to the anchor selective action model method, we also conduct comparative experiments with the flow‑matching action model as shown in Table~\ref{tab:fm_navsim}.
\begin{table}[htbp]
\caption{Flow matching as action expert version comparison on the NAVSIM benchmark.}
\label{tab:fm_navsim}
\centering
% \fontsize{8pt}{9pt}\selectfont
% \setlength{\tabcolsep}{3pt}
\renewcommand{\arraystretch}{0.85}
\renewcommand{\topfraction}{1.0}
\renewcommand{\floatpagefraction}{1.0}
% 关键：@{\extracolsep{\fill}} 自动填充空白
\begin{tabular*}{0.95\linewidth}{
  @{\extracolsep{\fill}}
  l 
  c c c c c c 
  @{}
}
\toprule
\textbf{Method} & \textbf{NC$\uparrow$} & \textbf{DAC$\uparrow$} & \textbf{TTC$\uparrow$} &
\textbf{C.$\uparrow$} & \textbf{EP$\uparrow$} & \textbf{PDMS$\uparrow$} \\
\midrule
DriveVLA-W0 (fm) & 98.4 & 95.3 & 95.2 & \textbf{100.0} & 80.9 & 87.2 \\
Ours (fm) & \textbf{98.8} & \textbf{96.3} & \textbf{95.5} & \textbf{100.0} & \textbf{82.9} & \textbf{88.76} \\
\bottomrule
\end{tabular*}
\end{table}

\subsection{Ablation Study For The Scene Noise Robustness}
To further clarify how latent representation contributes to noise robustness, we perform an ablation study on $\lambda_3$. Table~\ref{tab:ablation_on_cotraining_loss_weight_noise} shows that larger $\lambda_3$ consistently boosts model performance on noisy data, validating the beneficial role of latent representation in noise robustness. Nevertheless, unbounded increase of $\lambda_3$ degrades performance on normal cases, making the trade-off between these two capabilities a critical future research direction.
\begin{table}[htbp]
\centering
\caption{Ablation study of the latent loss weights on the noise robustness test.}
\label{tab:ablation_on_cotraining_loss_weight_noise}
\begin{tabularx}{0.7\linewidth}{
  >{\centering\arraybackslash}X |  % 第一列末尾加竖线
  >{\centering\arraybackslash}X
  >{\centering\arraybackslash}X
  >{\centering\arraybackslash}X
  >{\centering\arraybackslash}X
  >{\centering\arraybackslash}X
  >{\raggedright\arraybackslash}p{1.5cm}
}
\toprule
$\lambda_3$ & \textbf{NC$\uparrow$} & \textbf{DAC$\uparrow$} & \textbf{TTC$\uparrow$} & \textbf{C.$\uparrow$} & \textbf{EP$\uparrow$} & \textbf{PDMS$\uparrow$} \\
\midrule
0.05  & 99.5 & 88.7 & 96.6 & 99.1 & 82.4 & \cellcolor{blue!8} 84.52 \\
0.1   & 99.7 & 91.1 & 96.9 & 99.2 & 84.4 & \cellcolor{blue!8} 86.87 \rlap{\scriptsize\textcolor{gtcolor}{$\uparrow\!$+2.35}} \\
0.2   & 99.1 & 94.8 & 94.5 & 99.4 & 88.1 & \cellcolor{blue!8} 89.45 \rlap{\scriptsize\textcolor{gtcolor}{$\uparrow\!$+2.58}} \\
1.0   & 98.8 & 98.3 & 95.1 & 100.0 & 90.8 & \cellcolor{blue!8} 92.94 \rlap{\scriptsize\textcolor{gtcolor}{$\uparrow\!$+3.49}} \\
\bottomrule
\end{tabularx}
\end{table}

\section{Visualization}
\label{sec:more_vis}
\subsection{Future Image Generation}
\begin{figure}[htbp]
    \centering
    % \small
    \footnotesize

    \begin{subfigure}{\textwidth}
        \centering
        \setlength{\extrarowheight}{0pt}   % 移除单元格内额外间距
        \renewcommand{\arraystretch}{0}
        \setlength{\tabcolsep}{0.5pt}  % 控制列间距，可自行调整
        \settoheight{\wmimageheight}{\includegraphics[width=0.2\textwidth]{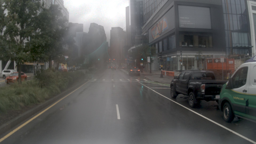}}  % 测量图片高度
        \begin{tabular}{c c c c c c c}
            % 第一行：列标签（表头）
            & V$_{t-1}$ & V$_t$ \\
            \raisebox{\dimexpr 0.5\wmimageheight - 0.5\ht\strutbox \relax}{Input} &
            \includegraphics[width=0.2\textwidth]{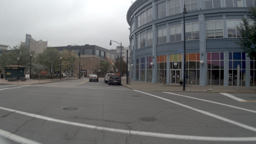} &
            \includegraphics[width=0.2\textwidth]{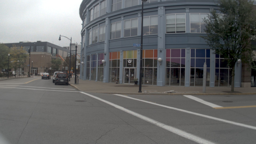} \\ [0.5pt]
            & V$_{t+1}$ & V$_{t+2}$ & V$_{t+3}$ & V$_{t+4}$ \\
            % 第二行：GT 行
            \raisebox{\dimexpr 0.5\wmimageheight - 0.5\ht\strutbox \relax}{GT} &
            \includegraphics[width=0.2\textwidth]{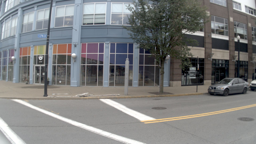} &
            \includegraphics[width=0.2\textwidth]{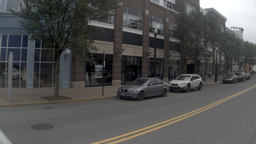} &
            \includegraphics[width=0.2\textwidth]{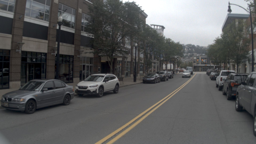} &
            \includegraphics[width=0.2\textwidth]{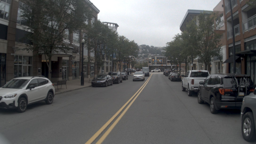} \\ [0.5pt]
            % 第三行：Pred 行
            \raisebox{\dimexpr 0.5\wmimageheight - 0.5\ht\strutbox \relax}{Pred} &
            \includegraphics[width=0.2\textwidth]{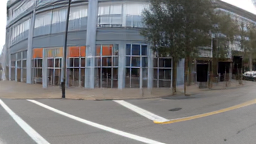} &
            \includegraphics[width=0.2\textwidth]{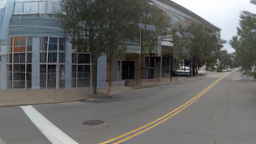} &
            \includegraphics[width=0.2\textwidth]{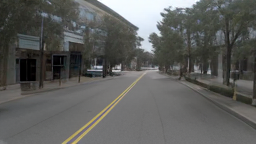} &
            \includegraphics[width=0.2\textwidth]{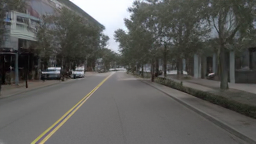} \\ [2pt]
        \end{tabular}
        \caption{Right turn in the intersection with a limited front camera view.}
        \label{fig:wm_case4}
    \end{subfigure}

    \vspace{0.2cm}

    \begin{subfigure}{\textwidth}
        \centering
        \setlength{\extrarowheight}{0pt}   % 移除单元格内额外间距
        \renewcommand{\arraystretch}{0}
        \setlength{\tabcolsep}{0.5pt}  % 控制列间距，可自行调整
        \settoheight{\wmimageheight}{\includegraphics[width=0.2\textwidth]{figs/wm/90831b78d185503a/frame_000.png}}  % 测量图片高度
        \begin{tabular}{c c c c c c c}
            % 第一行：列标签（表头）
            & V$_{t-1}$ & V$_t$ \\
            \raisebox{\dimexpr 0.5\wmimageheight - 0.5\ht\strutbox \relax}{Input} &
            \includegraphics[width=0.2\textwidth]{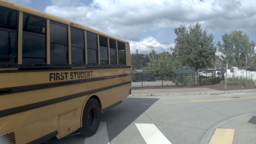} &
            \includegraphics[width=0.2\textwidth]{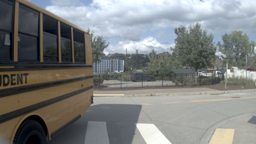} \\ [0.5pt]
            & V$_{t+1}$ & V$_{t+2}$ & V$_{t+3}$ & V$_{t+4}$ \\
            % 第二行：GT 行
            \raisebox{\dimexpr 0.5\wmimageheight - 0.5\ht\strutbox \relax}{GT} &
            \includegraphics[width=0.2\textwidth]{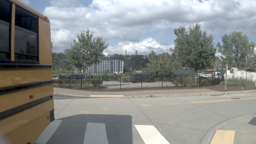} &
            \includegraphics[width=0.2\textwidth]{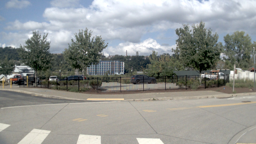} &
            \includegraphics[width=0.2\textwidth]{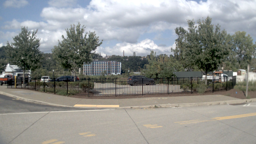} &
            \includegraphics[width=0.2\textwidth]{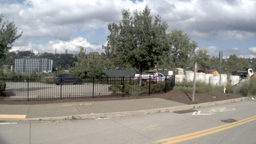} \\ [0.5pt]
            % 第三行：Pred 行
            \raisebox{\dimexpr 0.5\wmimageheight - 0.5\ht\strutbox \relax}{Pred} &
            \includegraphics[width=0.2\textwidth]{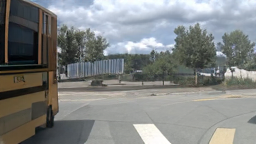} &
            \includegraphics[width=0.2\textwidth]{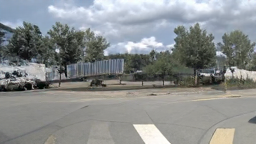} &
            \includegraphics[width=0.2\textwidth]{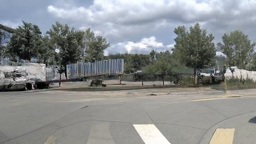} &
            \includegraphics[width=0.2\textwidth]{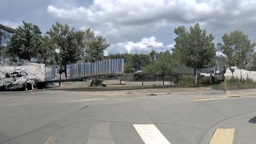} \\ [2pt]
        \end{tabular}
        \caption{Meeting an oncoming bus on a narrow road.}
        \label{fig:wm_case5}
    \end{subfigure}

    \caption{\textbf{Future image generation.} (a) During right turns, despite the absence of lane markings in front-view camera inputs, our model reliably forecasts future lane markings, evidencing its learned world knowledge. (b) Our model correctly forecasts the motion pattern of the oncoming bus and the ego car’s corresponding rightward yielding action, evidencing its complex scene understanding capability.}  % 整张大图的标题
    \label{fig:wm_case}
\end{figure}

Our model inherits the strong future frame prediction capability of the World Action Model. As shown in Figure~\ref{fig:wm_case4}, with only a single front-view camera, our model generates the central double yellow line absent from the input frames. Moreover, our model accurately forecasts future maneuvers of the ego vehicle and obstacles in the highly interactive scenario, as shown in Figure~\ref{fig:wm_case5}.

\subsection{Planning Trajectory}
\begin{figure}[htbp]
\centering
% ===== 第一行：第一行图片 =====
\begin{minipage}[b]{0.47\textwidth}
\centering
\textbf{Front-view Image}
\vspace{0.2cm}
\includegraphics[
    height=3.5cm, 
    width=\linewidth, 
    keepaspectratio, 
    valign=b,
    trim={0 100 0 0},
    clip
]{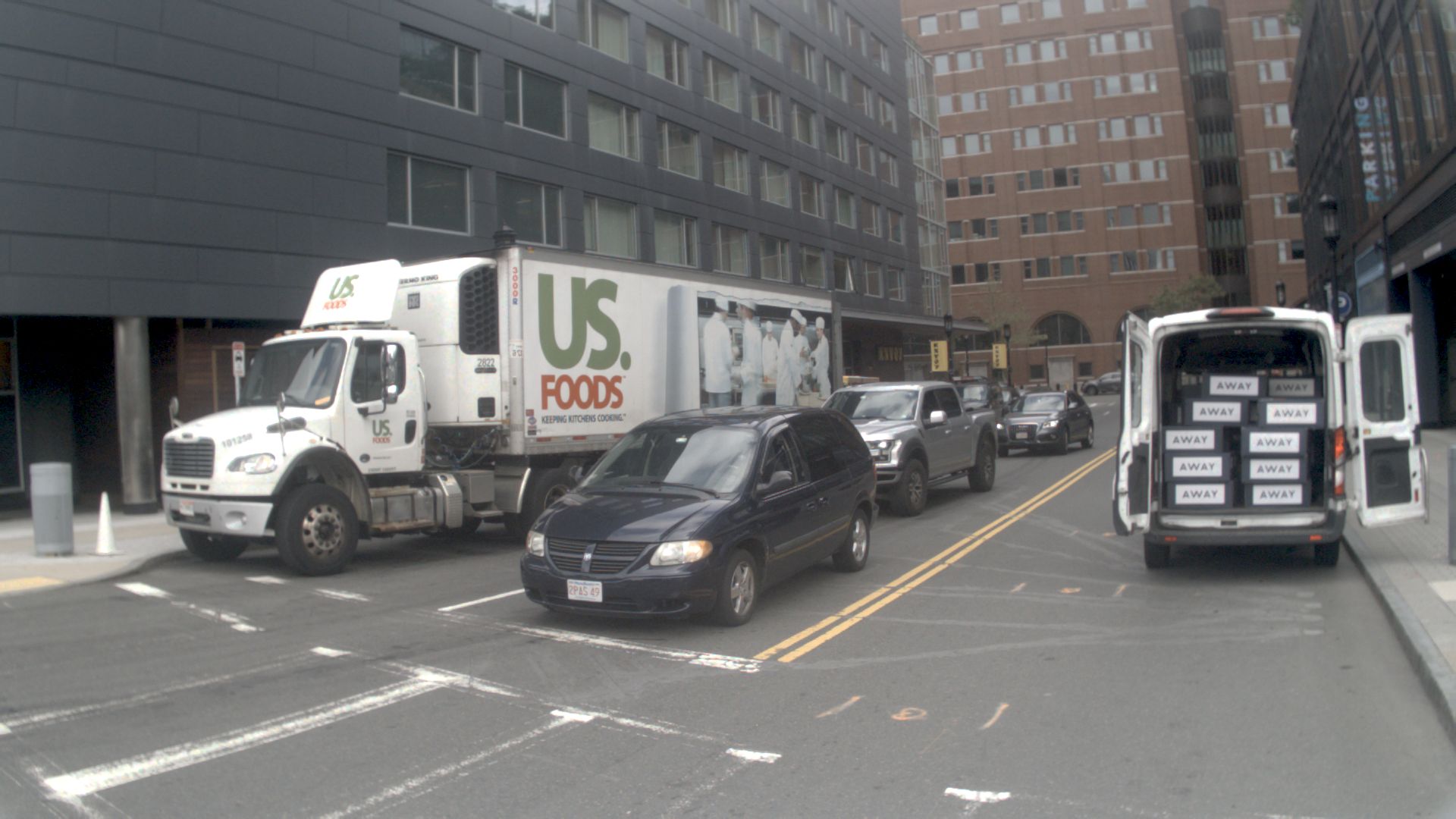}
\end{minipage}
\hfill
\begin{minipage}[b]{0.24\textwidth}
\centering
\textbf{Ours}
\vspace{0.2cm}
\includegraphics[
    height=4cm, 
    width=\linewidth, 
    keepaspectratio, 
    valign=b,
    trim={5 5 5 5},
    clip
]{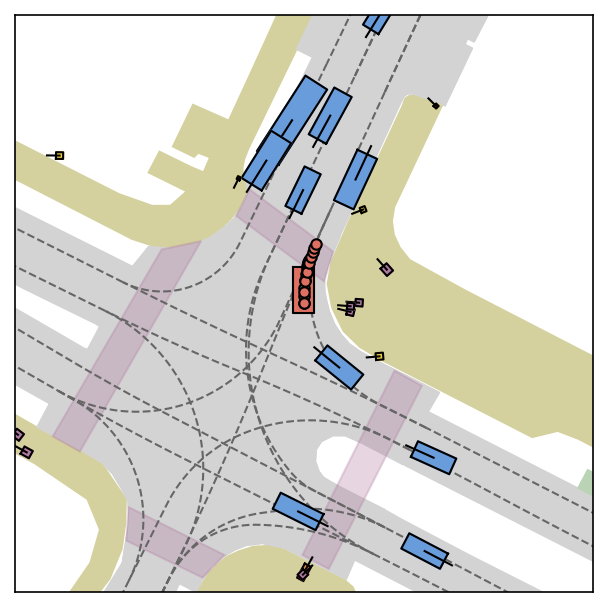}
\end{minipage}
\hfill
\begin{minipage}[b]{0.24\textwidth}
\centering
\textbf{DriveVLA-W0}
\vspace{0.2cm}
\includegraphics[
    height=4cm, 
    width=\linewidth, 
    keepaspectratio, 
    valign=b,
    trim={5 5 5 5},
    clip
]{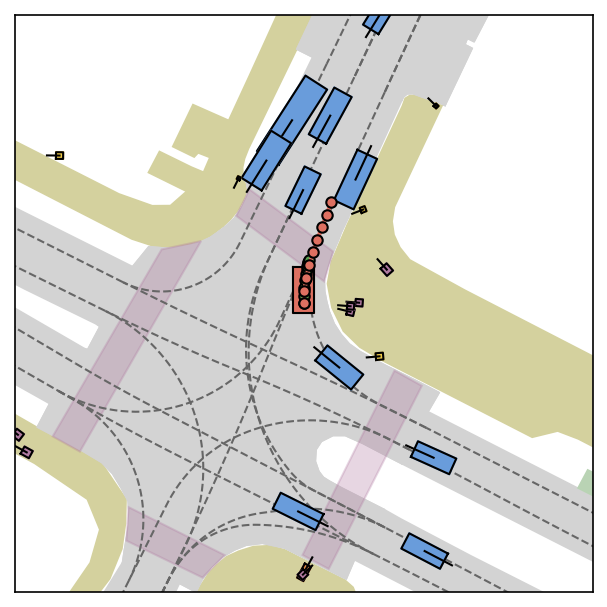}
\end{minipage}
% \vspace{0.1cm}
% ========== 第二行：统一图例 ==========
\vspace{0.2em}
\centering
\begin{tabular}{c @{\hspace{0.4em}} c @{\qquad} c @{\hspace{0.4em}} c @{\qquad} c @{\hspace{0.4em}} c @{\qquad} c @{\hspace{0.4em}} c @{\qquad} c @{\hspace{0.4em}} c}
  \textcolor{npccolor}{\rule{1.2em}{0.8em}} & NPC &
  \textcolor{egocolor}{\rule{1.2em}{0.8em}} & Ego Car &
  \textcolor{lanecolor}{\hdashrule[0.4ex]{1.6em}{0.05em}{2pt 2pt}} & Lane Center Line &
  \textcolor{gtcolor}{\Large$\bullet$} & GT Traj. &
  \textcolor{predcolor}{\Large$\bullet$} & Pred Traj. \\
\end{tabular}
% ========== 第三行：图题 ==========
\vspace{0.2em} % 与图例的间距
\caption{\textbf{Turn right and stop behind a truck with its trunk open.}
DriveVLA-W0 baseline fails in trying to nudge a stopping truck with its trunk open. Our method stops and keep a fit distance to it.}
\label{fig:good_case_stop_behind_truck_2}
\end{figure}

\begin{figure}[htbp]
\centering
% ===== 第一行：第一行图片 =====
\begin{minipage}[b]{0.47\textwidth}
\centering
\textbf{Front-view Image}
\vspace{0.2cm}
\includegraphics[
    height=3.5cm, 
    width=\linewidth, 
    keepaspectratio, 
    valign=b,
    trim={0 100 0 0},
    clip
]{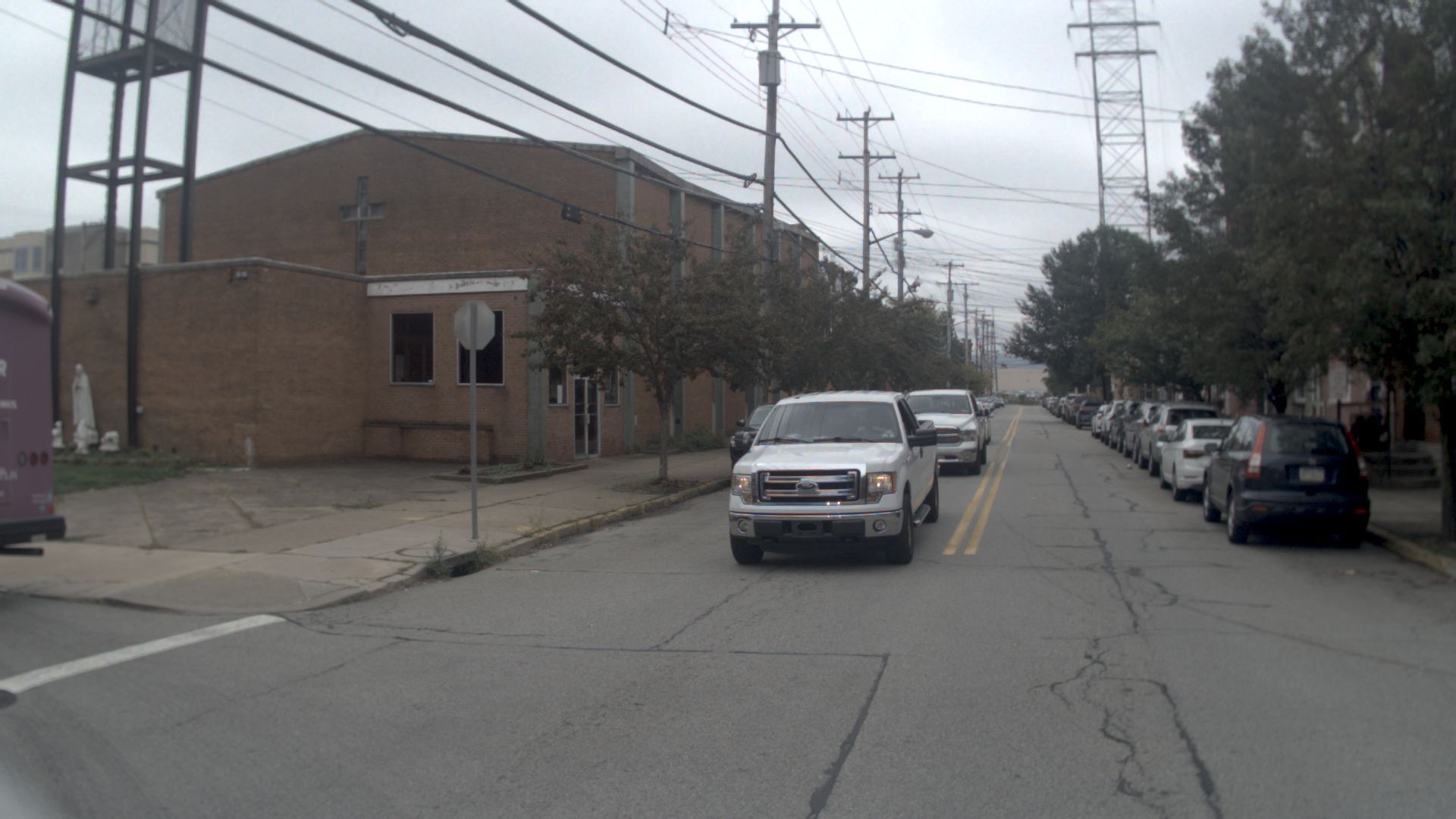}
\end{minipage}
\hfill
\begin{minipage}[b]{0.24\textwidth}
\centering
\textbf{Ours}
\vspace{0.2cm}
\includegraphics[
    height=4cm, 
    width=\linewidth, 
    keepaspectratio, 
    valign=b,
    trim={5 5 5 5},
    clip
]{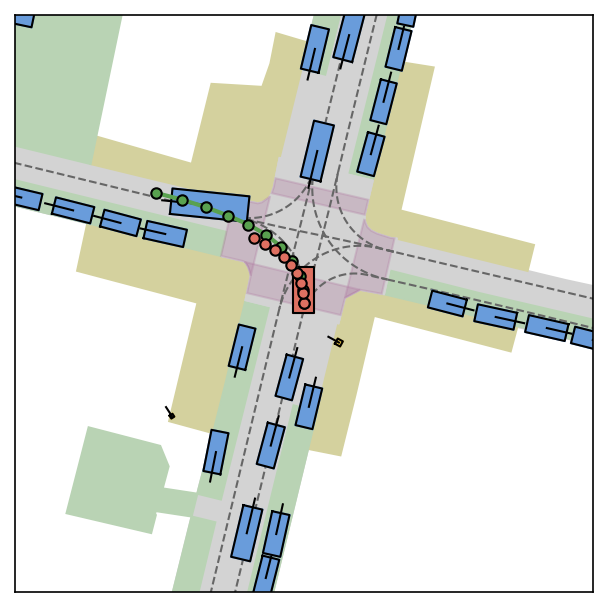}
\end{minipage}
\hfill
\begin{minipage}[b]{0.24\textwidth}
\centering
\textbf{DriveVLA-W0}
\vspace{0.2cm}
\includegraphics[
    height=4cm, 
    width=\linewidth, 
    keepaspectratio, 
    valign=b,
    trim={5 5 5 5},
    clip
]{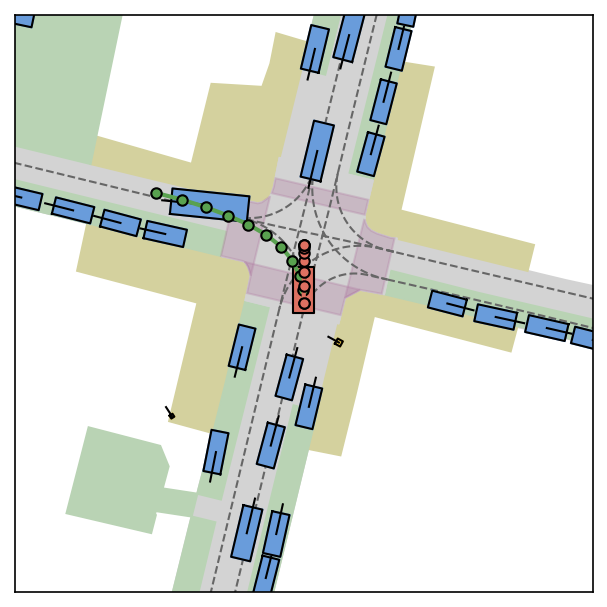}
\end{minipage}
% \vspace{0.1cm}
% ========== 第二行：统一图例 ==========
\vspace{0.2em}
\centering
\begin{tabular}{c @{\hspace{0.4em}} c @{\qquad} c @{\hspace{0.4em}} c @{\qquad} c @{\hspace{0.4em}} c @{\qquad} c @{\hspace{0.4em}} c @{\qquad} c @{\hspace{0.4em}} c}
  \textcolor{npccolor}{\rule{1.2em}{0.8em}} & NPC &
  \textcolor{egocolor}{\rule{1.2em}{0.8em}} & Ego Car &
  \textcolor{lanecolor}{\hdashrule[0.4ex]{1.6em}{0.05em}{2pt 2pt}} & Lane Center Line &
  \textcolor{gtcolor}{\Large$\bullet$} & GT Traj. &
  \textcolor{predcolor}{\Large$\bullet$} & Pred Traj. \\
\end{tabular}
% ========== 第三行：图题 ==========
\vspace{0.2em} % 与图例的间距
\caption{\textbf{Left turn in a small intersection with the limited view.}
DriveVLA-W0 baseline goes straight at a small intersection where it is supposed to turn left. Our method turns left even though the front view is limited.}
\label{fig:good_case_left_turn_samll_junction}
\end{figure}

\begin{figure}[htbp]
\centering
% ===== 第一行：第一行图片 =====
\begin{minipage}[b]{0.47\textwidth}
\centering
\textbf{Front-view Image}
\vspace{0.2cm}
\includegraphics[
    height=3.5cm, 
    width=\linewidth, 
    keepaspectratio, 
    valign=b,
    trim={0 100 0 0},
    clip
]{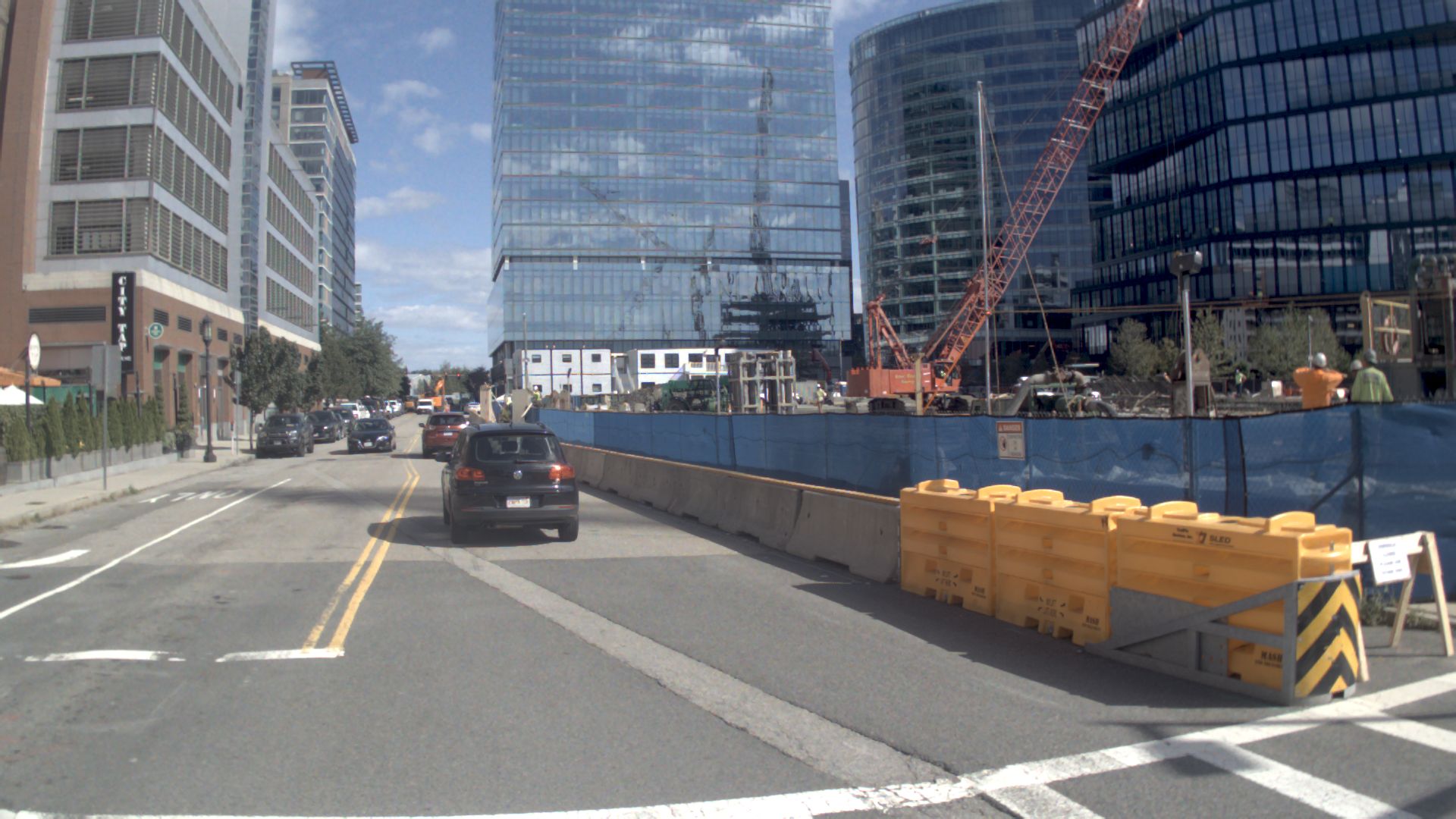}
\end{minipage}
\hfill
\begin{minipage}[b]{0.24\textwidth}
\centering
\textbf{Ours}
\vspace{0.2cm}
\includegraphics[
    height=4cm, 
    width=\linewidth, 
    keepaspectratio, 
    valign=b,
    trim={5 5 5 5},
    clip
]{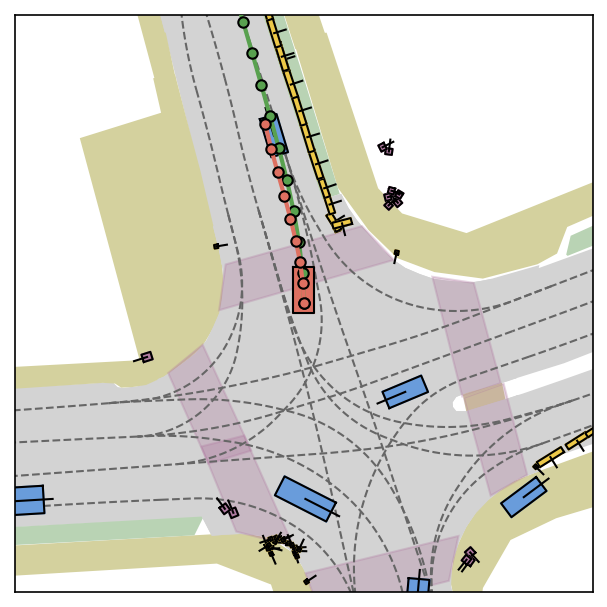}
\end{minipage}
\hfill
\begin{minipage}[b]{0.24\textwidth}
\centering
\textbf{DriveVLA-W0}
\vspace{0.2cm}
\includegraphics[
    height=4cm, 
    width=\linewidth, 
    keepaspectratio, 
    valign=b,
    trim={5 5 5 5},
    clip
]{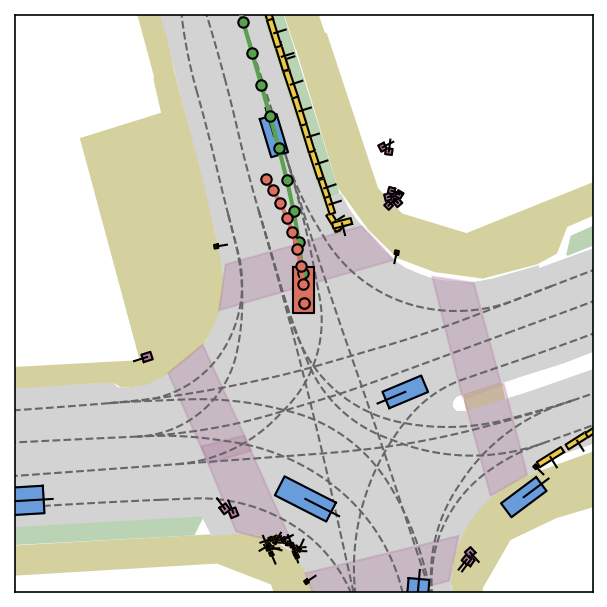}
\end{minipage}
% \vspace{0.1cm}
% ========== 第二行：统一图例 ==========
\vspace{0.2em}
\centering
\begin{tabular}{c @{\hspace{0.4em}} c @{\qquad} c @{\hspace{0.4em}} c @{\qquad} c @{\hspace{0.4em}} c @{\qquad} c @{\hspace{0.4em}} c @{\qquad} c @{\hspace{0.4em}} c}
  \textcolor{npccolor}{\rule{1.2em}{0.8em}} & NPC &
  \textcolor{egocolor}{\rule{1.2em}{0.8em}} & Ego Car &
  \textcolor{lanecolor}{\hdashrule[0.4ex]{1.6em}{0.05em}{2pt 2pt}} & Lane Center Line &
  \textcolor{gtcolor}{\Large$\bullet$} & GT Traj. &
  \textcolor{predcolor}{\Large$\bullet$} & Pred Traj. \\
\end{tabular}
% ========== 第三行：图题 ==========
\vspace{0.2em} % 与图例的间距
\caption{\textbf{Turn left at the construction intersection.}
DriveVLA-W0 baseline goes into the opposite lane at the construction intersection. Our method turns left and keep in the right lane.}
\label{fig:good_case_left_turn_construction}
\end{figure}

In this section, we conduct a qualitative case analysis to gain deeper insights into our model’s behavior and validate our key findings. Specifically, we compare our Hybrid World Model against the baseline DriveVLA-W0 across multiple challenging scenarios to demonstrate, as shown in Figure~\ref{fig:good_case_stop_behind_truck_2}, Figure~\ref{fig:good_case_left_turn_samll_junction} and Figure~\ref{fig:good_case_left_turn_construction}.
% =======================================
% =====appendix planning traj fig 3======
% =======================================

% =======================================
% =====appendix planning traj fig 4======
% =======================================

% =======================================
% =====appendix planning traj fig 5======
% =======================================

\subsection{Failure Case}
\begin{figure}[htbp]
\centering
% ===== 第一行：第一行图片 =====
\begin{minipage}[b]{0.04\textwidth}
\centering
\rotatebox{90}{\textbf{Planning}}
\vspace{0.8cm}
\end{minipage}%
\hfill
\begin{minipage}[b]{0.96\textwidth}
\begin{minipage}[b]{0.38\textwidth}
\centering
\textbf{$V_{t-1}$}
\vspace{0.2cm}
\includegraphics[
    height=3.5cm, 
    width=\linewidth, 
    keepaspectratio, 
    valign=b,
    trim={0 100 0 0},
    clip
]{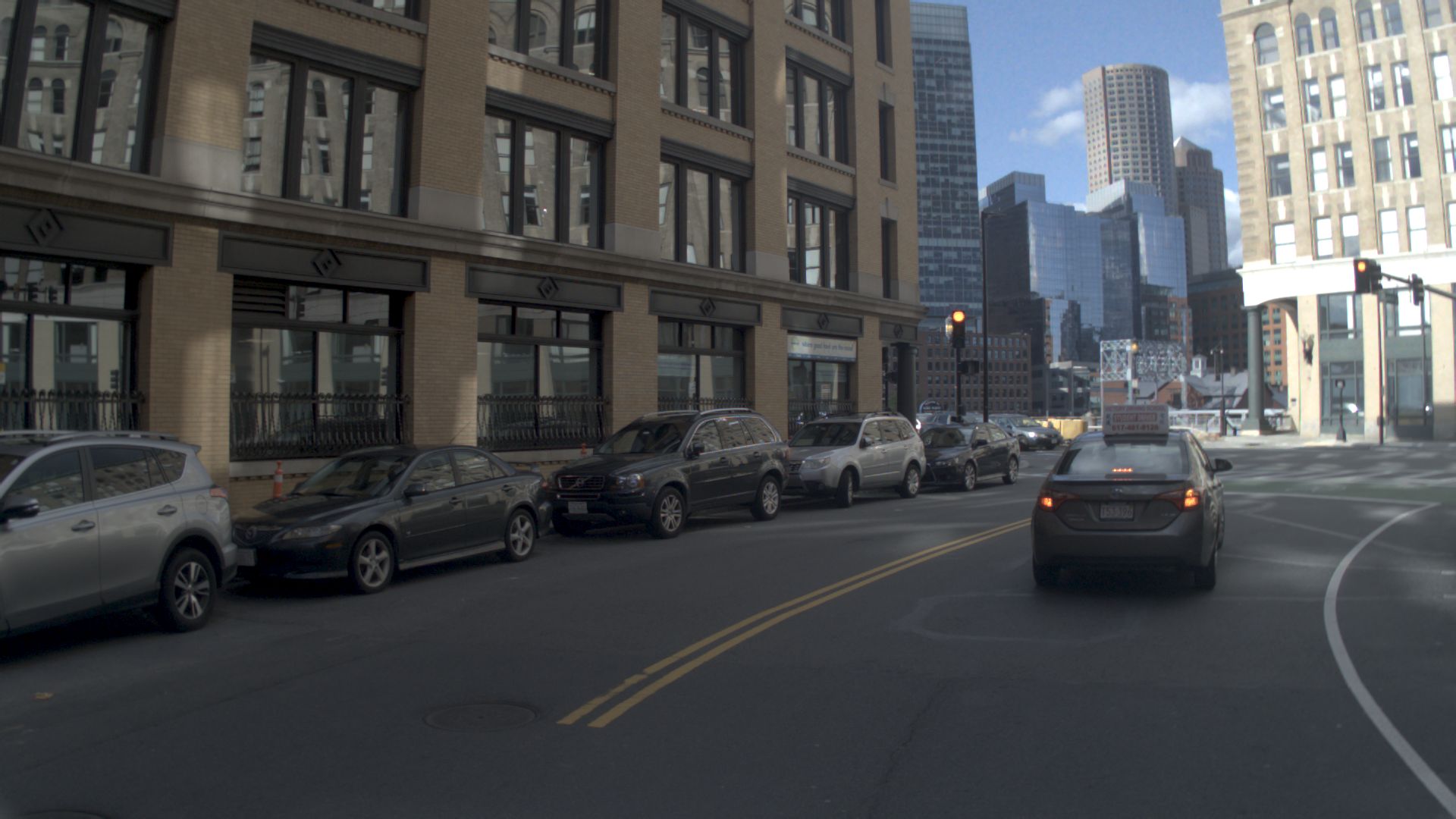}
\end{minipage}
\hspace{0.01\textwidth} % 将 \hfill 改为固定小间距
\begin{minipage}[b]{0.38\textwidth}
\centering
\textbf{$V_t$}
\vspace{0.2cm}
\includegraphics[
    height=3.5cm, 
    width=\linewidth, 
    keepaspectratio, 
    valign=b,
    trim={0 100 0 0},
    clip
]{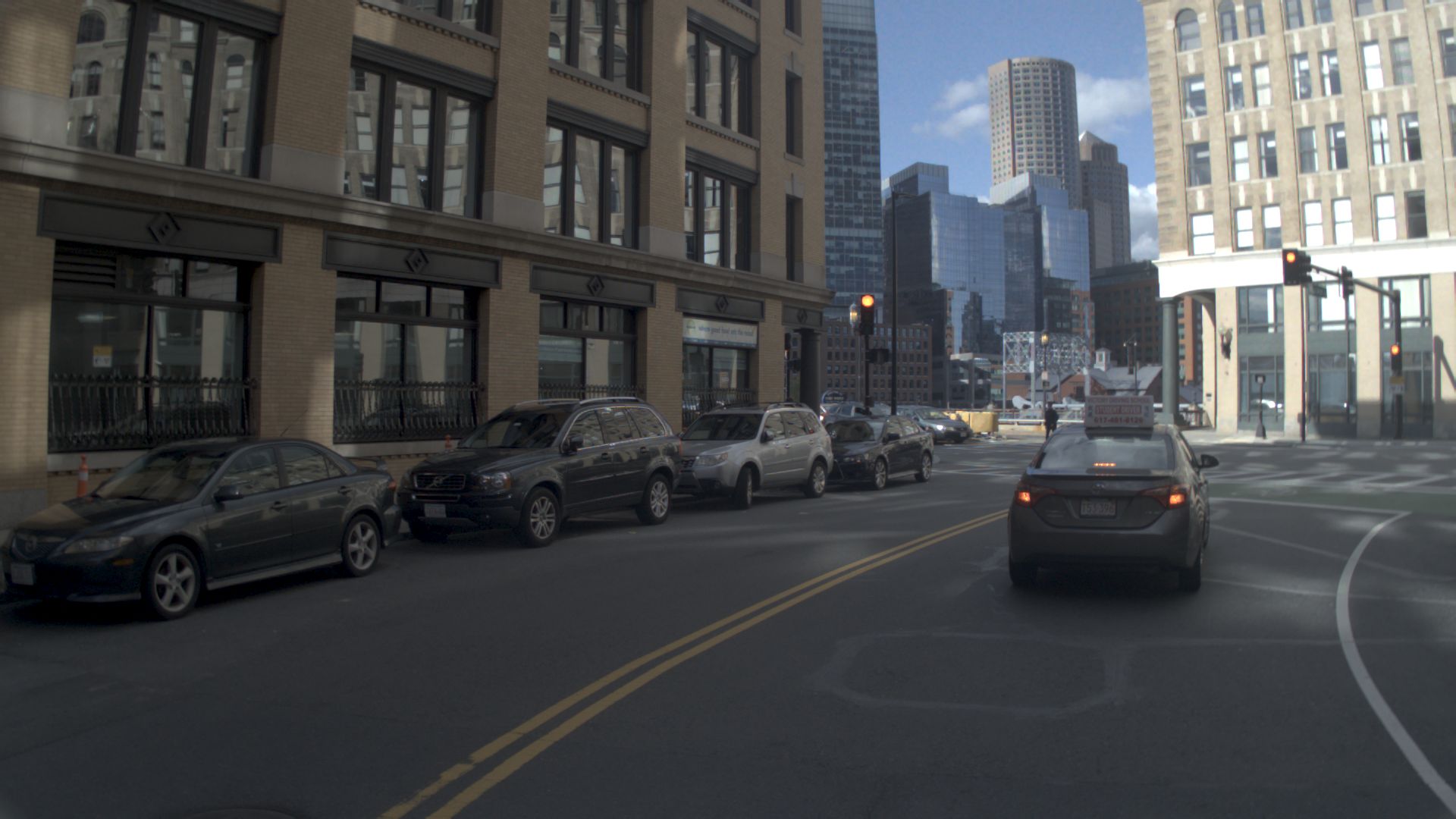}
\end{minipage}
\hspace{0.01\textwidth} % 将 \hfill 改为固定小间距
\begin{minipage}[b]{0.20\textwidth}
\centering
\textbf{Trajectory}
\vspace{0.2cm}
\includegraphics[
    height=4cm, 
    width=\linewidth, 
    keepaspectratio, 
    valign=b,
    trim={5 5 5 5},
    clip
]{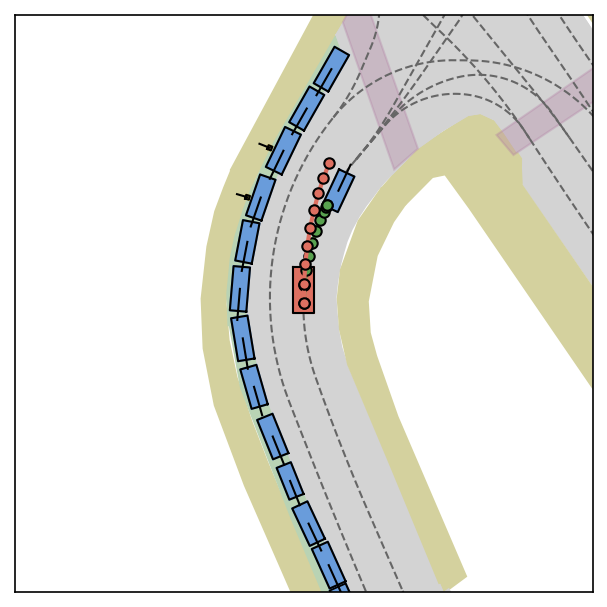}
\end{minipage}
\end{minipage}
% ========== 第二行：world model ======
\begin{minipage}[b]{0.04\textwidth}
\centering
\rotatebox{90}{\textbf{WM}}
\vspace{0.7cm}
\end{minipage}%
\hfill
\begin{minipage}[b]{0.96\textwidth}
\begin{minipage}[b]{0.23\textwidth}
\centering
\textbf{$V_{t+1}$}
\vspace{0.2cm}
\includegraphics[
    height=3.5cm, 
    width=\linewidth, 
    keepaspectratio, 
    valign=b,
    trim={0 0 0 0},
    clip
]{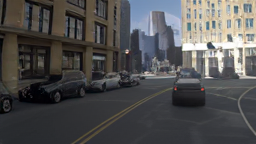}
\end{minipage}
\hspace{0.013\textwidth} % 将 \hfill 改为固定小间距
\begin{minipage}[b]{0.23\textwidth}
\centering
\textbf{$V_{t+2}$}
\vspace{0.2cm}
\includegraphics[
    height=3.5cm, 
    width=\linewidth, 
    keepaspectratio, 
    valign=b,
    trim={0 0 0 0},
    clip
]{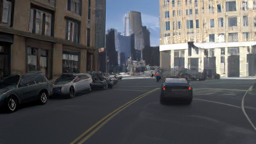}
\end{minipage}
\hspace{0.013\textwidth} % 将 \hfill 改为固定小间距
\begin{minipage}[b]{0.23\textwidth}
\centering
\textbf{$V_{t+3}$}
\vspace{0.2cm}
\includegraphics[
    height=3.5cm,
    width=\linewidth, 
    keepaspectratio, 
    valign=b,
    trim={0 0 0 0},
    clip
]{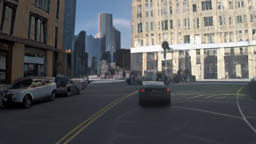}
\end{minipage}
\hspace{0.013\textwidth} % 将 \hfill 改为固定小间距
\begin{minipage}[b]{0.23\textwidth}
\centering
\textbf{$V_{t+4}$}
\vspace{0.2cm}
\includegraphics[
    height=3.5cm, 
    width=\linewidth, 
    keepaspectratio, 
    valign=b,
    trim={0 0 0 0},
    clip
]{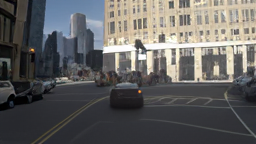}
\end{minipage}
\end{minipage}
% ========== 第三行：GT ======
\begin{minipage}[b]{0.04\textwidth}
\centering
\rotatebox{90}{\textbf{GT}}
\vspace{1.2cm}
\end{minipage}%
\hfill
\begin{minipage}[b]{0.96\textwidth}
\begin{minipage}[b]{0.23\textwidth}
\centering
\includegraphics[
    height=3.5cm, 
    width=\linewidth, 
    keepaspectratio, 
    valign=b,
    trim={0 0 0 0},
    clip
]{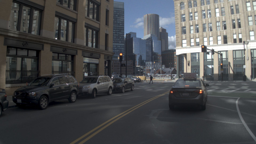}
\end{minipage}
\hspace{0.013\textwidth} % 将 \hfill 改为固定小间距
\begin{minipage}[b]{0.23\textwidth}
\centering
\includegraphics[
    height=3.5cm, 
    width=\linewidth, 
    keepaspectratio, 
    valign=b,
    trim={0 0 0 0},
    clip
]{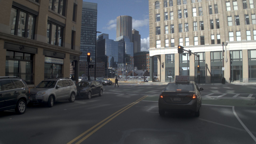}
\end{minipage}
\hspace{0.013\textwidth} % 将 \hfill 改为固定小间距
\begin{minipage}[b]{0.23\textwidth}
\centering
\includegraphics[
    height=3.5cm,
    width=\linewidth, 
    keepaspectratio, 
    valign=b,
    trim={0 0 0 0},
    clip
]{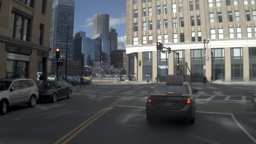}
\end{minipage}
\hspace{0.013\textwidth} % 将 \hfill 改为固定小间距
\begin{minipage}[b]{0.23\textwidth}
\centering
\includegraphics[
    height=3.5cm, 
    width=\linewidth, 
    keepaspectratio, 
    valign=b,
    trim={0 0 0 0},
    clip
]{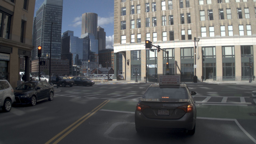}
\end{minipage} 
\vspace{0.2em}
\end{minipage}
% ========== 第四行：统一图例 ==========
\centering
\begin{tabular}{c @{\hspace{0.1em}} c @{\qquad} c @{\hspace{0.1em}} c @{\qquad} c @{\hspace{0.1em}} c @{\qquad} c @{\hspace{0.1em}} c @{\qquad} c @{\hspace{0.1em}} c}
  \textcolor{npccolor}{\rule{1.2em}{0.8em}} & NPC &
  \textcolor{egocolor}{\rule{1.2em}{0.8em}} & Ego Car &
  \textcolor{lanecolor}{\hdashrule[0.4ex]{1.6em}{0.05em}{2pt 2pt}} & Lane Center Line &
  \textcolor{gtcolor}{\Large$\bullet$} & GT Traj. &
  \textcolor{predcolor}{\Large$\bullet$} & Pred Traj. \\
\end{tabular}
% ========== 第五行：图题 ==========
\vspace{0.2em} % 与图例的间距
\caption{\textbf{Failure case analysis: deficits in capturing critical small elements.}
We illustrate a failure case in which our model attempts to nudge a queuing vehicle ahead. The world model’s predictions reveal that the model misclassifies the queuing vehicle as low-speed moving and tries to nudge past it for higher traffic efficiency, failing to recognize the queuing state indicated by the red light ahead and the vehicle’s brake lights. Since this critical cue is only discernible from a few pixels, fine-grained element capture and causal reasoning remain a core future research direction. }
\label{fig:bad_case_nudge}
\end{figure}

\begin{figure}[htbp]
\centering
% ===== 第一行：第一行图片 =====
\begin{minipage}[b]{0.04\textwidth}
\centering
\rotatebox{90}{\textbf{Planning}}
\vspace{0.8cm}
\end{minipage}%
\hfill
\begin{minipage}[b]{0.96\textwidth}
\begin{minipage}[b]{0.38\textwidth}
\centering
\textbf{$V_{t-1}$}
\vspace{0.2cm}
\includegraphics[
    height=3.5cm, 
    width=\linewidth, 
    keepaspectratio, 
    valign=b,
    trim={0 100 0 0},
    clip
]{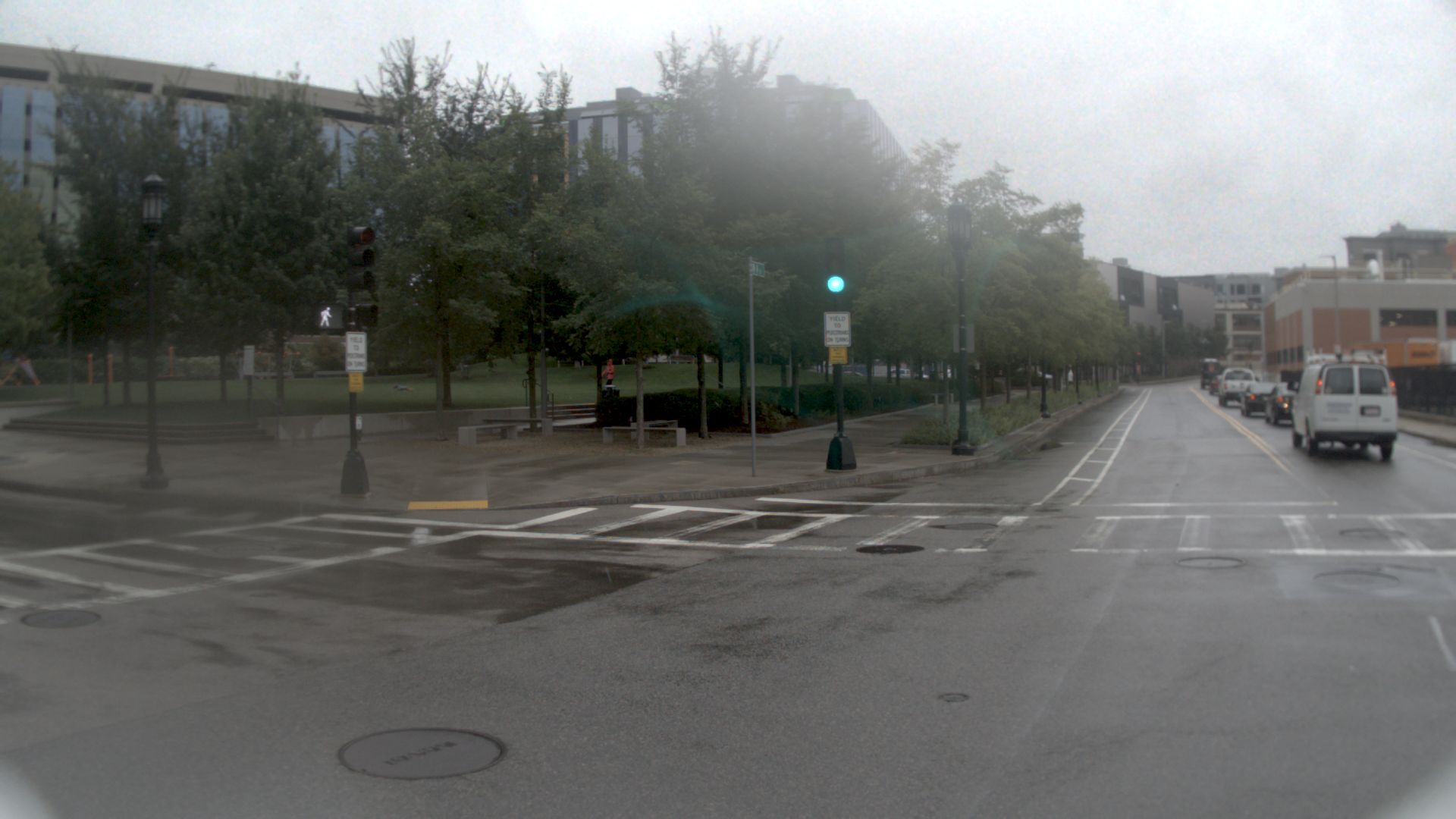}
\end{minipage}
\hspace{0.01\textwidth} % 将 \hfill 改为固定小间距
\begin{minipage}[b]{0.38\textwidth}
\centering
\textbf{$V_t$}
\vspace{0.2cm}
\includegraphics[
    height=3.5cm, 
    width=\linewidth, 
    keepaspectratio, 
    valign=b,
    trim={0 100 0 0},
    clip
]{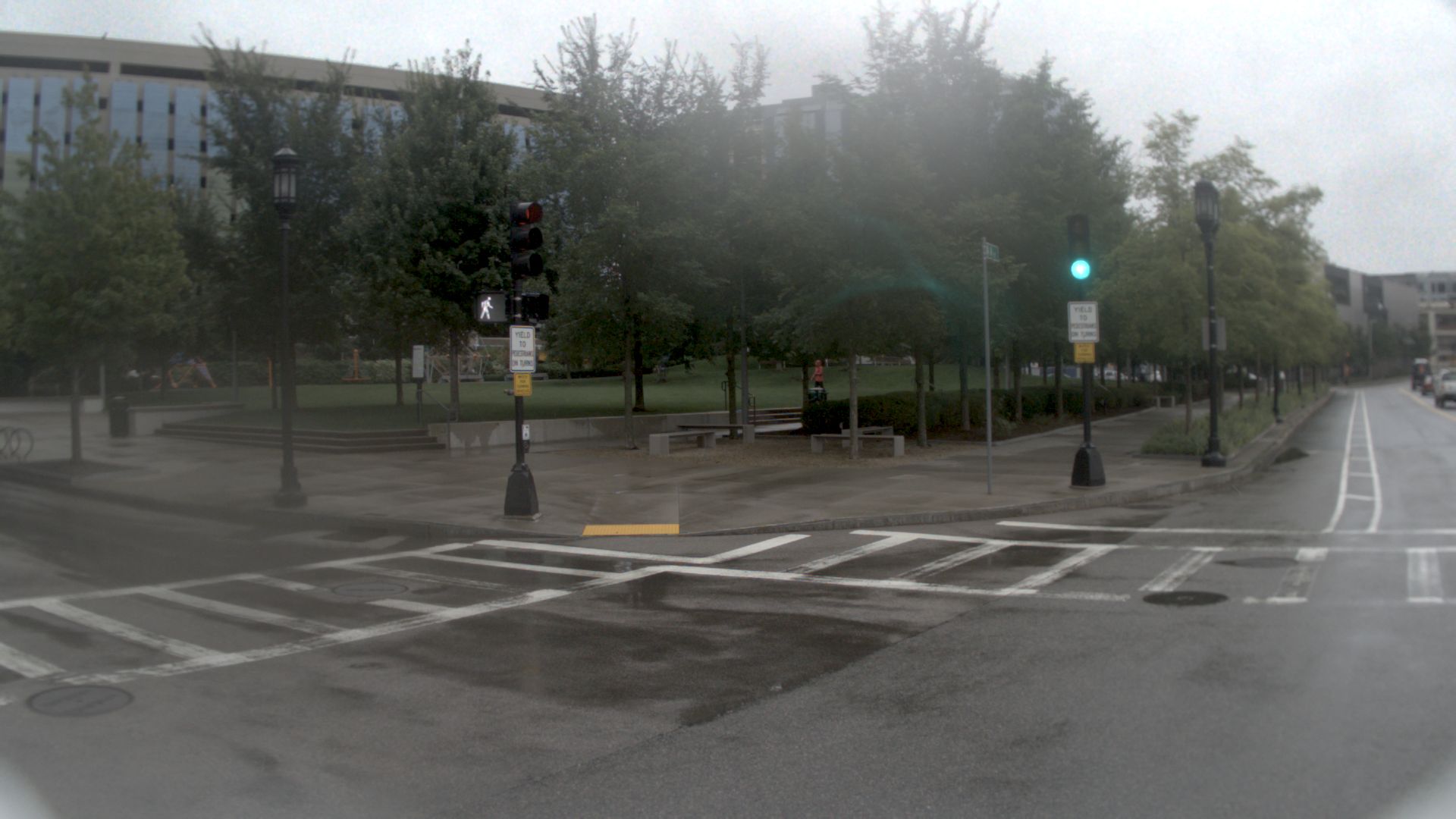}
\end{minipage}
\hspace{0.01\textwidth} % 将 \hfill 改为固定小间距
\begin{minipage}[b]{0.20\textwidth}
\centering
\textbf{Trajectory}
\vspace{0.2cm}
\includegraphics[
    height=4cm, 
    width=\linewidth, 
    keepaspectratio, 
    valign=b,
    trim={5 5 5 5},
    clip
]{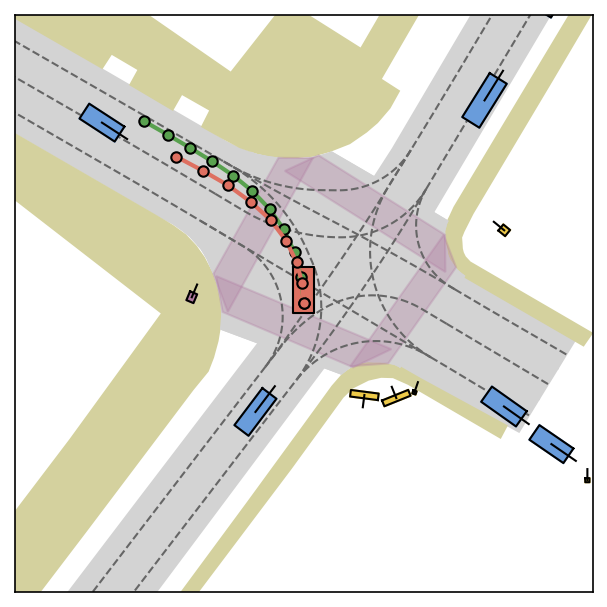}
\end{minipage}
\end{minipage}
% ========== 第二行：world model ======
\begin{minipage}[b]{0.04\textwidth}
\centering
\rotatebox{90}{\textbf{WM}}
\vspace{0.7cm}
\end{minipage}%
\hfill
\begin{minipage}[b]{0.96\textwidth}
\begin{minipage}[b]{0.23\textwidth}
\centering
\textbf{$V_{t+1}$}
\vspace{0.2cm}
\includegraphics[
    height=3.5cm, 
    width=\linewidth, 
    keepaspectratio, 
    valign=b,
    trim={0 0 0 0},
    clip
]{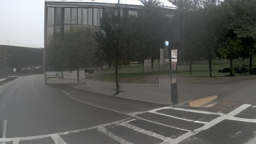}
\end{minipage}
\hspace{0.013\textwidth} % 将 \hfill 改为固定小间距
\begin{minipage}[b]{0.23\textwidth}
\centering
\textbf{$V_{t+2}$}
\vspace{0.2cm}
\includegraphics[
    height=3.5cm, 
    width=\linewidth, 
    keepaspectratio, 
    valign=b,
    trim={0 0 0 0},
    clip
]{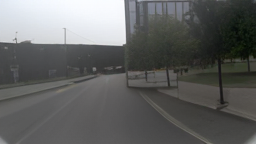}
\end{minipage}
\hspace{0.013\textwidth} % 将 \hfill 改为固定小间距
\begin{minipage}[b]{0.23\textwidth}
\centering
\textbf{$V_{t+3}$}
\vspace{0.2cm}
\includegraphics[
    height=3.5cm,
    width=\linewidth, 
    keepaspectratio, 
    valign=b,
    trim={0 0 0 0},
    clip
]{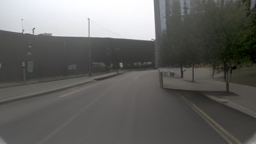}
\end{minipage}
\hspace{0.013\textwidth} % 将 \hfill 改为固定小间距
\begin{minipage}[b]{0.23\textwidth}
\centering
\textbf{$V_{t+4}$}
\vspace{0.2cm}
\includegraphics[
    height=3.5cm, 
    width=\linewidth, 
    keepaspectratio, 
    valign=b,
    trim={0 0 0 0},
    clip
]{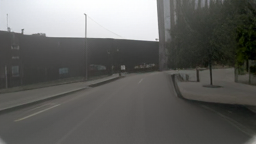}
\end{minipage}
\end{minipage}
% ========== 第三行：GT ======
\begin{minipage}[b]{0.04\textwidth}
\centering
\rotatebox{90}{\textbf{GT}}
\vspace{1.2cm}
\end{minipage}%
\hfill
\begin{minipage}[b]{0.96\textwidth}
\begin{minipage}[b]{0.23\textwidth}
\centering
\includegraphics[
    height=3.5cm, 
    width=\linewidth, 
    keepaspectratio, 
    valign=b,
    trim={0 0 0 0},
    clip
]{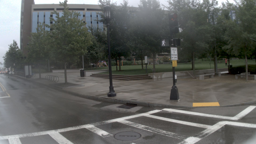}
\end{minipage}
\hspace{0.013\textwidth} % 将 \hfill 改为固定小间距
\begin{minipage}[b]{0.23\textwidth}
\centering
\includegraphics[
    height=3.5cm, 
    width=\linewidth, 
    keepaspectratio, 
    valign=b,
    trim={0 0 0 0},
    clip
]{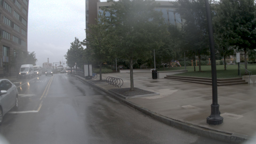}
\end{minipage}
\hspace{0.013\textwidth} % 将 \hfill 改为固定小间距
\begin{minipage}[b]{0.23\textwidth}
\centering
\includegraphics[
    height=3.5cm,
    width=\linewidth, 
    keepaspectratio, 
    valign=b,
    trim={0 0 0 0},
    clip
]{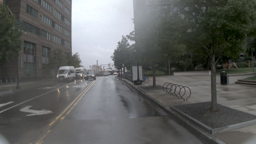}
\end{minipage}
\hspace{0.013\textwidth} % 将 \hfill 改为固定小间距
\begin{minipage}[b]{0.23\textwidth}
\centering
\includegraphics[
    height=3.5cm, 
    width=\linewidth, 
    keepaspectratio, 
    valign=b,
    trim={0 0 0 0},
    clip
]{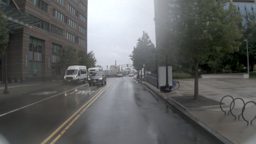}
\end{minipage} 
\vspace{0.2em}
\end{minipage}
% ========== 第四行：统一图例 ==========
\centering
\begin{tabular}{c @{\hspace{0.1em}} c @{\qquad} c @{\hspace{0.1em}} c @{\qquad} c @{\hspace{0.1em}} c @{\qquad} c @{\hspace{0.1em}} c @{\qquad} c @{\hspace{0.1em}} c}
  \textcolor{npccolor}{\rule{1.2em}{0.8em}} & NPC &
  \textcolor{egocolor}{\rule{1.2em}{0.8em}} & Ego Car &
  \textcolor{lanecolor}{\hdashrule[0.4ex]{1.6em}{0.05em}{2pt 2pt}} & Lane Center Line &
  \textcolor{gtcolor}{\Large$\bullet$} & GT Traj. &
  \textcolor{predcolor}{\Large$\bullet$} & Pred Traj. \\
\end{tabular}
% ========== 第五行：图题 ==========
\vspace{0.2em} % 与图例的间距
\caption{\textbf{Failure case analysis: constraints of front-view cameras under left turn.}
We present a failure case in which our model crosses the central double yellow line during left turns. Limited by field of view, historical and current frames fail to deliver valid cues for the target lane in the left-turn process; the world model’s predictions correspondingly lack the double road marking. This demonstrates the importance of multi-view camera inputs for turning scenarios, while trading off camera count against inference latency remains a key unresolved issue.}
\label{fig:bad_case_left_turn_samll_intersection}
\end{figure}

\begin{figure}[htbp]
\centering
% ===== 第一行：第一行图片 =====
\begin{minipage}[b]{0.04\textwidth}
\centering
\rotatebox{90}{\textbf{Planning}}
\vspace{0.8cm}
\end{minipage}%
\hfill
\begin{minipage}[b]{0.96\textwidth}
\begin{minipage}[b]{0.38\textwidth}
\centering
\textbf{$V_{t-1}$}
\vspace{0.2cm}
\includegraphics[
    height=3.5cm, 
    width=\linewidth, 
    keepaspectratio, 
    valign=b,
    trim={0 100 0 0},
    clip
]{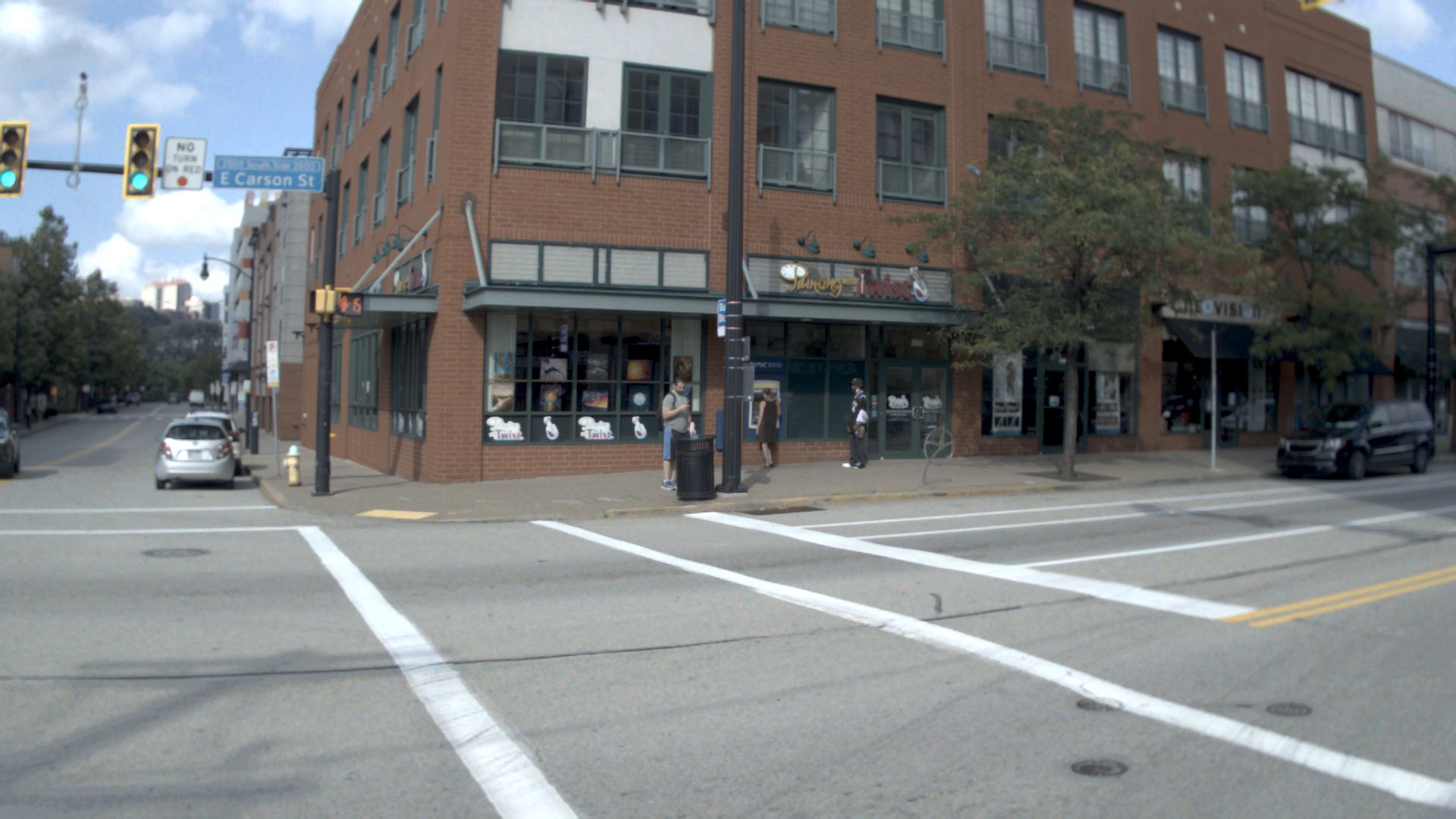}
\end{minipage}
\hspace{0.01\textwidth} % 将 \hfill 改为固定小间距
\begin{minipage}[b]{0.38\textwidth}
\centering
\textbf{$V_t$}
\vspace{0.2cm}
\includegraphics[
    height=3.5cm, 
    width=\linewidth, 
    keepaspectratio, 
    valign=b,
    trim={0 100 0 0},
    clip
]{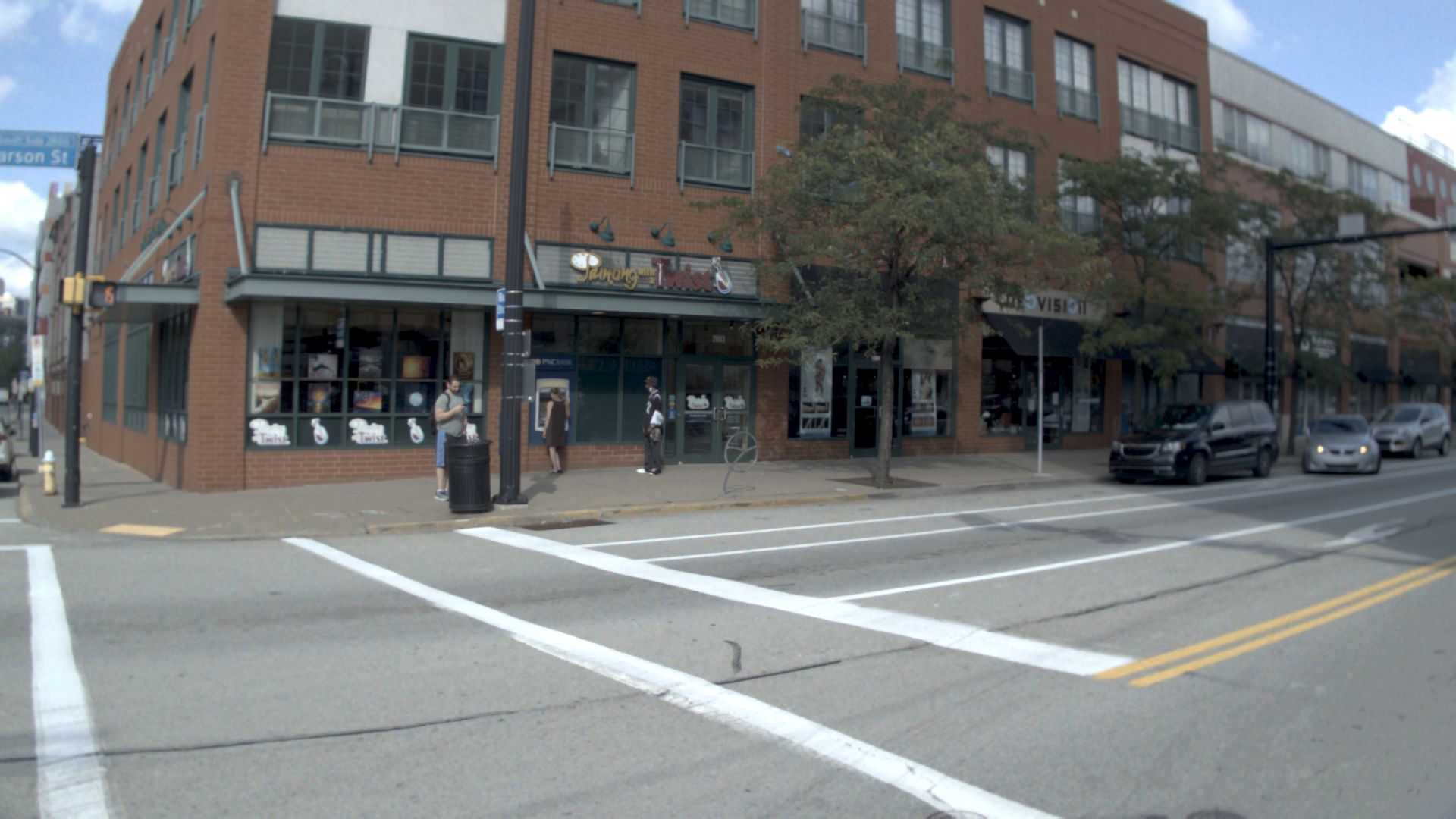}
\end{minipage}
\hspace{0.01\textwidth} % 将 \hfill 改为固定小间距
\begin{minipage}[b]{0.20\textwidth}
\centering
\textbf{Trajectory}
\vspace{0.2cm}
\includegraphics[
    height=4cm, 
    width=\linewidth, 
    keepaspectratio, 
    valign=b,
    trim={5 5 5 5},
    clip
]{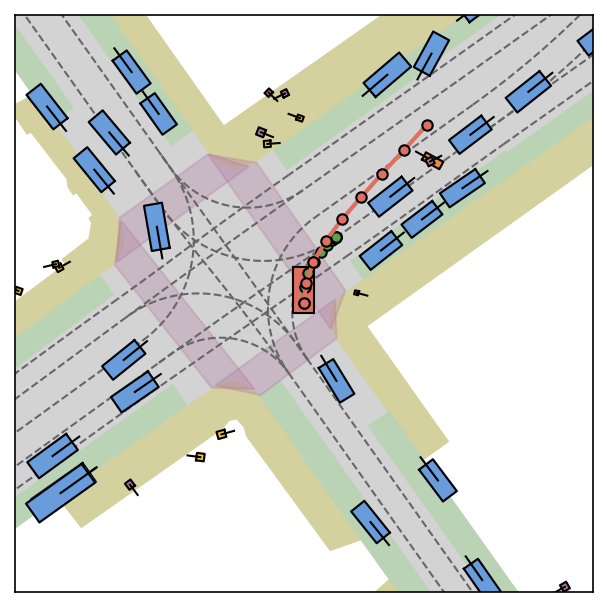}
\end{minipage}
\end{minipage}
% ========== 第二行：world model ======
\begin{minipage}[b]{0.04\textwidth}
\centering
\rotatebox{90}{\textbf{WM}}
\vspace{0.7cm}
\end{minipage}%
\hfill
\begin{minipage}[b]{0.96\textwidth}
\begin{minipage}[b]{0.23\textwidth}
\centering
\textbf{$V_{t+1}$}
\vspace{0.2cm}
\includegraphics[
    height=3.5cm, 
    width=\linewidth, 
    keepaspectratio, 
    valign=b,
    trim={0 0 0 0},
    clip
]{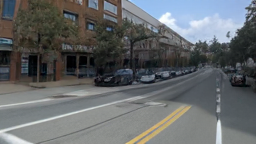}
\end{minipage}
\hspace{0.013\textwidth} % 将 \hfill 改为固定小间距
\begin{minipage}[b]{0.23\textwidth}
\centering
\textbf{$V_{t+2}$}
\vspace{0.2cm}
\includegraphics[
    height=3.5cm, 
    width=\linewidth, 
    keepaspectratio, 
    valign=b,
    trim={0 0 0 0},
    clip
]{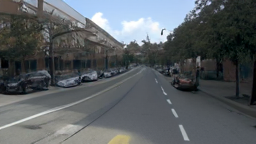}
\end{minipage}
\hspace{0.013\textwidth} % 将 \hfill 改为固定小间距
\begin{minipage}[b]{0.23\textwidth}
\centering
\textbf{$V_{t+3}$}
\vspace{0.2cm}
\includegraphics[
    height=3.5cm,
    width=\linewidth, 
    keepaspectratio, 
    valign=b,
    trim={0 0 0 0},
    clip
]{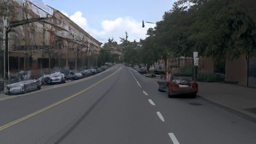}
\end{minipage}
\hspace{0.013\textwidth} % 将 \hfill 改为固定小间距
\begin{minipage}[b]{0.23\textwidth}
\centering
\textbf{$V_{t+4}$}
\vspace{0.2cm}
\includegraphics[
    height=3.5cm, 
    width=\linewidth, 
    keepaspectratio, 
    valign=b,
    trim={0 0 0 0},
    clip
]{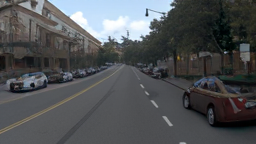}
\end{minipage}
\end{minipage}
% ========== 第三行：GT ======
\begin{minipage}[b]{0.04\textwidth}
\centering
\rotatebox{90}{\textbf{GT}}
\vspace{1.2cm}
\end{minipage}%
\hfill
\begin{minipage}[b]{0.96\textwidth}
\begin{minipage}[b]{0.23\textwidth}
\centering
\includegraphics[
    height=3.5cm, 
    width=\linewidth, 
    keepaspectratio, 
    valign=b,
    trim={0 0 0 0},
    clip
]{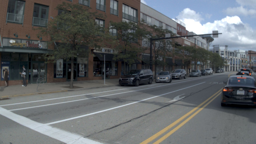}
\end{minipage}
\hspace{0.013\textwidth} % 将 \hfill 改为固定小间距
\begin{minipage}[b]{0.23\textwidth}
\centering
\includegraphics[
    height=3.5cm, 
    width=\linewidth, 
    keepaspectratio, 
    valign=b,
    trim={0 0 0 0},
    clip
]{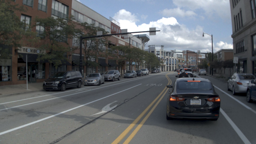}
\end{minipage}
\hspace{0.013\textwidth} % 将 \hfill 改为固定小间距
\begin{minipage}[b]{0.23\textwidth}
\centering
\includegraphics[
    height=3.5cm,
    width=\linewidth, 
    keepaspectratio, 
    valign=b,
    trim={0 0 0 0},
    clip
]{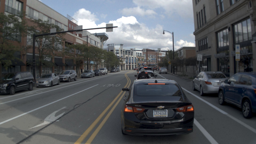}
\end{minipage}
\hspace{0.013\textwidth} % 将 \hfill 改为固定小间距
\begin{minipage}[b]{0.23\textwidth}
\centering
\includegraphics[
    height=3.5cm, 
    width=\linewidth, 
    keepaspectratio, 
    valign=b,
    trim={0 0 0 0},
    clip
]{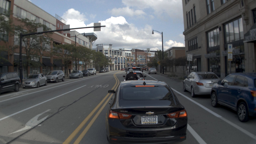}
\end{minipage} 
\vspace{0.2em}
\end{minipage}
% ========== 第四行：统一图例 ==========
\centering
\begin{tabular}{c @{\hspace{0.1em}} c @{\qquad} c @{\hspace{0.1em}} c @{\qquad} c @{\hspace{0.1em}} c @{\qquad} c @{\hspace{0.1em}} c @{\qquad} c @{\hspace{0.1em}} c}
  \textcolor{npccolor}{\rule{1.2em}{0.8em}} & NPC &
  \textcolor{egocolor}{\rule{1.2em}{0.8em}} & Ego Car &
  \textcolor{lanecolor}{\hdashrule[0.4ex]{1.6em}{0.05em}{2pt 2pt}} & Lane Center Line &
  \textcolor{gtcolor}{\Large$\bullet$} & GT Traj. &
  \textcolor{predcolor}{\Large$\bullet$} & Pred Traj. \\
\end{tabular}
% ========== 第五行：图题 ==========
\vspace{0.2em} % 与图例的间距
\caption{\textbf{Failure case analysis: constraints of front-view cameras under right turn.}
We present a failure case in which our model rear-ends a queuing leading vehicle during right turns. Limited by field of view, valid cues of surrounding vehicles are unavailable in the right-turn process, and the world model’s predictions correspondingly miss the preceding vehicle. This again demonstrates the necessity of multi-view camera inputs for turning scenarios.}
\label{fig:bad_case_right_turn_samll_intersection}
\end{figure}

To provide an unbiased account of our model’s performance and delineate paths for subsequent enhancement, representative failure cases encountered in evaluation are analyzed. They fall into two main types: deficits in capturing critical small elements, and constraints of front-view cameras under turning scenarios, as shown in Figure~\ref{fig:bad_case_nudge}, Figure~\ref{fig:bad_case_left_turn_samll_intersection} and Figure~\ref{fig:bad_case_right_turn_samll_intersection}.
% =======================================
% =====appendix failed case fig 2========
% =======================================

% =======================================
% =====appendix failed case fig 5========
% =======================================

% =======================================
% =====appendix failed case fig 6========
% =======================================

\subsection{More Scene Noise Robustness Case}
\begin{figure}[!t]
\centering
\setlength{\abovecaptionskip}{2pt}
\setlength{\belowcaptionskip}{0pt}
\begin{subfigure}{\textwidth}
    % ===== 第一行 =====
    \begin{minipage}[b]{0.2\textwidth}
    \centering
    \rotatebox{90}{\textbf{Light}}
    \vspace{0.8cm}
    \end{minipage}%
    \hspace{-1.7cm}%
    \begin{minipage}[b]{0.9\textwidth}
    \centering
    \begin{minipage}[b]{0.43\textwidth}
    \centering
    \textbf{Front-view Image}
    \includegraphics[height=3.0cm,width=\linewidth,keepaspectratio,valign=b,trim={0 100 0 0},clip]{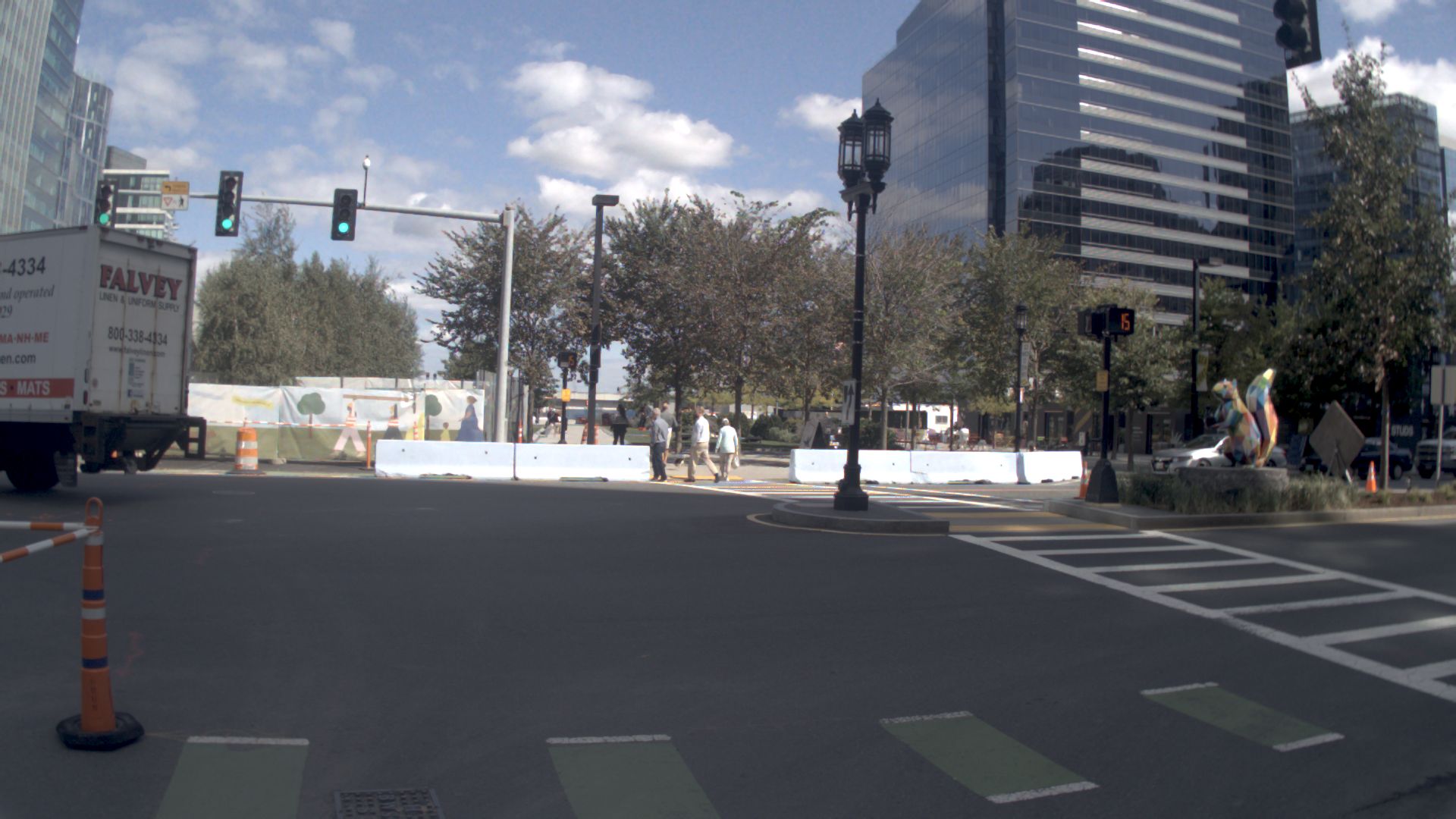}
    \end{minipage}
    \hspace{0.01\textwidth}%  % ← 可控的横向间距
    \begin{minipage}[b]{0.22\textwidth}
    \centering
    \textbf{Ours}
    \includegraphics[height=3.0cm,width=\linewidth,keepaspectratio,valign=b,trim={5 5 5 5},clip]{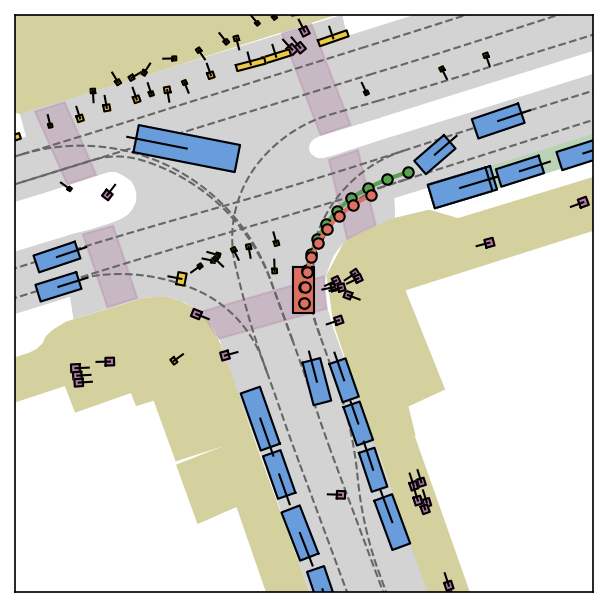}
    \end{minipage}
    \hspace{0.01\textwidth}%  % ← 可控的横向间距
    \begin{minipage}[b]{0.22\textwidth}
    \centering
    \textbf{DriveVLA-W0}
    \includegraphics[height=3.0cm,width=\linewidth,keepaspectratio,valign=b,trim={5 5 5 5},clip]{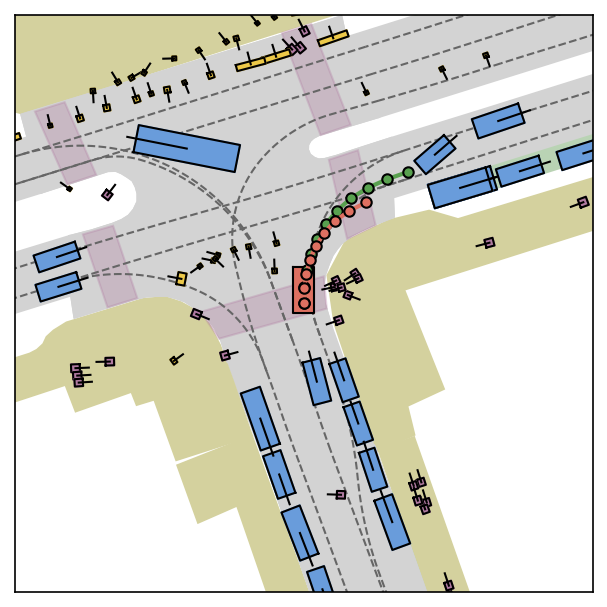}
    \end{minipage}
    \end{minipage}
    % ===== 第二行 =====
    \begin{minipage}[b]{0.2\textwidth}
    \centering
    \rotatebox{90}{\textbf{Dark}}
    \vspace{0.8cm}
    \end{minipage}%
    \hspace{-1.7cm}%
    \begin{minipage}[b]{0.9\textwidth}
    \centering
    \begin{minipage}[b]{0.43\textwidth}
    \centering
    \includegraphics[height=3.0cm,width=\linewidth,keepaspectratio,valign=b,trim={0 150 0 0},clip]{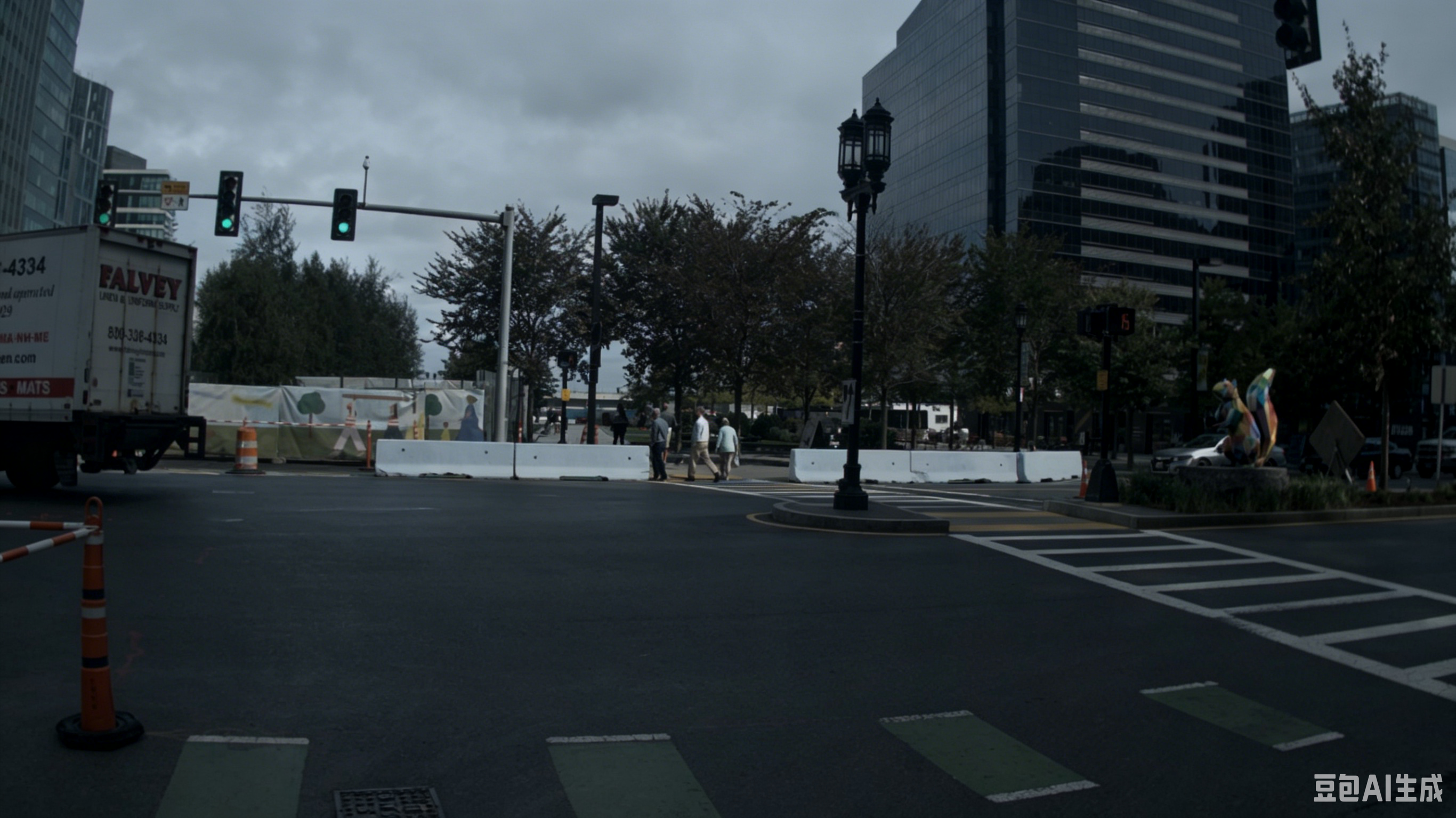}
    \end{minipage}
    \hspace{0.01\textwidth}%  % ← 可控的横向间距
    \begin{minipage}[b]{0.22\textwidth}
    \centering
    \includegraphics[height=3.0cm,width=\linewidth,keepaspectratio,valign=b,trim={5 5 5 5},clip]{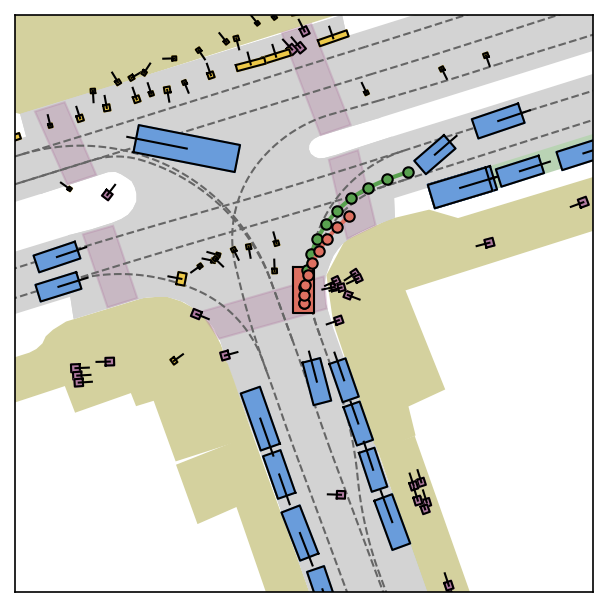}
    \end{minipage}
    \hspace{0.01\textwidth}%  % ← 可控的横向间距
    \begin{minipage}[b]{0.22\textwidth}
    \centering
    \includegraphics[height=3.0cm,width=\linewidth,keepaspectratio,valign=b,trim={5 5 5 5},clip]{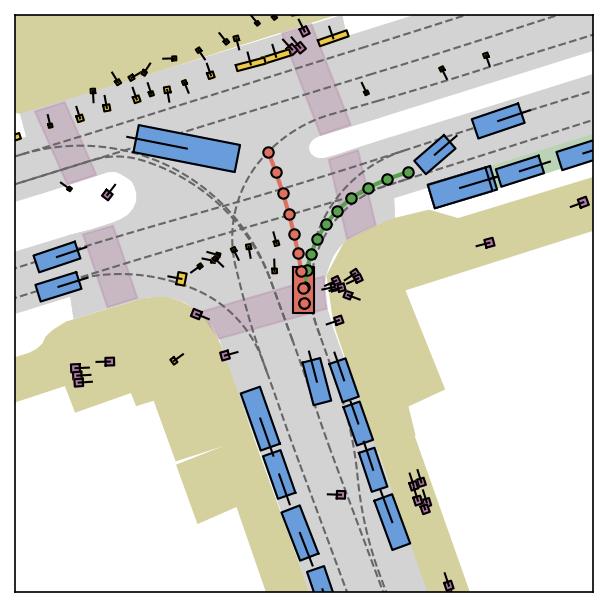}
    \end{minipage}
    \end{minipage}
    \caption{Turn right in the intersection.}
    \label{fig:robustness_to_light_intensity_right_turn}
\end{subfigure}
\begin{subfigure}{\textwidth}
    % ===== 第三行 =====
    \begin{minipage}[b]{0.2\textwidth}
    \centering
    \rotatebox{90}{\textbf{Light}}
    \vspace{0.8cm}
    \end{minipage}%
    \hspace{-1.7cm}%
    \begin{minipage}[b]{0.9\textwidth}
    \centering
    \begin{minipage}[b]{0.43\textwidth}
    \centering
    \includegraphics[height=3.0cm,width=\linewidth,keepaspectratio,valign=b,trim={0 100 0 0},clip]{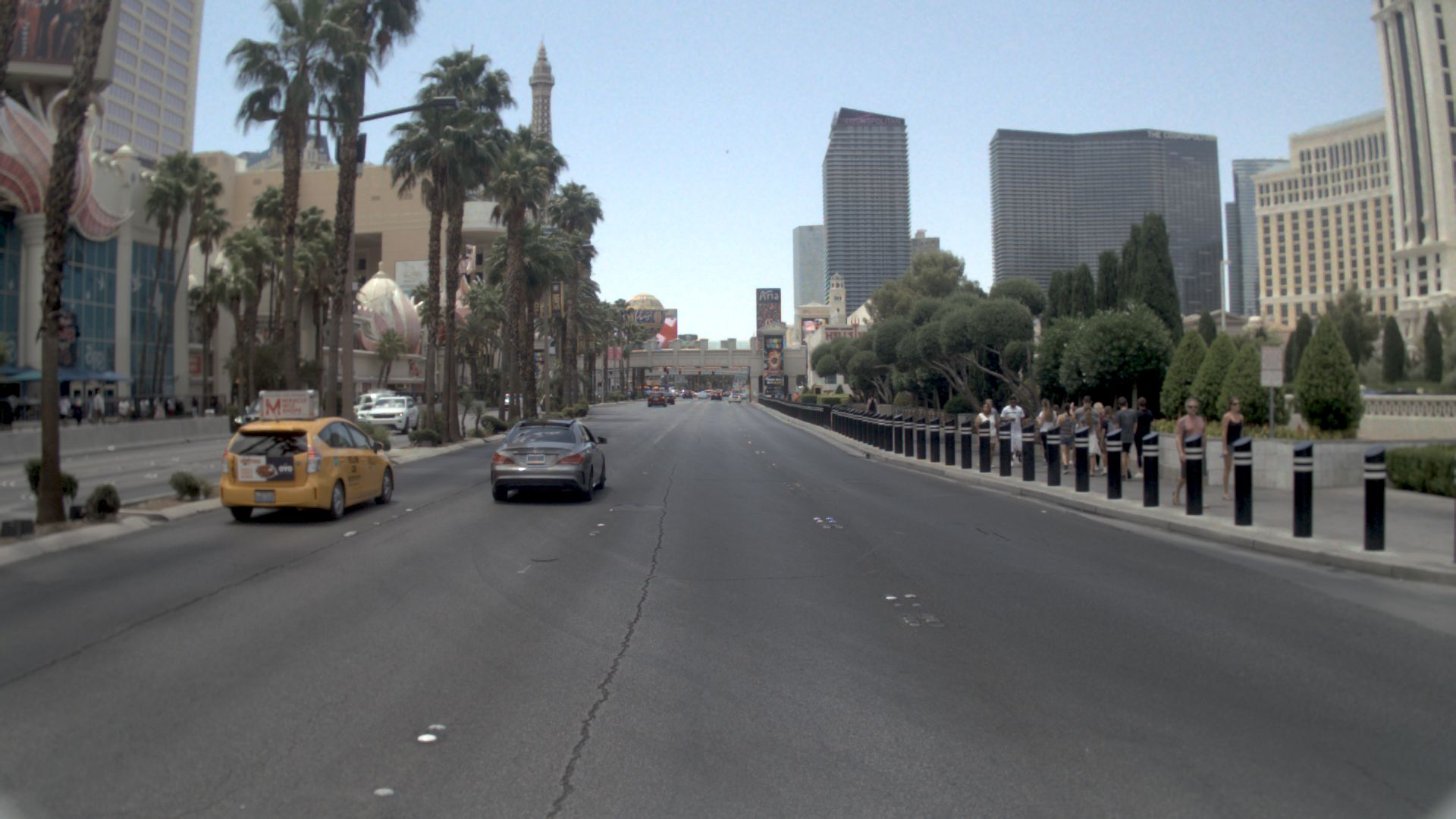}
    \end{minipage}
    \hspace{0.01\textwidth}%  % ← 可控的横向间距
    \begin{minipage}[b]{0.22\textwidth}
    \centering
    \includegraphics[height=3.0cm,width=\linewidth,keepaspectratio,valign=b,trim={5 5 5 5},clip]{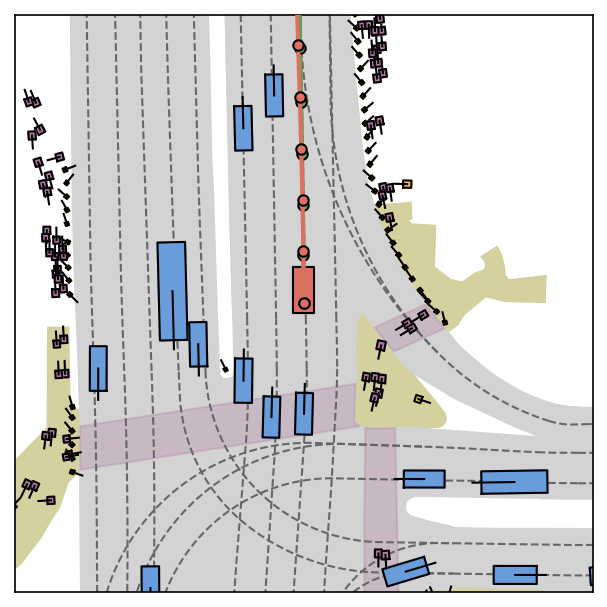}
    \end{minipage}
    \hspace{0.01\textwidth}%  % ← 可控的横向间距
    \begin{minipage}[b]{0.22\textwidth}
    \centering
    \includegraphics[height=3.0cm,width=\linewidth,keepaspectratio,valign=b,trim={5 5 5 5},clip]{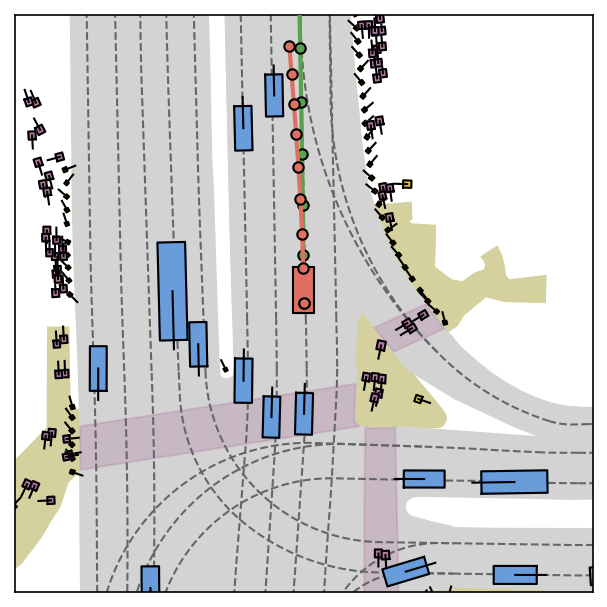}
    \end{minipage}
    \end{minipage}
    % ===== 第四行 =====
    \begin{minipage}[b]{0.2\textwidth}
    \centering
    \rotatebox{90}{\textbf{Dark}}
    \vspace{0.8cm}
    \end{minipage}%
    \hspace{-1.7cm}%
    \begin{minipage}[b]{0.9\textwidth}
    \centering
    \begin{minipage}[b]{0.43\textwidth}
    \centering
    \includegraphics[height=3.0cm,width=\linewidth,keepaspectratio,valign=b,trim={0 150 0 0},clip]{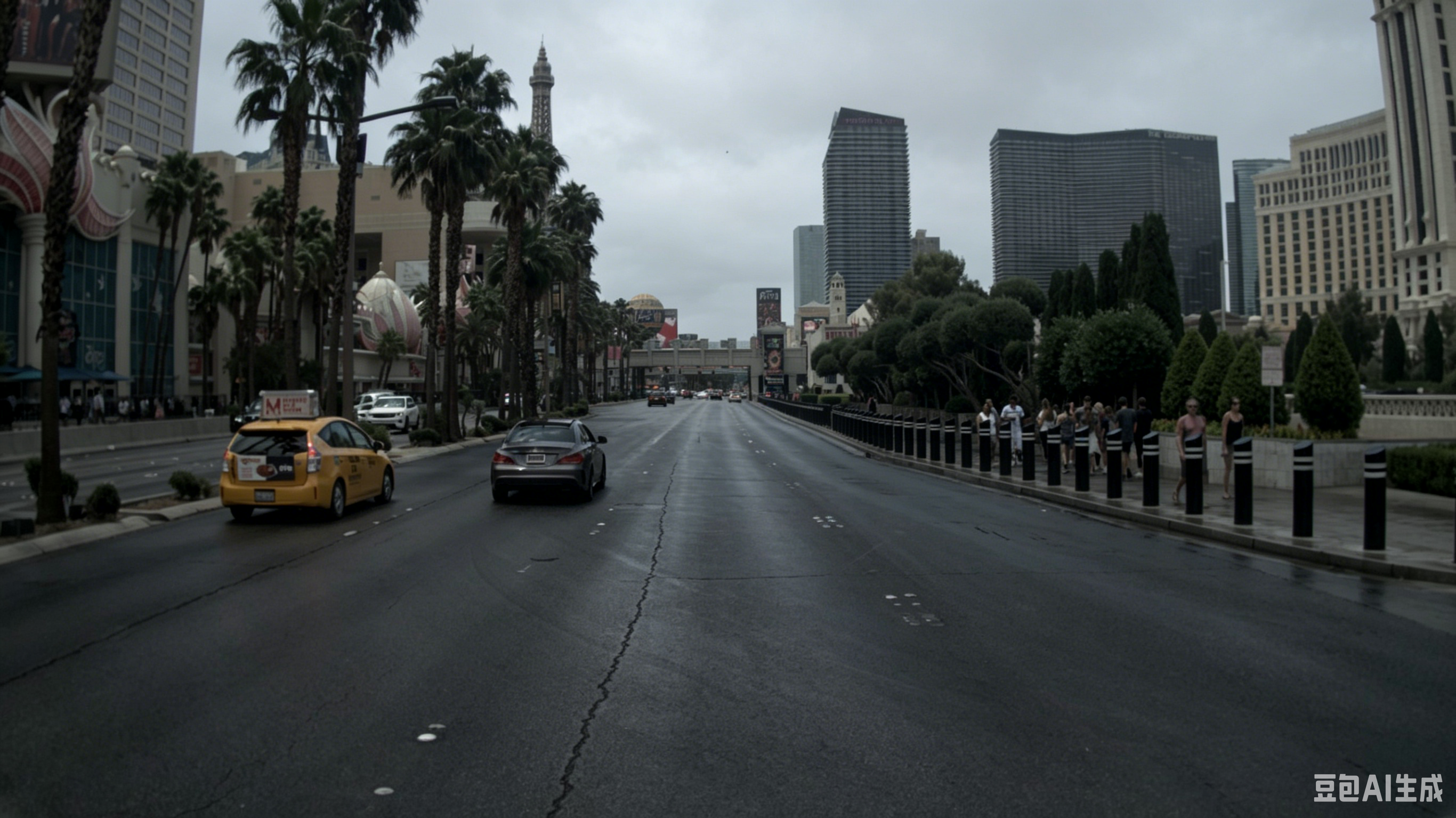}
    \end{minipage}
    \hspace{0.01\textwidth}%  % ← 可控的横向间距
    \begin{minipage}[b]{0.22\textwidth}
    \centering
    \includegraphics[height=3.0cm,width=\linewidth,keepaspectratio,valign=b,trim={5 5 5 5},clip]{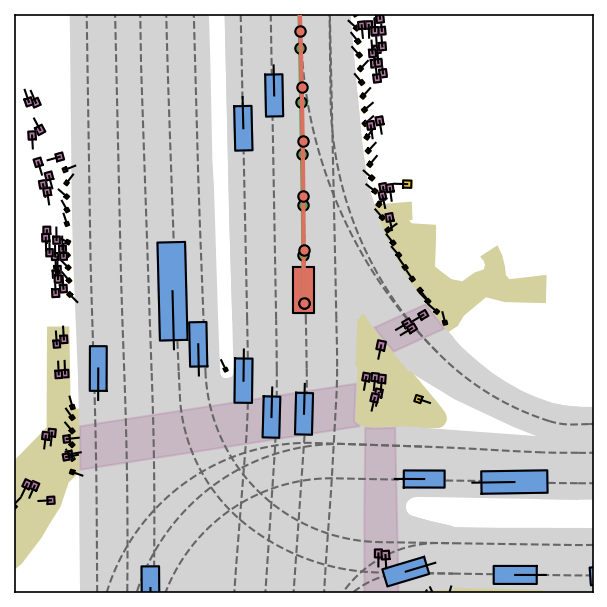}
    \end{minipage}
    \hspace{0.01\textwidth}%  % ← 可控的横向间距
    \begin{minipage}[b]{0.22\textwidth}
    \centering
    \includegraphics[height=3.0cm,width=\linewidth,keepaspectratio,valign=b,trim={5 5 5 5},clip]{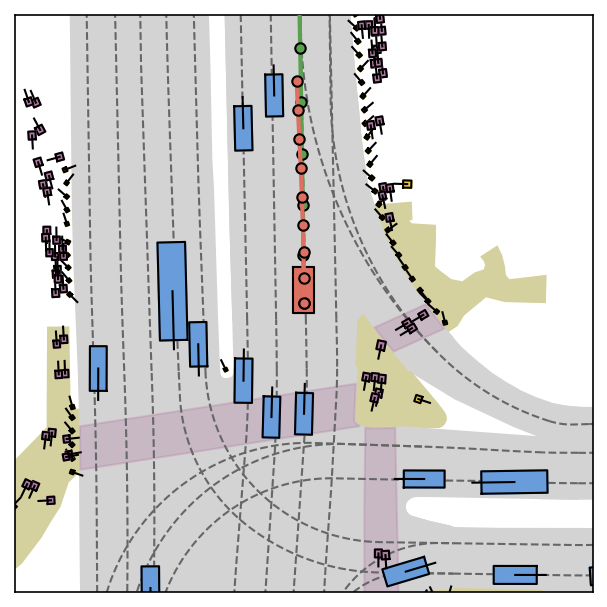}
    \end{minipage}
    \end{minipage}
    \caption{Go straight in the intersection.}
    \label{fig:robustness_to_light_intensity_go_straight}
\end{subfigure}
\vspace{0.1cm} % 与图例的间距
% ========== 图例 ==========
\centering
\begin{tabular}{c @{\hspace{0.4em}} c @{\qquad} c @{\hspace{0.4em}} c @{\qquad} c @{\hspace{0.4em}} c @{\qquad} c @{\hspace{0.4em}} c @{\qquad} c @{\hspace{0.4em}} c}
  \textcolor{npccolor}{\rule{1.2em}{0.8em}} & NPC &
  \textcolor{egocolor}{\rule{1.2em}{0.8em}} & Ego Car &
  \textcolor{lanecolor}{\hdashrule[0.4ex]{1.6em}{0.05em}{2pt 2pt}} & Lane Center Line &
  \textcolor{gtcolor}{\Large$\bullet$} & GT Traj. &
  \textcolor{predcolor}{\Large$\bullet$} & Pred Traj. \\
\end{tabular}
\vspace{0.1cm} % 与图例的间距
\caption{\textbf{Turn right and go straight under the fluctuating illumination.} (a) DriveVLA-W0 goes straight under right turn instruction, while our method remains robust. (b) DriveVLA-W0's behavior changes from changing lanes to going straight, while our approach remains going straight.}
\label{fig:robustness_to_light_intensity}
\end{figure}

\begin{figure}[!t]
\centering
\setlength{\abovecaptionskip}{2pt}
\setlength{\belowcaptionskip}{0pt}
% ===== 第一行 =====
\hspace*{1.8cm}%
\begin{minipage}[b]{0.9\textwidth}
\begin{minipage}[b]{0.43\textwidth}
\centering
\textbf{Front-view Image}
\includegraphics[height=3.0cm,width=\linewidth,keepaspectratio,valign=b,trim={0 100 0 0},clip]{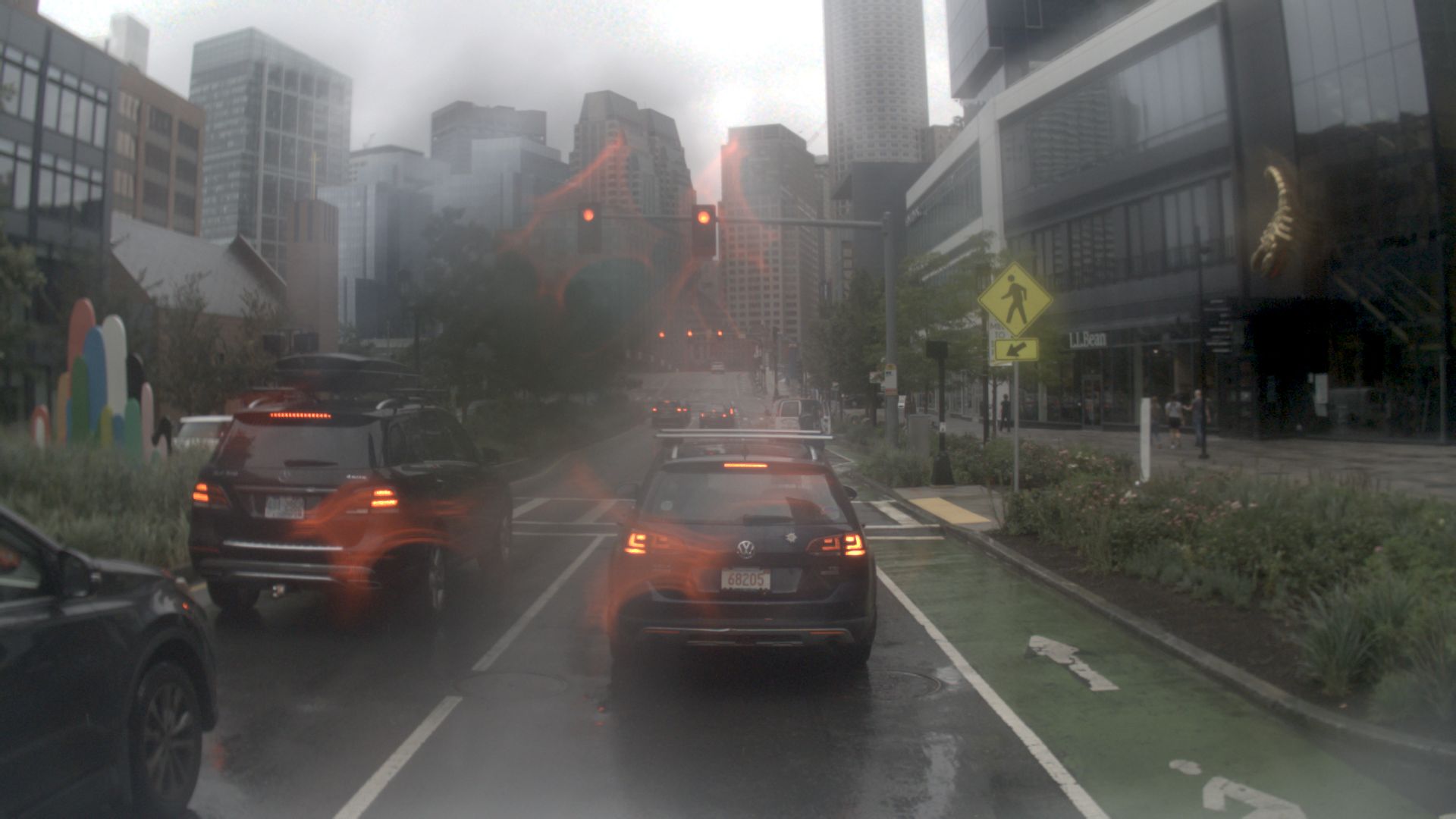}
\end{minipage}
\hspace*{0.1cm}%
\begin{minipage}[b]{0.22\textwidth}
\centering
\textbf{Ours}
\includegraphics[height=3.0cm,width=\linewidth,keepaspectratio,valign=b,trim={5 5 5 5},clip]{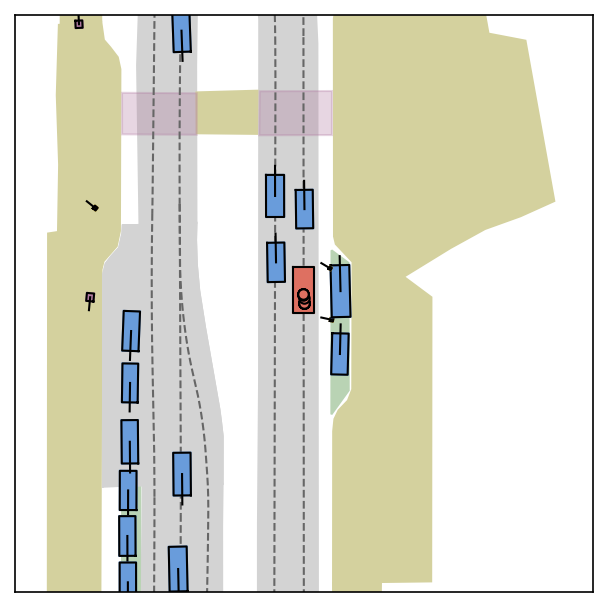}
\end{minipage}
\hspace*{0.1cm}%
\begin{minipage}[b]{0.22\textwidth}
\centering
\textbf{DriveVLA-W0}
\includegraphics[height=3.0cm,width=\linewidth,keepaspectratio,valign=b,trim={5 5 5 5},clip]{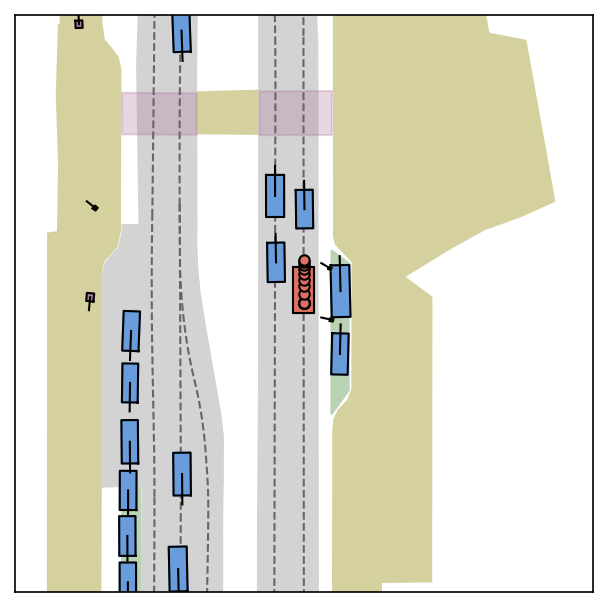}
\end{minipage}
\end{minipage}
% ===== 第二行 =====
\hspace*{1.8cm}%
\begin{minipage}[b]{0.9\textwidth}
\begin{minipage}[b]{0.43\textwidth}
\centering
\includegraphics[height=3.0cm,width=\linewidth,keepaspectratio,valign=b,trim={0 100 0 0},clip]{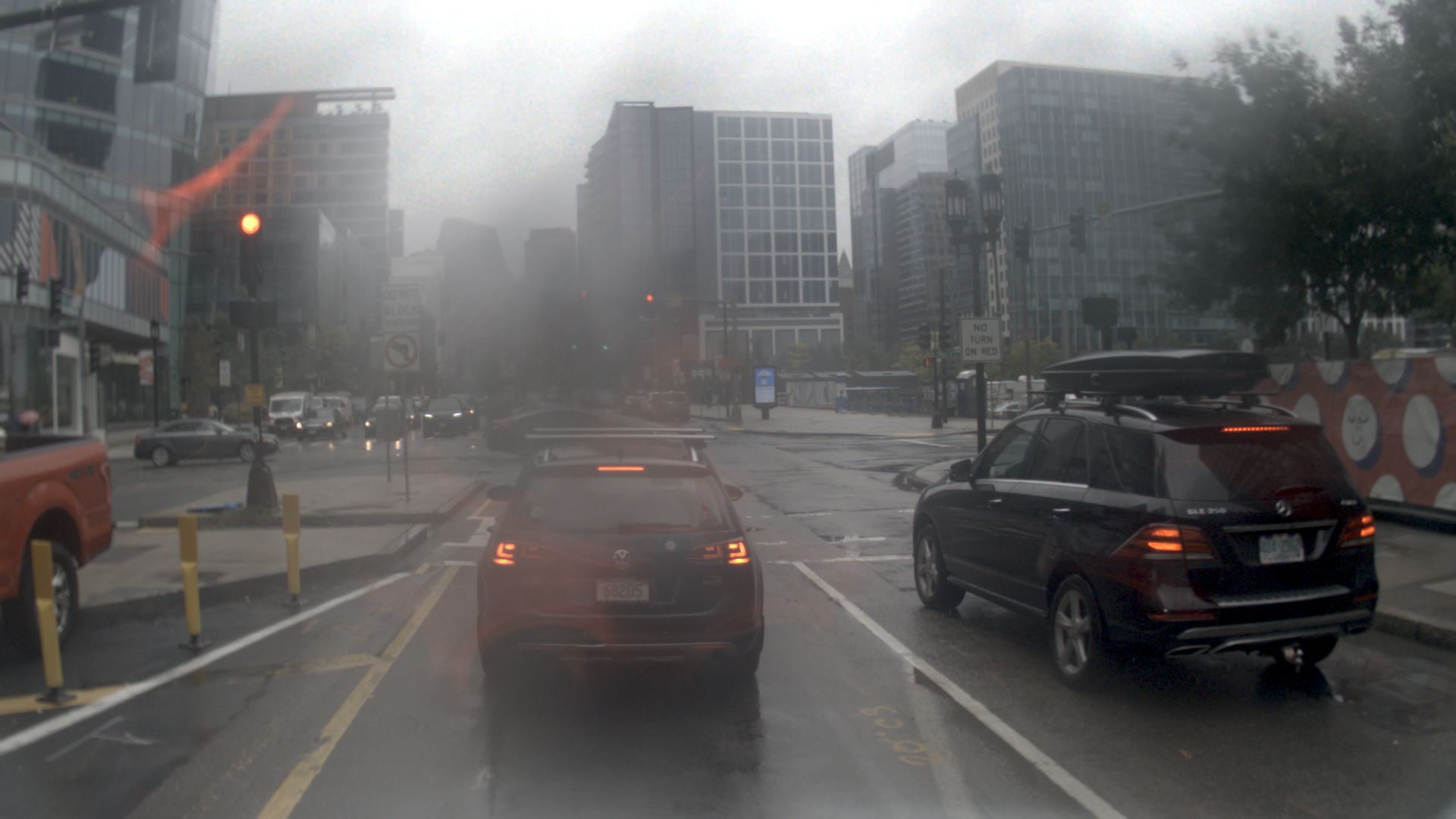}
\end{minipage}
\hspace*{0.1cm}%
\begin{minipage}[b]{0.22\textwidth}
\centering
\includegraphics[height=3.0cm,width=\linewidth,keepaspectratio,valign=b,trim={5 5 5 5},clip]{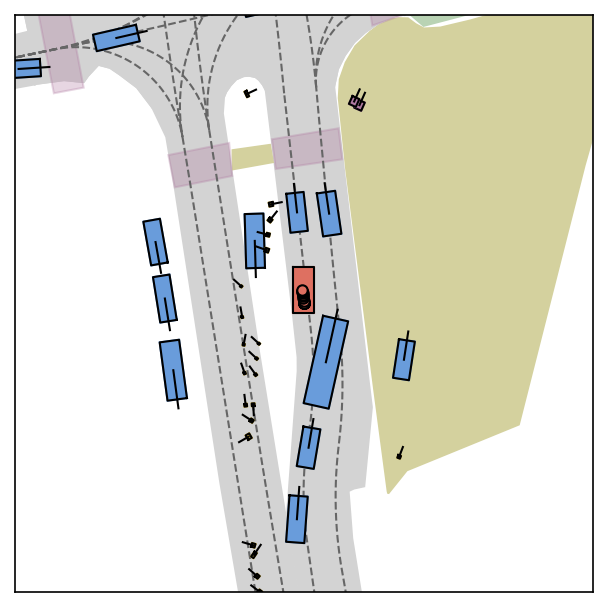}
\end{minipage}
\hspace*{0.1cm}%
\begin{minipage}[b]{0.22\textwidth}
\centering
\includegraphics[height=3.0cm,width=\linewidth,keepaspectratio,valign=b,trim={5 5 5 5},clip]{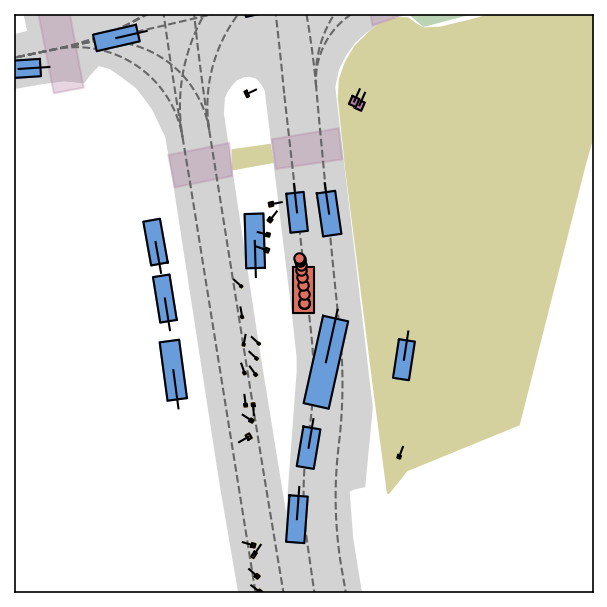}
\end{minipage}
\vspace{0.1cm} % 与图例的间距
\end{minipage}
% ========== 图例 ==========
\centering
\begin{tabular}{c @{\hspace{0.4em}} c @{\qquad} c @{\hspace{0.4em}} c @{\qquad} c @{\hspace{0.4em}} c @{\qquad} c @{\hspace{0.4em}} c @{\qquad} c @{\hspace{0.4em}} c}
  \textcolor{npccolor}{\rule{1.2em}{0.8em}} & NPC &
  \textcolor{egocolor}{\rule{1.2em}{0.8em}} & Ego Car &
  \textcolor{lanecolor}{\hdashrule[0.4ex]{1.6em}{0.05em}{2pt 2pt}} & Lane Center Line &
  \textcolor{gtcolor}{\Large$\bullet$} & GT Traj. &
  \textcolor{predcolor}{\Large$\bullet$} & Pred Traj. \\
\end{tabular}
\vspace{0.1cm} % 与图例的间距
\caption{\textbf{Remain stationary at red-light intersection under the blur noise.} DriveVLA-W0 baseline crawls during red-light queuing, while our method remains stationary.}
\label{fig:robustness_red_light_stop}
\end{figure}

\begin{figure}[htbp]
\centering
\setlength{\abovecaptionskip}{2pt}
\setlength{\belowcaptionskip}{0pt}
% ===== 第一行：第一行图片 =====
\begin{minipage}[b]{0.44\textwidth}
\centering
\textbf{Front-view Image}
\vspace{0.1cm}
\includegraphics[
    height=3.0cm,
    width=\linewidth,
    keepaspectratio,
    valign=b,
    trim={0 150 0 0},
    clip
]{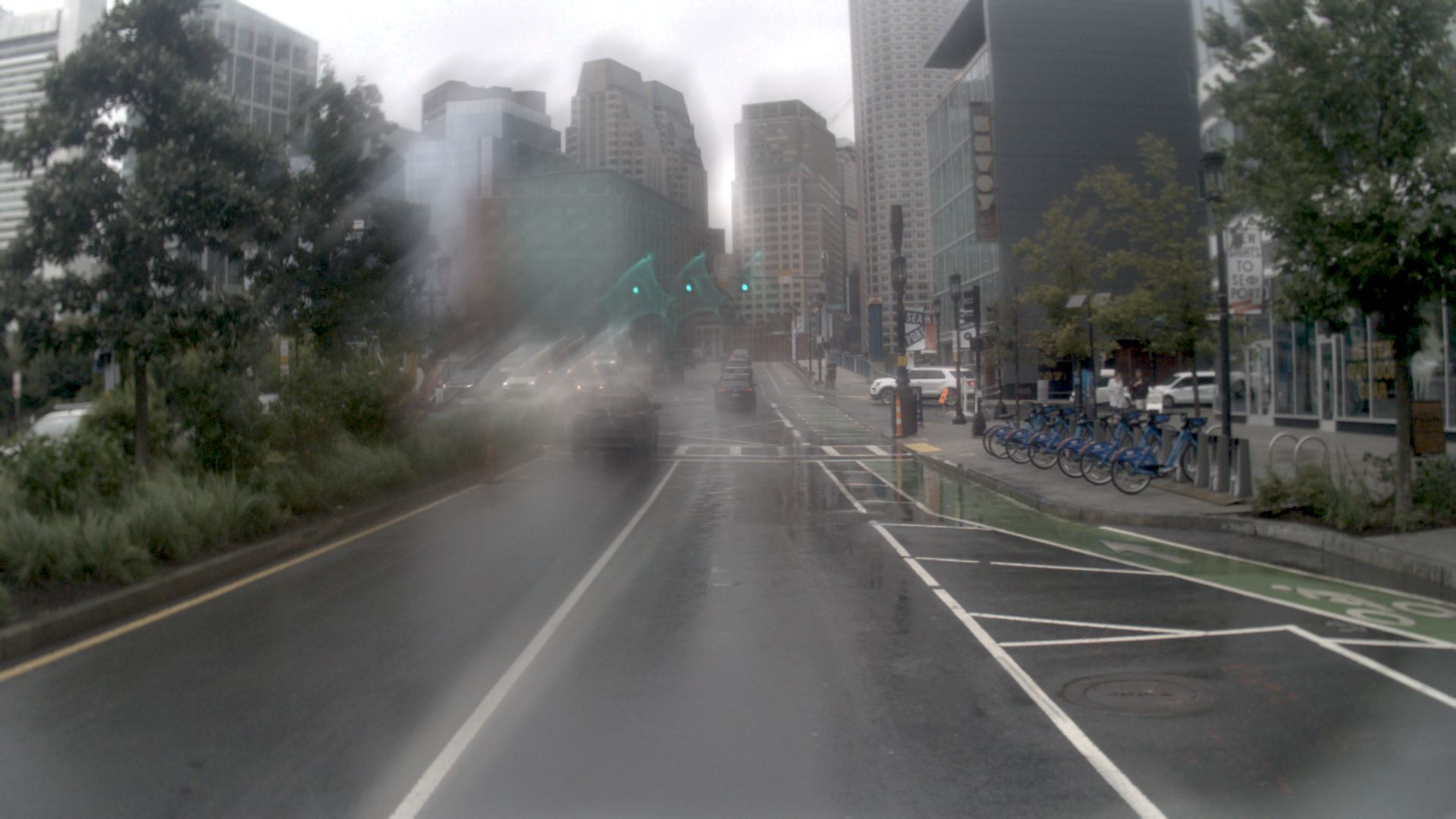}
\end{minipage}
\hspace{0.01\textwidth}%  % ← 可控的横向间距
\begin{minipage}[b]{0.22\textwidth}
\centering
\textbf{Ours}
\vspace{0.1cm}
\includegraphics[
    height=3.0cm,
    width=\linewidth,
    keepaspectratio,
    valign=b,
    trim={5 5 5 5},
    clip
]{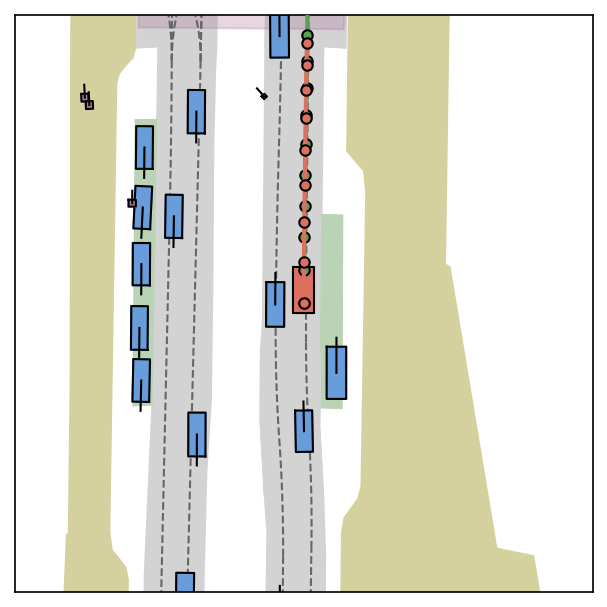}
\end{minipage}
\hspace{0.01\textwidth}%  % ← 可控的横向间距
\begin{minipage}[b]{0.22\textwidth}
\centering
\textbf{DriveVLA-W0}
\vspace{0.1cm}
\includegraphics[
    height=3.0cm,
    width=\linewidth,
    keepaspectratio,
    valign=b,
    trim={5 5 5 5},
    clip
]{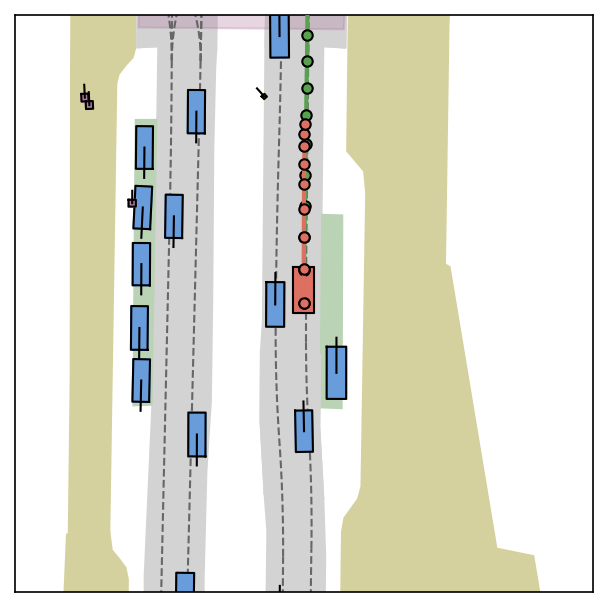}
\end{minipage}
% ===== 第二行：第二行图片 =====
\begin{minipage}[b]{0.44\textwidth}
\centering
\vspace{0.1cm}
\includegraphics[
    height=3.0cm,
    width=\linewidth,
    keepaspectratio,
    valign=b,
    trim={0 150 0 0},
    clip
]{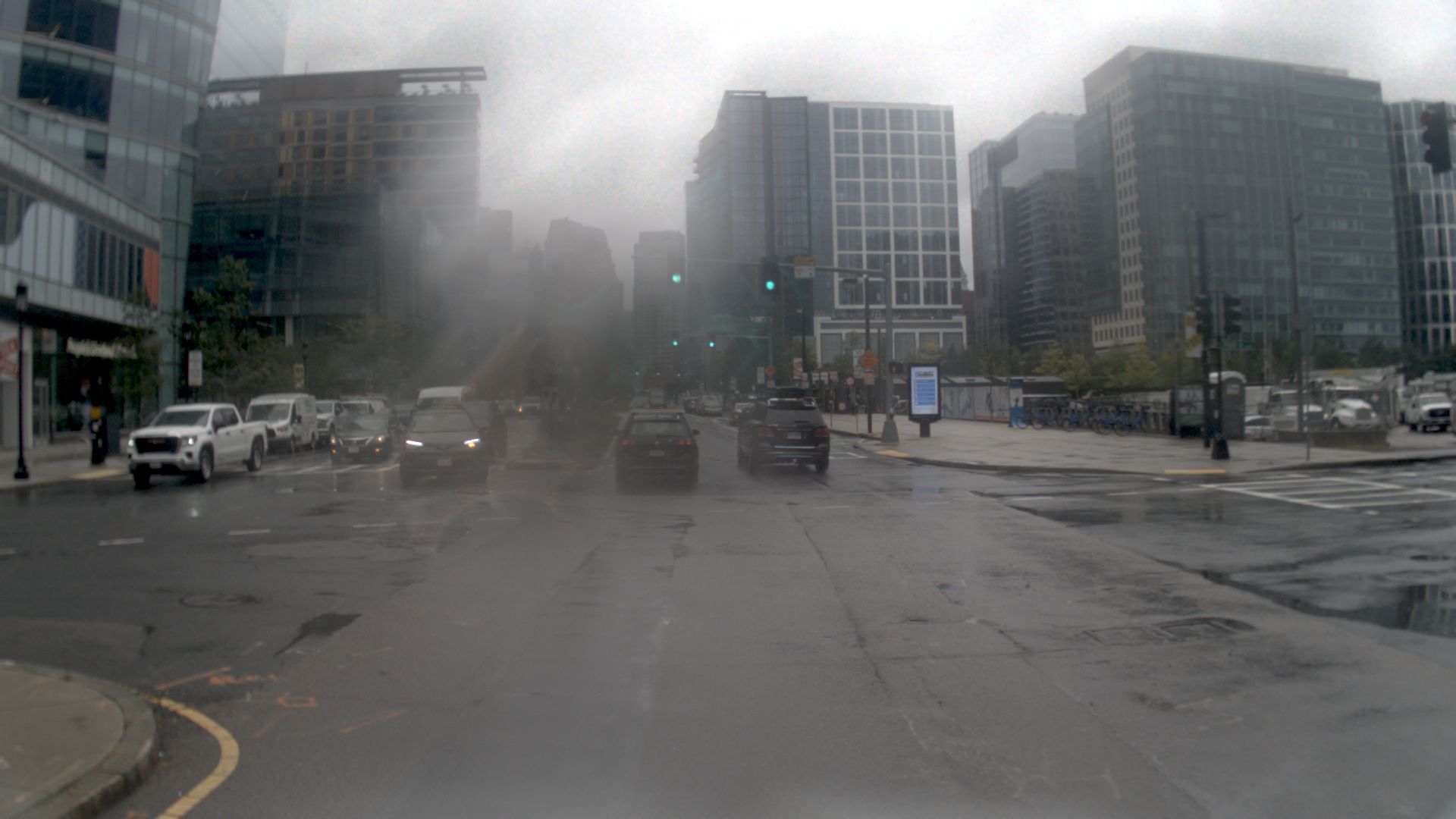}
\end{minipage}
\hspace{0.01\textwidth}%  % ← 可控的横向间距
\begin{minipage}[b]{0.22\textwidth}
\centering
\vspace{0.1cm}
\includegraphics[
    height=3.0cm,
    width=\linewidth,
    keepaspectratio,
    valign=b,
    trim={5 5 5 5},
    clip
]{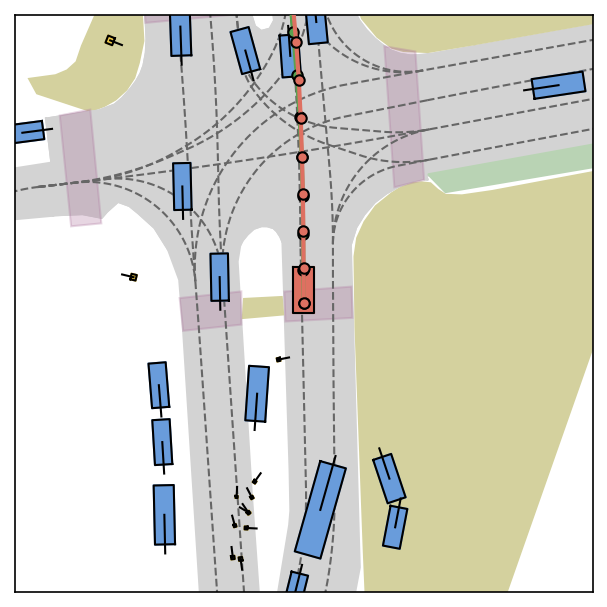}
\end{minipage}
\hspace{0.01\textwidth}%  % ← 可控的横向间距
\begin{minipage}[b]{0.22\textwidth}
\centering
\vspace{0.1cm}
\includegraphics[
    height=3.0cm,
    width=\linewidth,
    keepaspectratio,
    valign=b,
    trim={5 5 5 5},
    clip
]{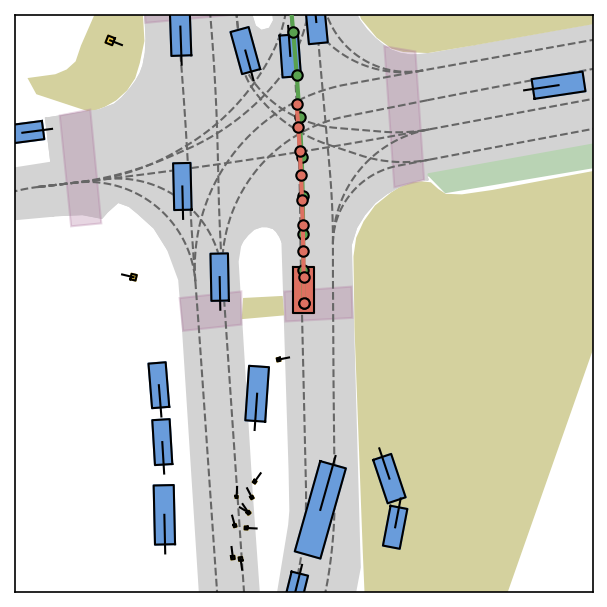}
\end{minipage}
% ===== 第三行：第三行图片 =====
\begin{minipage}[b]{0.44\textwidth}
\centering
\vspace{0.1cm}
\includegraphics[
    height=3.0cm,
    width=\linewidth,
    keepaspectratio,
    valign=b,
    trim={0 150 0 0},
    clip
]{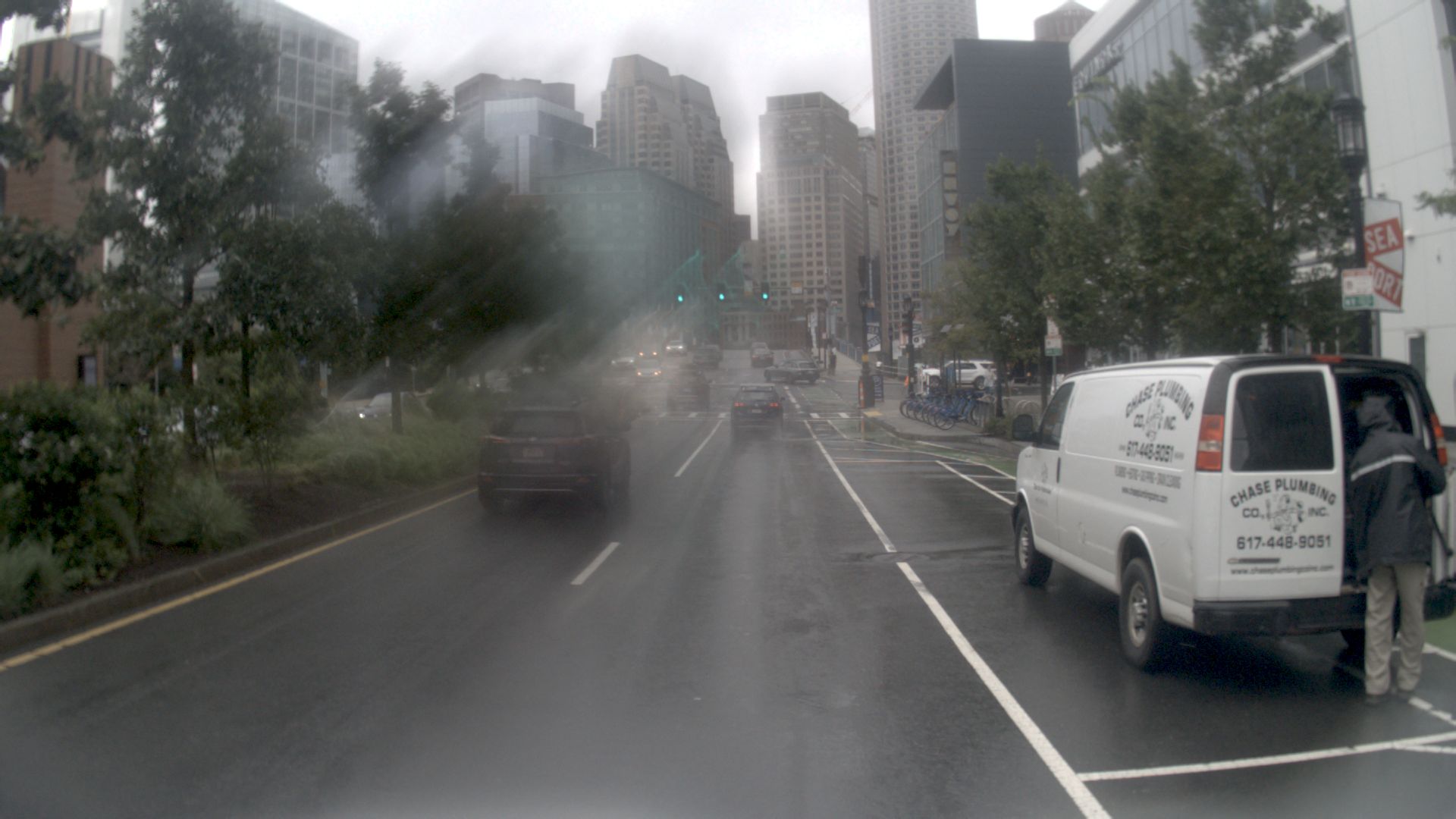}
\end{minipage}
\hspace{0.01\textwidth}%  % ← 可控的横向间距
\begin{minipage}[b]{0.22\textwidth}
\centering
\vspace{0.1cm}
\includegraphics[
    height=3.0cm,
    width=\linewidth,
    keepaspectratio,
    valign=b,
    trim={5 5 5 5},
    clip
]{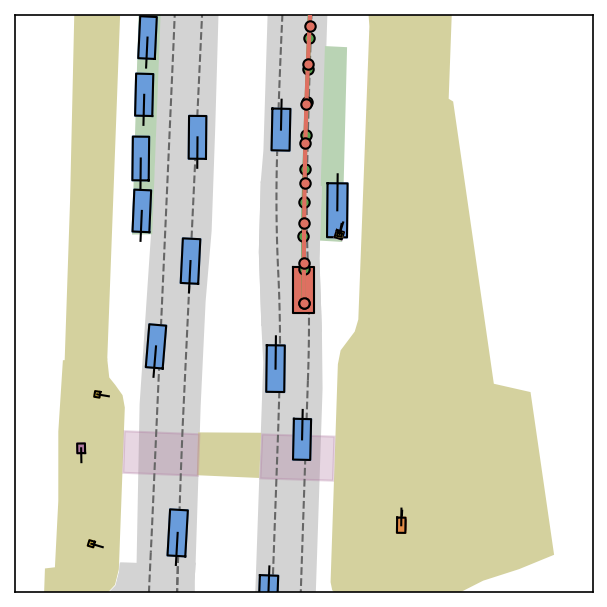}
\end{minipage}
\hspace{0.01\textwidth}%  % ← 可控的横向间距
\begin{minipage}[b]{0.22\textwidth}
\centering
\vspace{0.1cm}
\includegraphics[
    height=3.0cm,
    width=\linewidth,
    keepaspectratio,
    valign=b,
    trim={5 5 5 5},
    clip
]{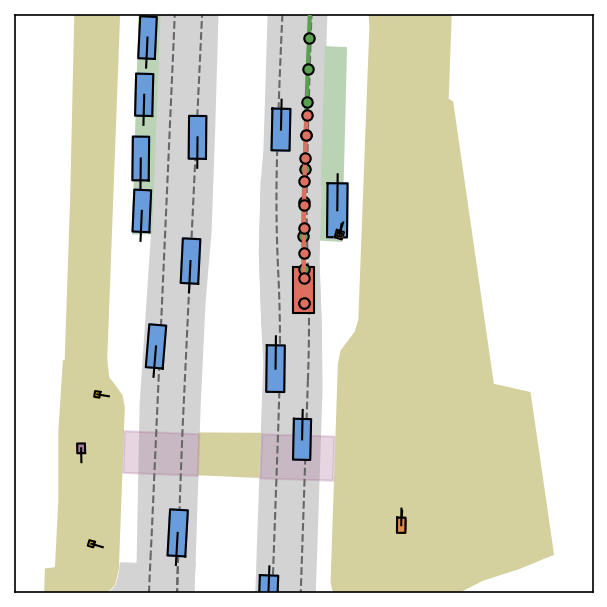}
\end{minipage}
% ===== 第一行：第一行图片 =====
\begin{minipage}[b]{0.44\textwidth}
\centering
\vspace{0.1cm}
\includegraphics[
    height=3.0cm,   
    width=\linewidth, 
    keepaspectratio, 
    valign=b,
    trim={0 150 0 0},
    clip
]{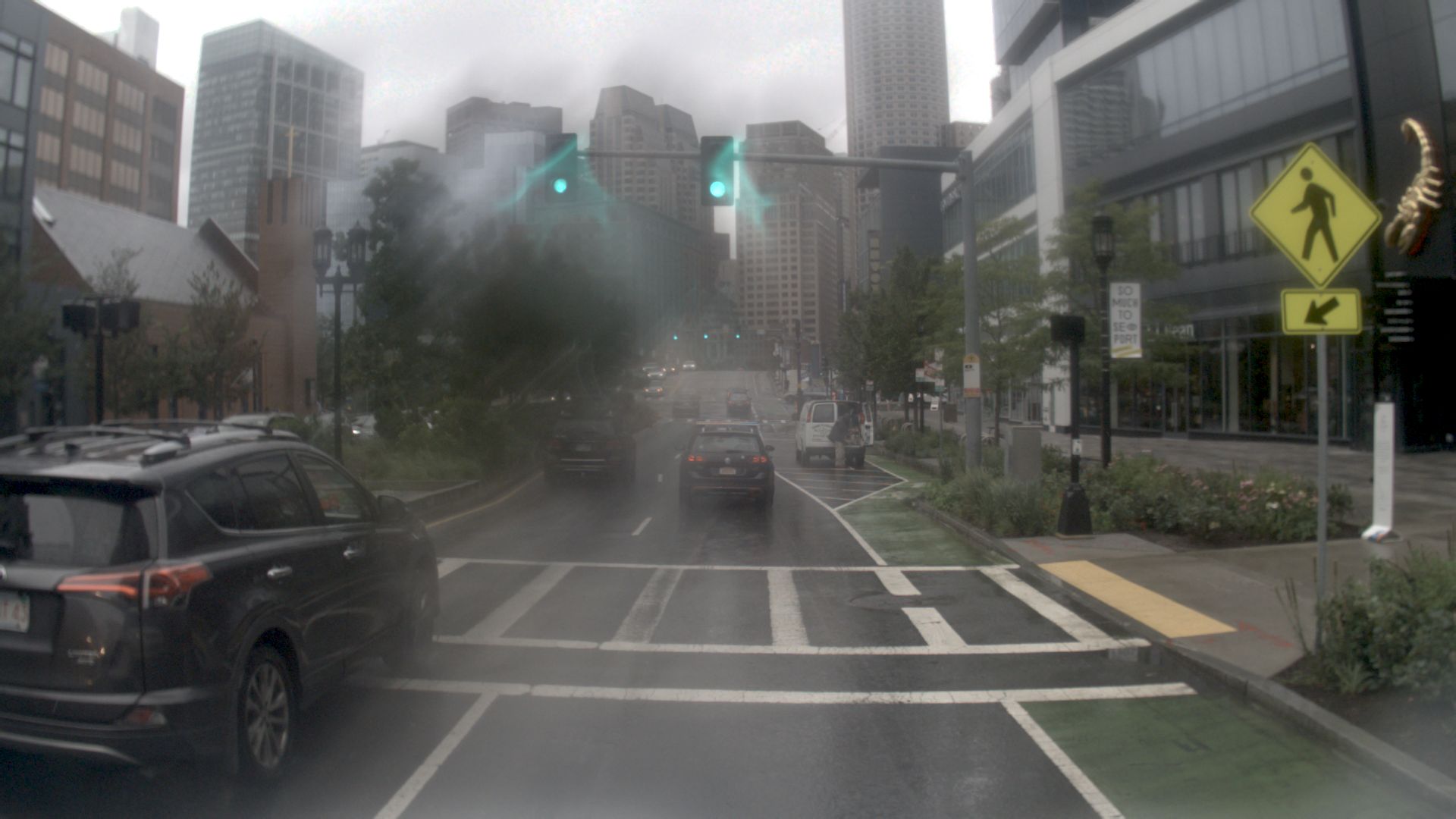}
\end{minipage}
\hspace{0.01\textwidth}%  % ← 可控的横向间距
\begin{minipage}[b]{0.22\textwidth}
\centering
\vspace{0.1cm}
\includegraphics[
    height=3.0cm,   
    width=\linewidth, 
    keepaspectratio, 
    valign=b,
    trim={5 5 5 5},
    clip
]{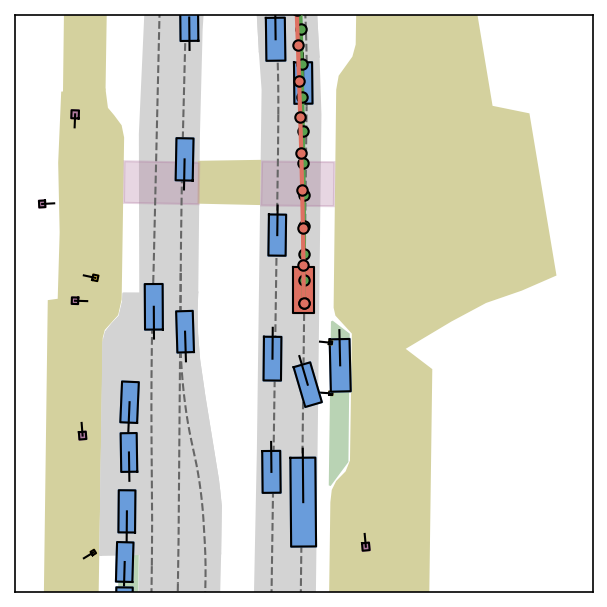}
\end{minipage}
\hspace{0.01\textwidth}%  % ← 可控的横向间距
\begin{minipage}[b]{0.22\textwidth}
\centering
\vspace{0.1cm}
\includegraphics[
    height=3.0cm,  
    width=\linewidth, 
    keepaspectratio, 
    valign=b,
    trim={5 5 5 5},
    clip
]{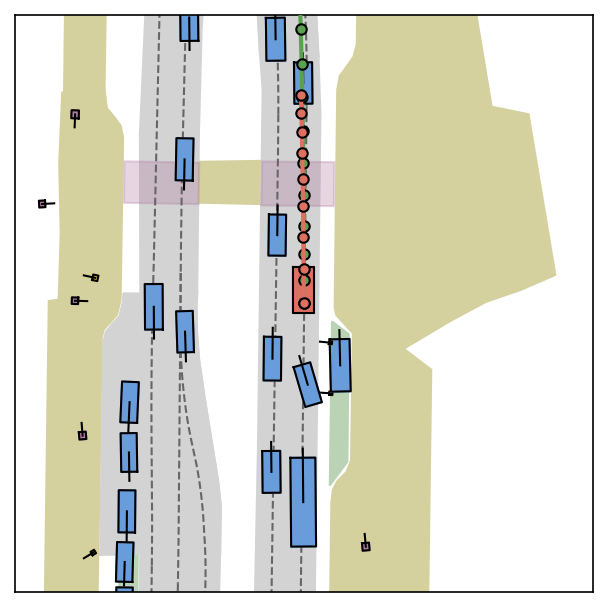}
\end{minipage}
% ===== 第三行：第三行图片 =====
\begin{minipage}[b]{0.44\textwidth}
\centering
\vspace{0.1cm}
\includegraphics[
    height=3.0cm,   
    width=\linewidth, 
    keepaspectratio, 
    valign=b,
    trim={0 150 0 0},
    clip
]{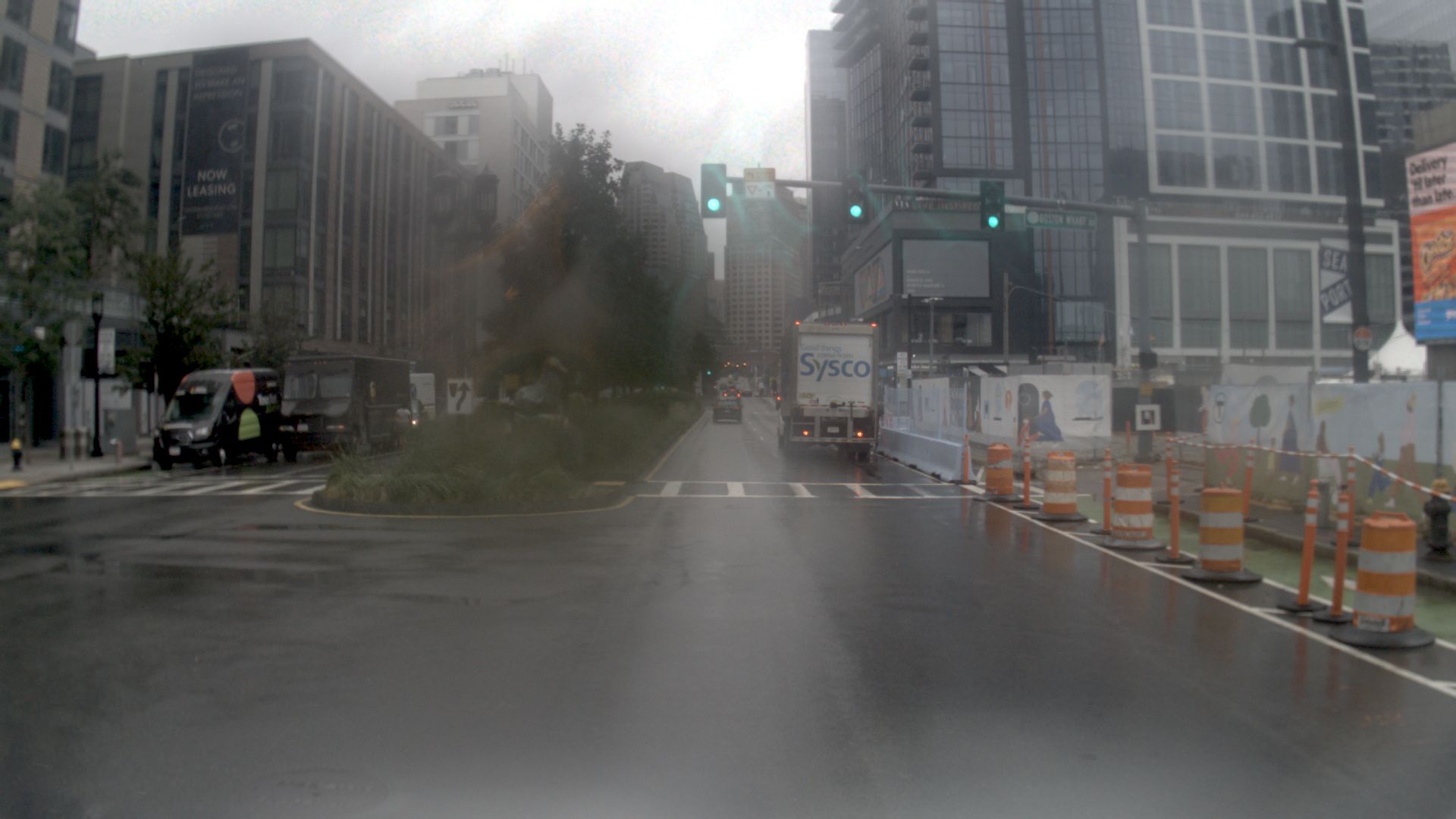}
\end{minipage}
\hspace{0.01\textwidth}%  % ← 可控的横向间距
\begin{minipage}[b]{0.22\textwidth}
\centering
\vspace{0.1cm}
\includegraphics[
    height=3.0cm,  
    width=\linewidth, 
    keepaspectratio, 
    valign=b,
    trim={5 5 5 5},
    clip
]{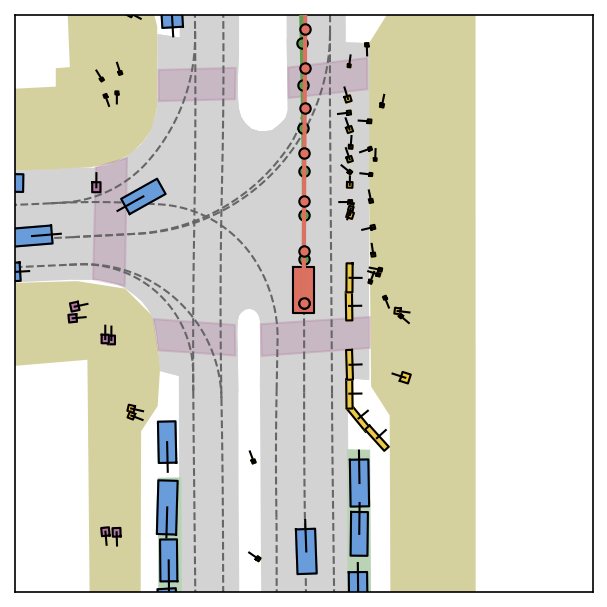}
\end{minipage}
\hspace{0.01\textwidth}%  % ← 可控的横向间距
\begin{minipage}[b]{0.22\textwidth}
\centering
\vspace{0.1cm}
\includegraphics[
    height=3.0cm,  
    width=\linewidth, 
    keepaspectratio, 
    valign=b,
    trim={5 5 5 5},
    clip
]{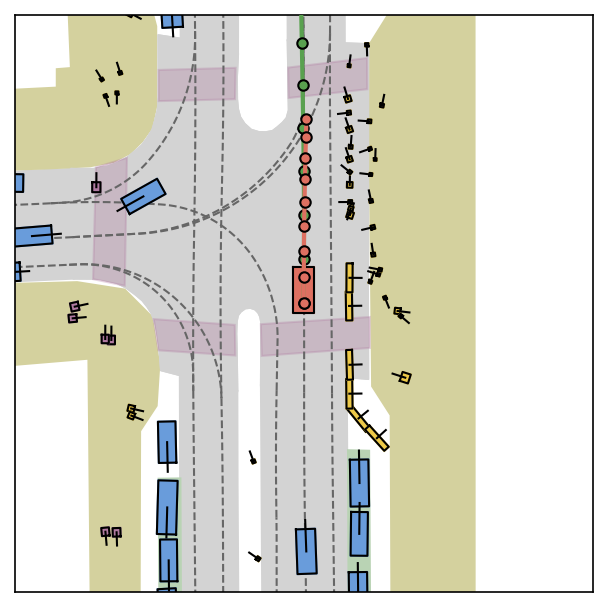}
\end{minipage}
% ===== 第四行：第四行图片 =====
\begin{minipage}[b]{0.44\textwidth}
\centering
\vspace{0.1cm}
\includegraphics[
    height=3.0cm,  
    width=\linewidth, 
    keepaspectratio, 
    valign=b,
    trim={0 150 0 0},
    clip
]{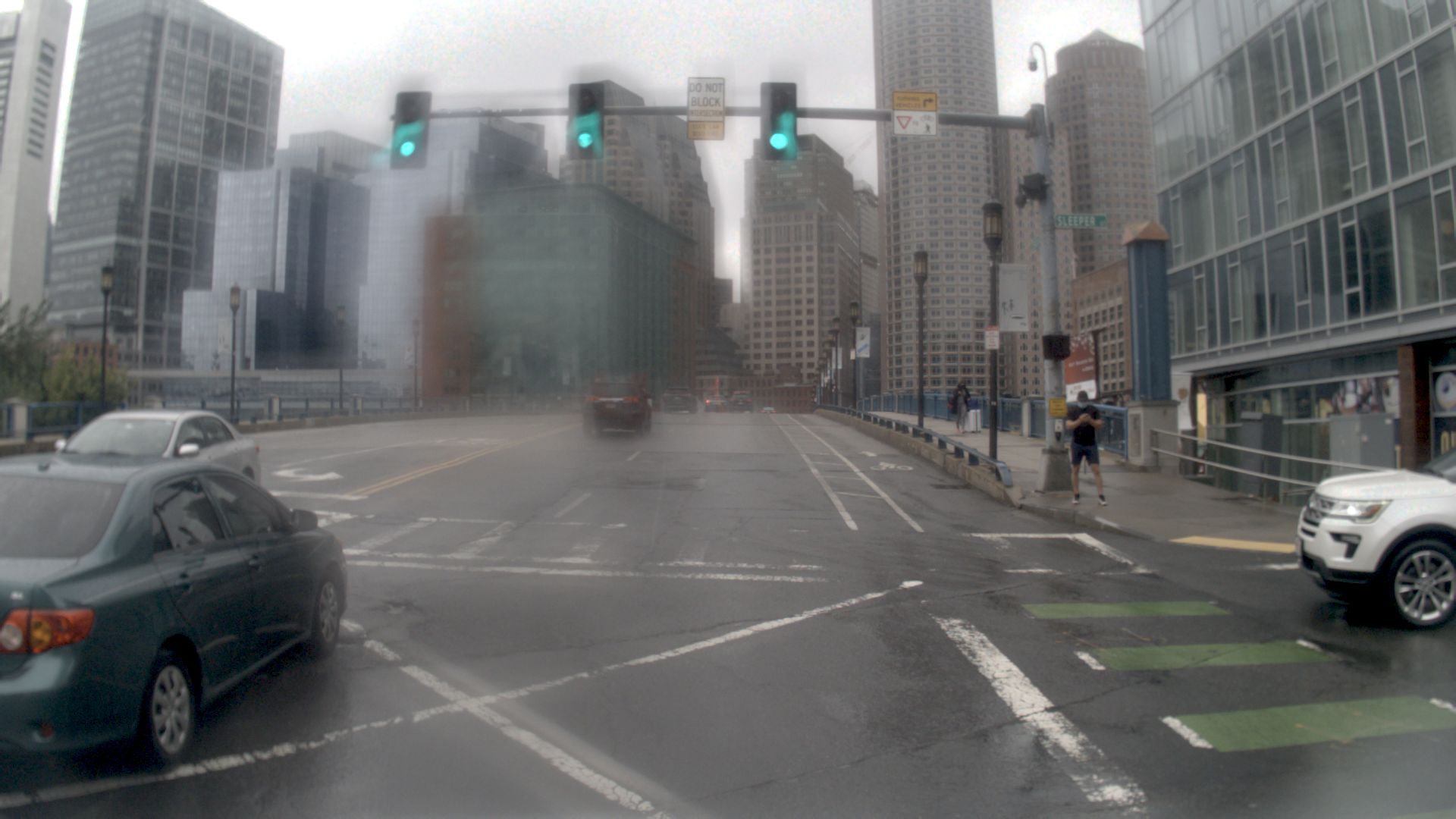}
\end{minipage}
\hspace{0.01\textwidth}%  % ← 可控的横向间距
\begin{minipage}[b]{0.22\textwidth}
\centering
\vspace{0.1cm}
\includegraphics[
    height=3.0cm,    
    width=\linewidth, 
    keepaspectratio, 
    valign=b,
    trim={5 5 5 5},
    clip
]{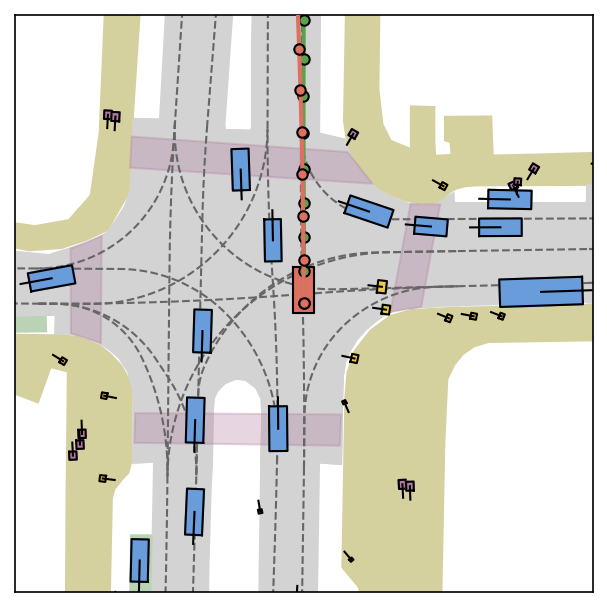}
\end{minipage}
\hspace{0.01\textwidth}%  % ← 可控的横向间距
\begin{minipage}[b]{0.22\textwidth}
\centering
\vspace{0.1cm}
\includegraphics[
    height=3.0cm, 
    width=\linewidth, 
    keepaspectratio, 
    valign=b,
    trim={5 5 5 5},
    clip
]{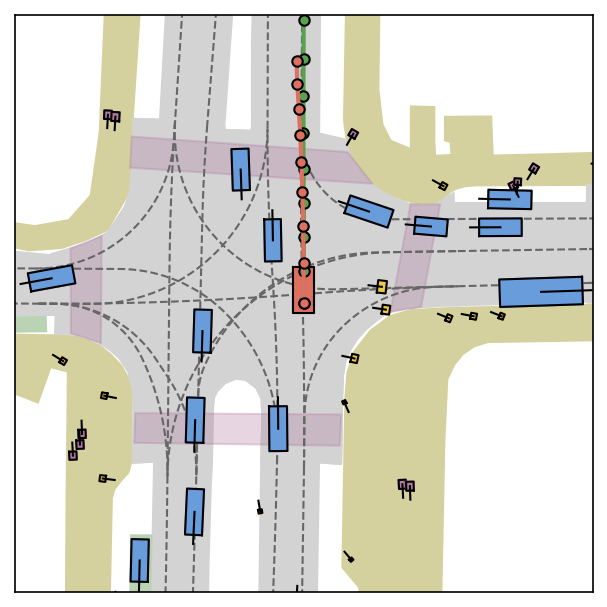}
\end{minipage}
\\[1em]%
% ========== 第五行：统一图例 ==========
\centering
\begin{tabular}{c @{\hspace{0.4em}} c @{\qquad} c @{\hspace{0.4em}} c @{\qquad} c @{\hspace{0.4em}} c @{\qquad} c @{\hspace{0.4em}} c @{\qquad} c @{\hspace{0.4em}} c}
  \textcolor{npccolor}{\rule{1.2em}{0.8em}} & NPC &
  \textcolor{egocolor}{\rule{1.2em}{0.8em}} & Ego Car &
  \textcolor{lanecolor}{\hdashrule[0.4ex]{1.6em}{0.05em}{2pt 2pt}} & Lane Center Line &
  \textcolor{gtcolor}{\Large$\bullet$} & GT Traj. &
  \textcolor{predcolor}{\Large$\bullet$} & Pred Traj. \\
\end{tabular}
% ========== 第六行：图题 ==========
\vspace{0.1cm} % 与图例的间距
\caption{\textbf{Go straight under the blur noise.}
DriveVLA-W0 baseline try to slow down due to the blurry image. Our method maintains a reasonable speed and is more aligned with human driving patterns.}
\label{fig:robustness_green_light_stop}
\end{figure}

To provide a balanced view of our model’s scene noise robustness, we perform qualitative analyses on illumination interference and blur noise cases. Our model exhibits more stable and reasonable performance under noise than the baseline. Besides the left-turn scenario in the main text, we compare output variations in typical right-turn and straight-ahead scenarios by adjusting illumination intensity alone, verifying that our model yields more consistent outputs under illumination perturbations, as shown in Figure~\ref{fig:robustness_to_light_intensity_right_turn} and Figure~\ref{fig:robustness_to_light_intensity_go_straight}.
We further present cases with natural non-uniform blur from rain and fog, where traffic lights and vehicles are partially or fully occluded. The pure pixel-regression baseline shows severe degradation: it crawls during red-light queuing and acts overly conservative to slow down. By contrast, benefiting from the LWM’s inherent scene noise robustness, our model maintains accurate and reasonable driving behaviors under the same noisy inputs, with trajectories more aligned with human driving patterns, as shown in Figure~\ref{fig:robustness_red_light_stop} and Figure~\ref{fig:robustness_green_light_stop}.
% =======================================
% ======appendix light dark fig 1========
% =======================================

% =======================================
% ======appendix blurry case fig 2=======
% =======================================

% =======================================
% ======appendix blurry case fig 1=======
% =======================================

% \section{MORE IMPLEMENTATION DETAILS}
% \label{sec:more_impl}
% \textbf{Flow matching based action experts} 

\section{Use of LLMs}
\label{sec:use_llms}
Large Language Models (LLMs) are employed to polish the writing in this manuscript.

%% file: ref.bib
@article{lipman2022flow,
  title={Flow matching for generative modeling},
  author={Lipman, Yaron and Chen, Ricky TQ and Ben-Hamu, Heli and Nickel, Maximilian and Le, Matt},
  journal={arXiv preprint arXiv:2210.02747},
  year={2022}
}

@inproceedings{rombach2022high,
  title={High-resolution image synthesis with latent diffusion models},
  author={Rombach, Robin and Blattmann, Andreas and Lorenz, Dominik and Esser, Patrick and Ommer, Bj{\"o}rn},
  booktitle={Proceedings of the IEEE/CVF conference on computer vision and pattern recognition},
  pages={10684--10695},
  year={2022}
}

@article{chung2024scaling,
  title={Scaling instruction-finetuned language models},
  author={Chung, Hyung Won and Hou, Le and Longpre, Shayne and Zoph, Barret and Tay, Yi and Fedus, William and Li, Yunxuan and Wang, Xuezhi and Dehghani, Mostafa and Brahma, Siddhartha and others},
  journal={Journal of Machine Learning Research},
  volume={25},
  number={70},
  pages={1--53},
  year={2024}
}

@article{tan2025latent,
  title={Latent Chain-of-Thought World Modeling for End-to-End Driving},
  author={Tan, Shuhan and Chitta, Kashyap and Chen, Yuxiao and Tian, Ran and You, Yurong and Wang, Yan and Luo, Wenjie and Cao, Yulong and Krahenbuhl, Philipp and Pavone, Marco and Ivanovic, Boris},
  journal={arXiv preprint arXiv:2512.10226},
  year={2025},
  note={Accepted to CVPR 2026}
}

@article{lu2026onevl,
  title={Xiaomi OneVL: One-Step Latent Reasoning and Planning with Vision-Language Explanation},
  author={Lu, Jinghui and Guan, Jiayi and Huang, Zhijian and Li, Jinlong and Li, Guang and Kong, Lingdong and Li, Yingyan and Wang, Han and Xu, Shaoqing and Luo, Yuechen and Li, Fang and Dang, Chenxu and Wang, Junli and Xu, Tao and Wu, Jing and Wu, Jianhua and Hao, Xiaoshuai and Zhang, Wen and Jiang, Tianyi and Zhang, Lingfeng and Zhou, Lei and Tang, Yingbo and Wang, Jie and Gao, Yinfeng and Jia, Feiyang and Liu, Lin and Ge, Yigu and Li, Hanbing and Shen, Yuannan and Cui, Jianwei and Xie, Hongwei and Wang, Bing},
  journal={arXiv preprint arXiv:2604.18486},
  year={2026}
}

@article{zeng2025futuresightdrive,
  title={FutureSightDrive: Thinking Visually with Spatio-Temporal CoT for Autonomous Driving},
  author={Zeng, Shuang and Chang, Xinyuan and Xie, Mengwei and Liu, Xinran and Bai, Yifan and Pan, Zheng and Xu, Mu and Wei, Xing and Guo, Ning},
  journal={arXiv preprint arXiv:2505.17685},
  year={2025},
  note={Accepted to NeurIPS 2025 as Spotlight}
}

@article{black2024pi0,
  title={$\pi_0$: A Vision-Language-Action Flow Model for General Robot Control},
  author={Black, Kevin and Brown, Noah and Driess, Danny and Esmail, Adnan and Equi, Michael and Finn, Chelsea and Fusai, Niccolo and Groom, Lachy and Hausman, Karol and Ichter, Brian and others},
  journal={arXiv preprint arXiv:2410.24164},
  year={2024}
}

@article{pertsch2025fast,
  title={FAST: Efficient Action Tokenization for Vision-Language-Action Models},
  author={Pertsch, Karl and Stachowicz, Kyle and Ichter, Brian and Driess, Danny and Nair, Suraj and Vuong, Quan and Mees, Oier and Finn, Chelsea and Levine, Sergey},
  journal={arXiv preprint arXiv:2501.09747},
  year={2025}
}

@article{fu2025orion,
  title={ORION: A Holistic End-to-End Autonomous Driving Framework by Vision-Language Instructed Action Generation},
  author={Fu, Haoyu and Zhang, Diankun and Zhao, Zongchuang and Cui, Jianfeng and Liang, Dingkang and Zhang, Chong and Zhang, Dingyuan and Xie, Hongwei and Wang, Bing and Bai, Xiang},
  journal={arXiv preprint arXiv:2503.19755},
  year={2025}
}

@article{zhou2025opendrivevla,
  title={OpenDriveVLA: Towards End-to-end Autonomous Driving with Large Vision Language Action Model},
  author={Zhou, Xingcheng and Han, Xuyuan and Yang, Feng and Ma, Yunpu and Tresp, Volker and Knoll, Alois},
  journal={arXiv preprint arXiv:2503.23463},
  year={2025}
}

@article{zheng2024doe1,
  title={Doe-1: Closed-Loop Autonomous Driving with Large World Model},
  author={Zheng, Wenzhao and Xia, Zetian and Huang, Yuanhui and Zuo, Sicheng and Zhou, Jie and Lu, Jiwen},
  journal={arXiv preprint arXiv:2412.09627},
  year={2024}
}

@inproceedings{wang2024drivedreamer,
  title={Drivedreamer: Towards real-world-drive world models for autonomous driving},
  author={Wang, Xiaofeng and Zhu, Zheng and Huang, Guan and Chen, Xinze and Zhu, Jiagang and Lu, Jiwen},
  booktitle={European conference on computer vision},
  pages={55--72},
  year={2024},
  organization={Springer}
}

@article{jia2023adriver,
  title={Adriver-i: A general world model for autonomous driving},
  author={Jia, Fan and Mao, Weixin and Liu, Yingfei and Zhao, Yucheng and Wen, Yuqing and Zhang, Chi and Zhang, Xiangyu and Wang, Tiancai},
  journal={arXiv preprint arXiv:2311.13549},
  year={2023}
}

@article{hu2023gaia,
  title={Gaia-1: A generative world model for autonomous driving},
  author={Hu, Anthony and Russell, Lloyd and Yeo, Hudson and Murez, Zak and Fedoseev, George and Kendall, Alex and Shotton, Jamie and Corrado, Gianluca},
  journal={arXiv preprint arXiv:2309.17080},
  year={2023}
}

@inproceedings{chen2025drivinggpt,
  title={Drivinggpt: Unifying driving world modeling and planning with multi-modal autoregressive transformers},
  author={Chen, Yuntao and Wang, Yuqi and Zhang, Zhaoxiang},
  booktitle={Proceedings of the IEEE/CVF International Conference on Computer Vision},
  pages={26890--26900},
  year={2025}
}

@inproceedings{esser2021taming,
  title={Taming transformers for high-resolution image synthesis},
  author={Esser, Patrick and Rombach, Robin and Ommer, Bjorn},
  booktitle={Proceedings of the IEEE/CVF conference on computer vision and pattern recognition},
  pages={12873--12883},
  year={2021}
}

@article{dauner2024navsim,
  title={Navsim: Data-driven non-reactive autonomous vehicle simulation and benchmarking},
  author={Dauner, Daniel and Hallgarten, Marcel and Li, Tianyu and Weng, Xinshuo and Huang, Zhiyu and Yang, Zetong and Li, Hongyang and Gilitschenski, Igor and Ivanovic, Boris and Pavone, Marco and others},
  journal={arXiv preprint arXiv:2406.15349},
  year={2024}
}

@misc{openscene2023,
  title={OpenScene: The Largest Up-to-Date 3D Occupancy Prediction Benchmark in Autonomous Driving},
  author={OpenScene Contributors},
  howpublished={\url{https://github.com/OpenDriveLab/OpenScene}},
  year={2023}
}

@article{cao2025pseudo,
  title={Pseudo-simulation for autonomous driving},
  author={Cao, Wei and Hallgarten, Marcel and Li, Tianyu and Dauner, Daniel and Gu, Xunjiang and Wang, Caojun and Miron, Yakov and Aiello, Marco and Li, Hongyang and Gilitschenski, Igor and others},
  journal={arXiv preprint arXiv:2506.04218},
  year={2025}
}

@article{caesar2021nuplan,
  title={nuplan: A closed-loop ml-based planning benchmark for autonomous vehicles},
  author={Caesar, Holger and Kabzan, Juraj and Tan, Kok Seang and Fong, Whye Kit and Wolff, Eric and Lang, Alex and Fletcher, Luke and Beijbom, Oscar and Omari, Sammy},
  journal={arXiv preprint arXiv:2106.11810},
  year={2021}
}

@article{xing2024large,
  title={Large motion video autoencoding with cross-modal video vae},
  author={Xing, Yazhou and Fei, Yang and He, Yingqing and Chen, Jingye and Xie, Jiaxin and Chi, Xiaowei and Chen, Qifeng},
  journal={arXiv preprint arXiv:2412.17805},
  year={2024}
}

@article{wang2024emu3,
  title={Emu3: Next-token prediction is all you need},
  author={Wang, Xinlong and Zhang, Xiaosong and Luo, Zhengxiong and Sun, Quan and Cui, Yufeng and Wang, Jinsheng and Zhang, Fan and Wang, Yueze and Li, Zhen and Yu, Qiying and others},
  journal={arXiv preprint arXiv:2409.18869},
  year={2024}
}

@misc{ali_ppu_2026,
  title        = {PPU Introduction},
  author       = {{Alibaba Cloud}},
  year         = {2026},
  howpublished = {[Online]},
  note         = {\url{https://help.aliyun.com/zh/document_detail/2864586.html}}
}

@article{hu2025vision,
  title={Vision-language-action models for autonomous driving: Past, present, and future},
  author={Hu, Tianshuai and Liu, Xiaolu and Wang, Song and Zhu, Yiyao and Liang, Ao and Kong, Lingdong and Zhao, Guoyang and Gong, Zeying and Cen, Jun and Huang, Zhiyu and others},
  journal={arXiv preprint arXiv:2512.16760},
  year={2025}
}

@article{wang2023drivemlm,
  title={Drive{MLM}: Aligning multi-modal large language models with behavioral planning states for autonomous driving},
  author={Wang, Wenhai and Xie, Jiangwei and Hu, ChuanYang and Zou, Haoming and Fan, Jianan and Tong, Wenwen and Wen, Yang and Wu, Silei and Deng, Hanming and Li, Zhiqi and others},
  journal={arXiv preprint arXiv:2312.09245},
  year={2023}
}

@article{xu2024drivegpt4,
  title={Drive{GPT}4: Interpretable end-to-end autonomous driving via large language model},
  author={Xu, Zhenhua and Zhang, Yujia and Xie, Enze and Zhao, Zhen and Guo, Yong and Wong, Kwan-Yee K and Li, Zhenguo and Zhao, Hengshuang},
  journal={IEEE Robotics and Automation Letters},
  volume={9},
  number={10},
  pages={8186--8193},
  year={2024},
  publisher={IEEE}
}

@article{jiang2025alphadrive,
  title={Alpha{D}rive: Unleashing the power of {VLM}s in autonomous driving via reinforcement learning and reasoning},
  author={Jiang, Bo and Chen, Shaoyu and Zhang, Qian and Liu, Wenyu and Wang, Xinggang},
  journal={arXiv preprint arXiv:2503.07608},
  year={2025}
}

@inproceedings{sima2024drivelm,
  title={Drive{LM}: Driving with graph visual question answering},
  author={Sima, Chonghao and Renz, Katrin and Chitta, Kashyap and Chen, Li and Zhang, Hanxue and Xie, Chengen and Bei{\ss}wenger, Jens and Luo, Ping and Geiger, Andreas and Li, Hongyang},
  booktitle={European conference on computer vision},
  pages={256--274},
  year={2024},
  organization={Springer}
}

@article{hwang2024emma,
  title={{EMMA}: End-to-end multimodal model for autonomous driving},
  author={Hwang, Jyh-Jing and Xu, Runsheng and Lin, Hubert and Hung, Wei-Chih and Ji, Jingwei and Choi, Kristy and Huang, Di and He, Tong and Covington, Paul and Sapp, Benjamin and others},
  journal={arXiv preprint arXiv:2410.23262},
  year={2024}
}

@inproceedings{hu2023planning,
  title={Planning-oriented autonomous driving},
  author={Hu, Yihan and Yang, Jiazhi and Chen, Li and Li, Keyu and Sima, Chonghao and Zhu, Xizhou and Chai, Siqi and Du, Senyao and Lin, Tianwei and Wang, Wenhai and others},
  booktitle={Proceedings of the IEEE/CVF conference on computer vision and pattern recognition},
  pages={17853--17862},
  year={2023}
}

@inproceedings{prakash2021multi,
  title={Multi-modal fusion transformer for end-to-end autonomous driving},
  author={Prakash, Aditya and Chitta, Kashyap and Geiger, Andreas},
  booktitle={Proceedings of the IEEE/CVF conference on computer vision and pattern recognition},
  pages={7077--7087},
  year={2021}
}

@inproceedings{weng2024drive,
  title={Para-drive: Parallelized architecture for real-time autonomous driving},
  author={Weng, Xinshuo and Ivanovic, Boris and Wang, Yan and Wang, Yue and Pavone, Marco},
  booktitle={Proceedings of the IEEE/CVF Conference on Computer Vision and Pattern Recognition},
  pages={15449--15458},
  year={2024}
}

@article{li2024hydra,
  title={Hydra-mdp: End-to-end multimodal planning with multi-target hydra-distillation},
  author={Li, Zhenxin and Li, Kailin and Wang, Shihao and Lan, Shiyi and Yu, Zhiding and Ji, Yishen and Li, Zhiqi and Zhu, Ziyue and Kautz, Jan and Wu, Zuxuan and others},
  journal={arXiv preprint arXiv:2406.06978},
  year={2024}
}

@inproceedings{liao2025diffusiondrive,
  title={Diffusiondrive: Truncated diffusion model for end-to-end autonomous driving},
  author={Liao, Bencheng and Chen, Shaoyu and Yin, Haoran and Jiang, Bo and Wang, Cheng and Yan, Sixu and Zhang, Xinbang and Li, Xiangyu and Zhang, Ying and Zhang, Qian and others},
  booktitle={Proceedings of the Computer Vision and Pattern Recognition Conference},
  pages={12037--12047},
  year={2025}
}

@article{zhou2025autovla,
  title={AutoVLA: A Vision-Language-Action Model for End-to-End Autonomous Driving with Adaptive Reasoning and Reinforcement Fine-Tuning},
  author={Zhou, Zewei and Cai, Tianhui and Zhao, Seth Z. and Zhang, Yun and Huang, Zhiyu and Zhou, Bolei and Ma, Jiaqi},
  journal={arXiv preprint arXiv:2506.13757},
  year={2025},
  note={Accepted by NeurIPS 2025}
}

@article{li2025recogdrive,
  title={ReCogDrive: A Reinforced Cognitive Framework for End-to-End Autonomous Driving},
  author={Li, Yongkang and Xiong, Kaixin and Guo, Xiangyu and Li, Fang and Yan, Sixu and Xu, Gangwei and Zhou, Lijun and Chen, Long and Sun, Haiyang and Wang, Bing and Ma, Kun and Chen, Guang and Ye, Hangjun and Liu, Wenyu and Wang, Xinggang},
  journal={arXiv preprint arXiv:2506.08052},
  year={2025}
}

@inproceedings{li2025end,
  title={End-to-end driving with online trajectory evaluation via bev world model},
  author={Li, Yingyan and Wang, Yuqi and Liu, Yang and He, Jiawei and Fan, Lue and Zhang, Zhaoxiang},
  booktitle={Proceedings of the IEEE/CVF International Conference on Computer Vision},
  pages={27137--27146},
  year={2025}
}

@article{zeng2026futuresightdrive,
  title={Futuresightdrive: Thinking visually with spatio-temporal cot for autonomous driving},
  author={Zeng, Shuang and Chang, Xinyuan and Xie, Mengwei and Liu, Xinran and Bai, Yifan and Pan, Zheng and Xu, Mu and Wei, Xing},
  journal={Advances in Neural Information Processing Systems},
  volume={38},
  pages={67299--67318},
  year={2026}
}

@article{yang2026resim,
  title={Resim: Reliable world simulation for autonomous driving},
  author={Yang, Jiazhi and Chitta, Kashyap and Gao, Shenyuan and Chen, Long and Shao, Yuqian and Jia, Xiaosong and Li, Hongyang and Geiger, Andreas and Yue, Xiangyu and Chen, Li},
  journal={Advances in Neural Information Processing Systems},
  volume={38},
  pages={167710--167741},
  year={2026}
}

@article{zhao2026forecasting,
  title={From forecasting to planning: Policy world model for collaborative state-action prediction},
  author={Zhao, Zhida and Fu, Talas and Wang, Yifan and Wang, Lijun and Lu, Huchuan},
  journal={Advances in Neural Information Processing Systems},
  volume={38},
  pages={134585--134611},
  year={2026}
}

@article{huang2026coworld,
  title={CoWorld-VLA: Thinking in a Multi-Expert World Model for Autonomous Driving},
  author={Huang, Minqing and Xiang, Yujiao and Liang, Zihan and Huang, Jiajie and Wang, Jingqi and Xu, Zhi and Tan, Feiyang and Zhou, Hangning and Yang, Mu and Che, Gong},
  journal={arXiv preprint arXiv:2605.10426},
  year={2026}
}

@inproceedings{xia2026drivelaw,
  title={Drivelaw: Unifying planning and video generation in a latent driving world},
  author={Xia, Tianze and Li, Yongkang and Zhou, Lijun and Yao, Jingfeng and Xiong, Kaixin and Sun, Haiyang and Wang, Bing and Ma, Kun and Chen, Guang and Ye, Hangjun and others},
  booktitle={Proceedings of the IEEE/CVF Conference on Computer Vision and Pattern Recognition},
  pages={39701--39712},
  year={2026}
}

@article{li2025drivevla,
  title={DriveVLA-W0: World Models Amplify Data Scaling Law in Autonomous Driving},
  author={Li, Yingyan and Shang, Shuyao and Liu, Weisong and Zhan, Bing and Wang, Haochen and Wang, Yuqi and Chen, Yuntao and Wang, Xiaoman and An, Yasong and Tang, Chufeng and Hou, Lu and Fan, Lue and Zhang, Zhaoxiang},
  journal={arXiv preprint arXiv:2510.12796},
  year={2025}
}

@article{yang2026dreamerad,
  title={Dreamerad: Efficient reinforcement learning via latent world model for autonomous driving},
  author={Yang, Pengxuan and Zheng, Yupeng and Qian, Deheng and Xing, Zebin and Zhang, Qichao and Wang, Linbo and Zhang, Yichen and Guo, Shaoyu and Xia, Zhongpu and Chen, Qiang and others},
  journal={arXiv preprint arXiv:2603.24587},
  year={2026}
}

@inproceedings{zhang2025epona,
  title={Epona: Autoregressive diffusion world model for autonomous driving},
  author={Zhang, Kaiwen and Tang, Zhenyu and Hu, Xiaotao and Pan, Xingang and Guo, Xiaoyang and Liu, Yuan and Huang, Jingwei and Yuan, Li and Zhang, Qian and Long, Xiao-Xiao and others},
  booktitle={Proceedings of the IEEE/CVF International Conference on Computer Vision},
  pages={27220--27230},
  year={2025}
}

@inproceedings{zheng2025world4drive,
  title={World4drive: End-to-end autonomous driving via intention-aware physical latent world model},
  author={Zheng, Yupeng and Yang, Pengxuan and Xing, Zebin and Zhang, Qichao and Zheng, Yuhang and Gao, Yinfeng and Li, Pengfei and Zhang, Teng and Xia, Zhongpu and Jia, Peng and others},
  booktitle={Proceedings of the IEEE/CVF International Conference on Computer Vision},
  pages={28632--28642},
  year={2025}
}

@inproceedings{li2025enhancing,
  title={Enhancing end-to-end autonomous driving with latent world model},
  author={Li, Yingyan and Fan, Lue and He, Jiawei and Wang, Yuqi and Chen, Yuntao and Zhang, Zhaoxiang and Tan, Tieniu},
  booktitle={International Conference on Learning Representations},
  volume={2025},
  pages={42942--42959},
  year={2025}
}

@article{lu2026xiaomi,
  title={Xiaomi OneVL: One-Step Latent Reasoning and Planning with Vision-Language Explanation},
  author={Lu, Jinghui and Guan, Jiayi and Huang, Zhijian and Li, Jinlong and Li, Guang and Kong, Lingdong and Li, Yingyan and Wang, Han and Xu, Shaoqing and Luo, Yuechen and others},
  journal={arXiv preprint arXiv:2604.18486},
  year={2026}
}

@inproceedings{yang2026worldrft,
  title={Worldrft: Latent world model planning with reinforcement fine-tuning for autonomous driving},
  author={Yang, Pengxuan and Lu, Ben and Xia, Zhongpu and Han, Chao and Gao, Yinfeng and Zhang, Teng and Zhan, Kun and Lang, XianPeng and Zheng, Yupeng and Zhang, Qichao},
  booktitle={Proceedings of the AAAI Conference on Artificial Intelligence},
  volume={40},
  number={14},
  pages={11649--11657},
  year={2026}
}

@article{zhang2026resworld,
  title={Resworld: Temporal residual world model for end-to-end autonomous driving},
  author={Zhang, Jinqing and Fu, Zehua and Xu, Zelin and Dai, Wenying and Liu, Qingjie and Wang, Yunhong},
  journal={arXiv preprint arXiv:2602.10884},
  year={2026}
}

@article{li2025hydra,
  title={Hydra-MDP++: Advancing End-to-End Driving via Hydra-Distillation with Expert-Guided Decision Analysis},
  author={Li, Kailin and Li, Zhenxin and Lan, Shiyi and Liu, Jiayi and Xie, Yuan and Wu, Zuxuan and Yu, Zhiding and Alvarez, Jose M and others},
  year={2025}
}

@inproceedings{yao2026drivesuprim,
  title={Drivesuprim: Towards precise trajectory selection for end-to-end planning},
  author={Yao, Wenhao and Li, Zhenxin and Lan, Shiyi and Wang, Zi and Sun, Xinglong and Alvarez, Jose M and Wu, Zuxuan},
  booktitle={Proceedings of the AAAI Conference on Artificial Intelligence},
  volume={40},
  number={14},
  pages={11910--11918},
  year={2026}
}

@article{feng2025artemis,
  title={Artemis: Autoregressive end-to-end trajectory planning with mixture of experts for autonomous driving},
  author={Feng, Renju and Xi, Ning and Chu, Duanfeng and Wang, Rukang and Deng, Zejian and Wang, Anzheng and Lu, Liping and Wang, Jinxiang and Huang, Yanjun},
  journal={IEEE Robotics and Automation Letters},
  volume={11},
  number={1},
  pages={226--233},
  year={2025},
  publisher={IEEE}
}

@article{sheng2026explorevla,
  title={ExploreVLA: Dense World Modeling and Exploration for End-to-End Autonomous Driving},
  author={Sheng, Zihao and Ye, Xin and Luo, Jingru and Chen, Sikai and Ren, Liu},
  journal={arXiv preprint arXiv:2604.02714},
  year={2026}
}

@article{liu2026driveworld,
  title={DriveWorld-VLA: Unified Latent-Space World Modeling with Vision-Language-Action for Autonomous Driving},
  author={Liu, Lin and Song, Ziying and Jia, Caiyan and Ye, Hangjun and Hao, Xiaoshuai and Chen, Long and others},
  journal={arXiv preprint arXiv:2602.06521},
  year={2026}
}

@article{wang2026latent,
  title={Latent-wam: Latent world action modeling for end-to-end autonomous driving},
  author={Wang, Linbo and Zheng, Yupeng and Chen, Qiang and Li, Shiwei and Zhang, Yichen and Xing, Zebin and Zhang, Qichao and Li, Xiang and Qian, Deheng and Yang, Pengxuan and others},
  journal={arXiv preprint arXiv:2603.24581},
  year={2026}
}

@article{tu2025role,
  title={The role of world models in shaping autonomous driving: A comprehensive survey},
  author={Tu, Sifan and Zhou, Xin and Liang, Dingkang and Jiang, Xingyu and Zhang, Yumeng and Li, Xiaofan and Bai, Xiang},
  journal={arXiv preprint arXiv:2502.10498},
  year={2025}
}
